\documentclass[10pt,twocolumn,letterpaper]{article}

\usepackage[pagenumbers]{cvpr} 
\usepackage{graphicx, animate}
\usepackage{algpseudocode}
\usepackage{algorithm}
\usepackage{multirow}
\usepackage{mathtools}
\usepackage{wrapfig}
\usepackage{tocloft}

\definecolor{cvprblue}{rgb}{0.21,0.49,0.74}
\newcommand{\zeropad}[1]{\ifnum#1<10 00#1\else\ifnum#1<100 0#1\else#1\fi\fi}
\usepackage[pagebackref,breaklinks,colorlinks,allcolors=cvprblue]{hyperref}

\def\paperID{7568} 
\def\confName{CVPR}
\def\confYear{2025}

\title{DOLLAR: Few-Step Video Generation via Distillation and Latent Reward Optimization}

\author{Zihan Ding\textsuperscript{1}\thanks{The work is done during internship at Adobe Research.}, Chi Jin\textsuperscript{1}, Difan Liu\textsuperscript{2}, Haitian Zheng\textsuperscript{2}, Krishna Kumar Singh\textsuperscript{2}, \\
 Qiang Zhang\textsuperscript{2}, Yan Kang\textsuperscript{2}, Zhe Lin\textsuperscript{2}, Yuchen Liu\textsuperscript{2}\thanks{Corresponding Author.}
\\
\noindent\textsuperscript{1}Princeton University, \noindent\textsuperscript{2}Adobe Research\\
{\tt\small \noindent\textsuperscript{1}\{zihand, chij\}@princeton.edu} \\
{\tt\small \noindent\textsuperscript{2}\{diliu, hazheng, krishsin, qiangz, yankang, zlin, yuliu\}@adobe.com}\\
Project Page: \href{https://quantumiracle.github.io/dollar/}{https://quantumiracle.github.io/dollar/}
}

\begin{document}
\maketitle
\begin{abstract}
Diffusion probabilistic models have shown significant progress in video generation; however, their computational efficiency is limited by the large number of sampling steps required. Reducing sampling steps often compromises video quality or generation diversity. In this work, we introduce a distillation method that combines variational score distillation and consistency distillation to achieve few-step video generation, maintaining both high quality and diversity. We also propose a latent reward model fine-tuning approach to further enhance video generation performance according to any specified reward metric. This approach reduces memory usage and does not require the reward to be differentiable. Our method demonstrates state-of-the-art performance in few-step generation for 10-second videos (128 frames at 12 FPS). The distilled student model achieves a score of 82.57 on VBench, surpassing the teacher model as well as baseline models Gen-3, T2V-Turbo~\cite{li2024t2v}, and Kling~\cite{kuaishou2024}. One-step distillation accelerates the teacher model’s diffusion sampling by up to 278.6 times, enabling near real-time generation. Human evaluations further validate the superior performance of our 4-step student models compared to teacher model using 50-step DDIM sampling.
\end{abstract}
    
\addtocontents{toc}{\protect\setcounter{tocdepth}{-2}}
\section{Introduction}
\label{sec:intro}

Diffusion probabilistic models~\cite{sohl2015deep, song2019generative, ho2020denoising, song2021score} have recently revolutionized generative modeling in continuous domains.  With remarkable expressive power and flexibility across diverse data formats and modalities, diffusion models have significant breakthroughs in tasks such as text-to-image and text-to-video (T2V) generation. However, despite substantial improvements in generation quality, the efficiency of diffusion models remains a limiting factor in practical applications due to the inherently large number of iterative sampling steps.  This efficiency challenge is exacerbated in video generative modeling, where the higher-dimensional space demands larger model sizes, more extensive training data, larger input and output tensors, and more sampling iterations. Furthermore, practical applications often require generation qualities that may differ from the training distribution—such as higher aesthetic standards or diverse stylistic choices—necessitating efficient post-training adjustments or fine-tuning to meet specific requirements while managing the substantial cost of pre-training.

\begin{figure}[b]
    \centering
\includegraphics[width=\columnwidth]{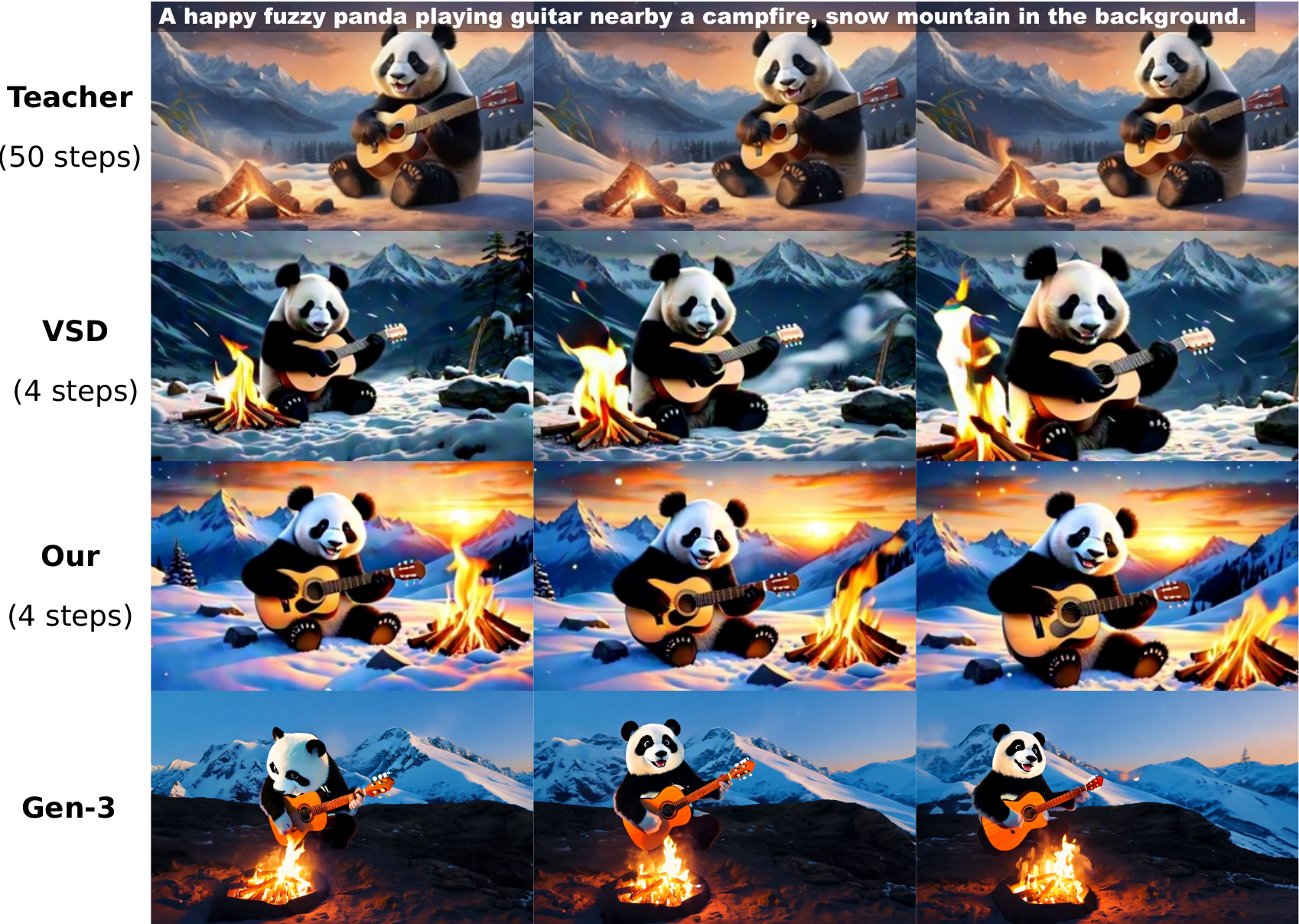}
    \caption{By incorporating variational score distillation, consistency distillation and latent reward fine-tuning, our method generates high-quality videos with 4-step sampling, $\times15.6$ acceleration compared with teacher. More visualized examples see Appendix Sec.~\ref{app_sec:visual} and \href{https://quantumiracle.github.io/dollar/}{project page}.
    }
    \label{fig:teaser}
\end{figure}

To address the efficiency challenges in diffusion models, model distillation~\cite{polino2018model, luhman2021knowledge, salimans2022progressive} has been widely researched across various models and domains. Score distillation, specifically, aims to improve efficiency in 3D~\cite{wang2023score, poole2022dreamfusion, wang2024prolificdreamer} and image synthesis~\cite{yin2024one, luo2024diff, xie2024distillation, salimans2024multistep} by aligning the distribution between teacher and student diffusion models. However, despite achieving high sample fidelity, it often encounters model collapse issues~\cite{yin2024one,lu2024simplifying}. Another approach, consistency distillation (CD)~\cite{song2023consistency}, seeks to ensure consistent sample predictions along the diffusion trajectory. While CD promotes greater sample diversity, it has limitations: it tends to lower sample fidelity and can produce overly smooth outputs in large-scale T2V applications.

A further challenge with distillation methods is that the student's performance is typically upper-bounded by the teacher model. Previous efforts to address this limitation have involved integrating variational score distillation (VSD)~\cite{wang2024prolificdreamer, yin2024improved} or consistency distillation~\cite{kim2023consistency} with GAN~\cite{goodfellow2014generative} loss, which modestly enhances sample fidelity within the training distribution, limited by the sparsity of discriminative signals of adversarial training. Consequently, the generated samples may still fall short in capturing nuanced visual quality details and text-to-image alignment, both of which require denser feedback signals. Recently developed image or video reward models offer promising potential to address this gap, providing richer signals for fine-grained improvements in generation quality.
 
In this work, we address the limitations of consistency distillation (CD) by incorporating a larger number of teacher denoising steps and combining CD with variational score distillation (VSD) to produce high-quality, diverse samples with a few-step model after distillation. However, this alone does not suffice to outperform the teacher model or reliably meet specific preferences for downstream applications, as generated samples may still face challenges in visual quality and text-to-video alignment. While model post-training with a high-quality dataset is a potential solution, it is often costly to implement. To overcome these limitations, we further introduce an efficient reward model fine-tuning method that enhances the student model beyond the teacher's capabilities and aligns it with any pre-defined requirements through tailored reward metrics. The improved performance is shown in Fig.~\ref{fig:teaser}.

We propose learning a dual reward model within the latent space, guided by the pixel-space reward model, and utilize the gradients from this latent reward model (LRM) to fine-tune the diffusion model directly. This approach combines the strengths of reward-gradient methods in pixel space and stochastic policy gradient methods, offering several advantages: (1) it harnesses the rich gradient information from the latent reward model, enabling efficient and effective tuning; (2) it does not require the original reward model to be differentiable, broadening applicability to a variety of reward models; (3) it significantly reduces computational and memory costs during fine-tuning by eliminating the need for backpropagation through large pixel-space reward models and the decoder. The LRM approach is versatile, accommodating various reward types—including image, video, text-image, and text-video rewards—thereby enhancing practical usability.

In summary, our contributions are threefold: (1) We introduce a diffusion model distillation method that combines VSD and CD losses to enable efficient, few-step T2V models; (2) We enhance CD with a generalized approach incorporating multiple teacher denoising steps to improve its effectiveness; (3) We propose to use a compact latent-space reward model for reward-based fine-tuning, which posts no requirement on the differentiability of original reward metrics and is more memory- and computation-efficient. All evaluations are conducted on large-scale T2V settings. Putting together these innovations, we present \texttt{DOLLAR} method with \textbf{D}isti\textbf{l}lation and \textbf{La}tent \textbf{R}eward \textbf{O}ptimization, to significantly advance the quality and efficiency of video generation and pave the way for real-time applications.

\section{Related Work}
\paragraph{Video Generation.}

Recent advancements have extended diffusion models from image synthesis to video generation, addressing the complexities of spatiotemporal data. Pioneering works like Video Diffusion Models \cite{ho2022video} adapted diffusion processes to handle temporal dynamics, enabling the creation of coherent and high-fidelity video clips. To enhance computational efficiency, Latent Diffusion Models (LDM) \cite{rombach2022high} perform diffusion modeling in compressed latent spaces, a strategy further refined for video by \cite{blattmann2023align}, \cite{harvey2022flexible}, Stable Video Diffusion \cite{blattmann2023stable} and VideoCrafter2 \cite{chen2024videocrafter2}. Text-to-video generation has progressed with models like Imagen Video \cite{ho2022video}, Make-A-Video \cite{singer2022make}, Phenaki \cite{villegas2022phenaki}, CogVideo \cite{hong2022cogvideo}, CogVideoX \cite{yang2024cogvideox}, Text2Video-Zero \cite{khachatryan2023text2video}, and ModelScopeT2V~\cite{wang2023modelscope}, which generate videos conditioned on textual descriptions, as known as the text-to-video (T2V) models. Hybrid approaches, such as Dual Diffusion Models \cite{xiao2023dual}, combine diffusion models with other generative frameworks to improve temporal coherence and resolution. There are also some recent advanced methods, like Lumiere~\cite{bar2024lumiere}, SF-V~\cite{zhang2024sf}, LaVie~\cite{wang2023lavie}, Pyramidal Flow Matching~\cite{jin2024pyramidal}. Diffusion transformer (DiT)~\cite{peebles2023scalable} further improves the scalability of the diffusion models by incorporating the transformer architecture, which allows to accommodate training videos with various resolutions and durations~\cite{polyak2024movie}. Despite these advancements, challenges like computational cost, temporal consistency, and suitable evaluation metrics remain, guiding future research in diffusion model-based video generation.

\begin{figure*}[htbp]
    \centering
\includegraphics[width=0.95\textwidth]{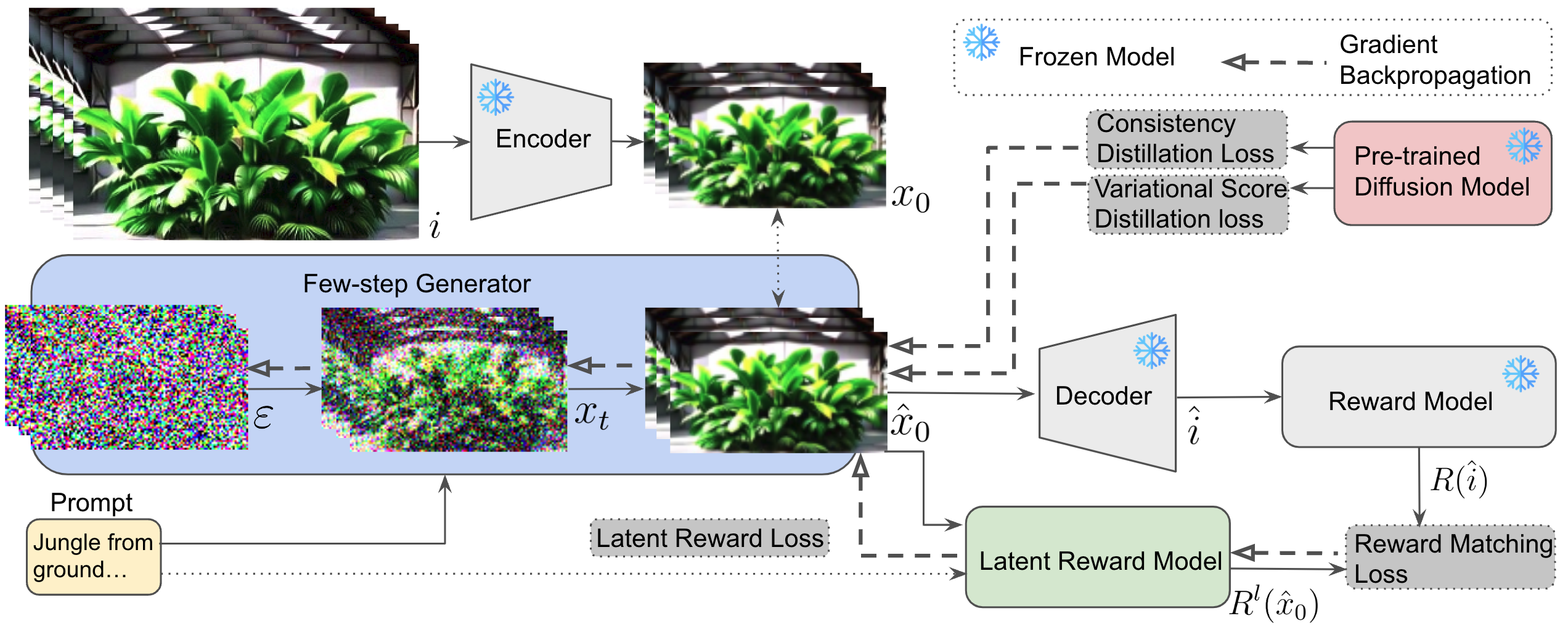}
    \caption{
    Method Overview: The few-step generator $G_\theta$ is trained to generate high-quality samples from random noise in latent space, guided by a combination of variational score distillation (VSD), consistency distillation (CD), and latent reward model (LRM) fine-tuning objectives. VSD loss enhances sample quality, albeit with a risk of mode collapse, while CD loss increases sample diversity without compromising generation quality. The LRM enables reward-based optimization to further improve sample quality, by bypassing the large, pixel-space reward model and the decoder, thereby reducing memory usage and removing the need for differentiable reward models.
    }
    \label{fig:diagram}
\end{figure*}

\paragraph{Efficiency of Diffusion Models.}

Diffusion models have achieved state-of-the-art results in generative tasks but are computationally intensive, requiring hundreds of sampling steps. DDIM \cite{song2021denoising} reduced the number of steps at inference time, but performance degrades significantly if it comes into the few-step regime. To address this, knowledge distillation for generative models is proposed to transfer knowledge from pre-trained teachers to students~\cite{luhman2021knowledge}.
Progressive Distillation \cite{salimans2022progressive} condensed multiple iterations into a single forward pass. With the distribution matching objective between the teacher and student models, score distillation is initially proposed for 3D generation with diffusion models~\cite{poole2022dreamfusion}. Variational score distillation (VSD) is later applied in 3D~\cite{wang2024prolificdreamer} and image generation~\cite{yin2024one, luo2024diff, xie2024distillation, salimans2024multistep}. Combining adversarial training with diffusion models is proposed for few-step image generation~\cite{xiao2021tackling, yin2024improved}.
Another branch of methods post alternative restrictions on the diffusion trajectories.
Consistency models \cite{song2023consistency} enabled one-step generation by training models to output consistent results across different noise levels. Latent consistency model (LCM)~\cite{luo2023latent} distills image diffusion models into consistency models. VideoLCM~\cite{wang2023videolcm} and AnimateLCM~\cite{wang2024animatelcm} apply consistency distillation from diffusion video models.  DPM-Solver \cite{lu2022dpmsolver} introduced a fast ODE solver, reducing diffusion sampling to around 10 steps.
Rectified flow~\cite{liu2022flow, lipman2022flow} is a special case of diffusion model, which enforces the straightness of the denoising trajectory during training to achieve high-quality few-step sampling.  Instaflow~\cite{liu2023instaflow} adopts this method to achieve one-step sampling for image generation. However, existing methods have limitations: VSD-based approaches often suffer from model collapse, producing less diverse samples post-distillation, while consistency models tend to yield lower fidelity and samples that are qualitatively poorer than those of the teacher model. Our distillation method, which combines VSD and CD, addresses these issues by generating high-quality and diverse samples.

\paragraph{Reward-based Fine-tuning.} To further improve image and video generation quality in aspects like aesthetic quality and text-image alignment, researchers recently proposed various reward-based fine-tuning methods for diffusion models~\cite{xu2024imagereward,chung2022diffusion, clark2023directly, prabhudesai2024video, li2024t2v, black2023training, clark2023directly, domingo2024adjoint}. The most common ones are direct reward gradients from a differentiable reward model. 
ReFL~\cite{xu2024imagereward} backpropagates the reward gradient through one-step predicted $x_0$ in DMs, similar as diffusion posterior sampling~\cite{chung2022diffusion}. 
DRaFT-$K$~\cite{clark2023directly} truncated the reward gradient backpropagation in diffusion process to latest $K$ steps. VADER~\cite{prabhudesai2024video} applies this on diffusion video models.
T2V-Turbo~\cite{li2024t2v} applies reward gradient for video diffusion models, with the gradients backpropagated through both the reward model and the decoder. The gradient is applied on distilled consistency models for one-step generation, to avoid multi-step backpropagation through DMs. Different from these, denoising diffusion policy optimization (DDPO)~\cite{black2023training} treats the denoising process as decision making process and applies stochastic policy gradient algorithms like REINFORCE and PPO to optimize it, without requiring the differentiable reward function. However, DDPO is found to be less sample efficient as reward-gradient method due to lack of the gradient information~\cite{clark2023directly}. To leverage the rich reward gradient information and bypass backpropagation through the large reward model and the decoder, we propose to use the latent reward model for gradient-based fine-tuning of diffusion models.
Adjoint Matching~\cite{domingo2024adjoint} casts the reward fine-tuning as a stochastic optimal control problem and proposes the memoryless flow matching method to ensure fine-tuned models converge to the tilted distribution.

\section{Methodology}
\label{sec:method}
The overview of our method is shown in Fig.~\ref{fig:diagram}.

\subsection{Diffusion Model}
\label{sec:diffusion}
Suppose the data distribution is $x_0\sim q(x_0)$, the diffusion model approximates this distribution by gradually denoising along a Markov chain. Forward diffusion follows $x_t\coloneqq F(x_0, t)=a_t x_0+b_t\varepsilon, \varepsilon\sim\mathcal{N}(0, \mathbf{I})$. For DDPM,
\begin{align}
    a_t=\sqrt{\bar{\alpha}_t}, b_t=\sqrt{1-\bar{\alpha}_t}
    \label{eq:ddpm_schedule}
\end{align}
with $\bar{\alpha}_t=\Pi_{i=1}^t \alpha_i$ following a pre-specified noise schedule $\alpha_t, t\in[T]$. For the general variance-preserving schedule~\cite{song2021score}, it satisfies $a_t^2+b_t^2=1$, therefore it can be equivalently written as $x_t=\cos(t) x_0 + \sin(t) \varepsilon, t\in[0, \frac{\pi}{2}]$ with a simple mapping of time sequences. Standard diffusion model optimization with velocity prediction $v_\theta$ follows the loss:
\begin{align}
    \mathcal{L}_\text{V}(\theta)&=\mathbb{E}_{x_0\sim q(x_0), \varepsilon\sim\mathcal
    {N}(0, \mathbf{I}), t}\big[w_t||v_\theta(x_t, t)-v_t||_2^2\big]\label{eq:diff_v_loss}\\
    v_t &= -\sin(t)x_0 + \cos(t)\varepsilon
    \label{eq:diff_v}
\end{align}
For rectified flow (RF)~\cite{liu2022flow} or flow matching~\cite{lipman2022flow}, $a_t=1-t, b_t=t, t\in[0,1]$, with a constant velocity target $v_t=\varepsilon-x_0, \forall t\in[0,1]$.

\paragraph{Conjugate Prediction Objective.}
Instead of applying noise prediction in previous work~\cite{ho2020denoising, rombach2022high} and the standard velocity prediction objective as in Instaflow~\cite{liu2023instaflow}, we apply a \textit{conjugate} velocity prediction objective:
\begin{align}
    \mathcal{L}_\text{CV}(\theta)=\mathbb{E}_{x_0\sim q(x_0), \varepsilon\sim\mathcal
    {N}(0, \mathbf{I}), t}\big[||v_\theta(x_t, t)-(x_0 - \varepsilon)||_2^2\big]
    \label{eq:insta_loss}
\end{align}
\begin{wrapfigure}{r}{0.4\columnwidth}
    \centering
\includegraphics[width=0.4\columnwidth]{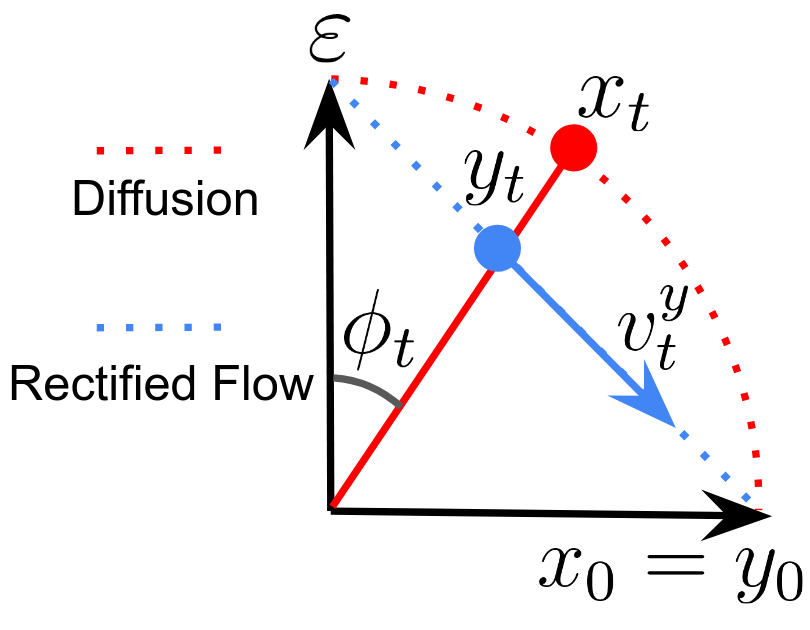}
    \caption{Demonstration of the conjugate velocity prediction: relationship of $v$-prediction for diffusion and rectified flow.
    }
    \label{fig:conjugate}
\end{wrapfigure}
with the sample $x_t$ being diffused along the diffusion trajectory according to the schedule defined as Eq.~\eqref{eq:ddpm_schedule}. The model is parameterized to predict velocity $v_t$ on RF trajectory at each timestep $t$, with a constant target $(x_0-\varepsilon)$ (we take a reverse here as opposed to standard RF for notation clarity), as visualized in Fig.~\ref{fig:conjugate}. The predicted velocity $v_\theta(x_t, t)=v_t^y$ is the velocity on RF as the conjugate point $y_t$ of sample $x_t$ along the diffusion trajectory. This is practically easier to learn compared to the time-varying velocity as in Eq.~\eqref{eq:diff_v}.

\paragraph{Inference.}
After training, the reverse diffusion process follows:
\begin{footnotesize}
\begin{align}
     &x_{t-1}\coloneqq\text{Denoise}(x_t, t, \theta) \nonumber\\
     &=(\sqrt{\bar{\alpha}_{t-1}}-\sqrt{1-\bar{\alpha}_{t-1}-\sigma_t^2}\frac{\sqrt{\bar{\alpha}_t}}{\sqrt{1-\bar{\alpha}_t}})\hat{x}_0 + \frac{\sqrt{1-\bar{\alpha}_{t-1}}}{\sqrt{1-\bar{\alpha}_t}}x_t+\sigma_t \varepsilon
     \label{eq:denoise}
\end{align}
\end{footnotesize}
with $\hat{x}_0=\frac{x_t+\sqrt{1-\bar{\alpha}_t}v_\theta(x_t, t)}{\sqrt{\bar{\alpha}_t}+\sqrt{1-\bar{\alpha}_t}}$ as the predicted original samples. The variance term is $\sigma_t^2=\frac{(1-\alpha_t)(1-\bar{\alpha}_{t-1})}{1-\bar{\alpha}_t}$.
Proofs see Appendix Sec.~\ref{sec:math}.

\subsection{Consistency Distillation}
\label{sec:cd}

Consistency model \cite{song2023consistency} enforces the consistency loss as the distillation method from a pre-trained teacher model $v_{\theta'}$, with a discrete sub-sampled time schedule $t_1=\epsilon< t_2<\dots < t_N=T$:
\begin{small}
\begin{align}
    \mathcal{L}_\text{CD}(\theta)&=\mathbb{E}_{x_0\sim q(x_0), t_n}[\lambda(t_n)d(f_\theta(x_{t_{n+m}}, t_{n+m}),f_{\theta^-}(\hat{x}_{t_n}, t_{n})]\label{eq:cd_loss}\\
    \hat{x}_{t_n}&=\text{Denoise}^m(x_{t_{n+m}}, t_{n+m}, t_n, \theta')
    \label{eq:xt_ode}
\end{align}
\end{small}
where $\lambda(t_n)$ is a time dependent coefficient usually set as constant in practice, and $d(\cdot, \cdot)$ is a distance metric like MSE or Huber loss. Instead of traditional one-step denoising with the teacher model, we apply a generalized CD with $\text{Denoise}^m(\cdot)$ indicating the $m$-step denoising function as defined by Eq.~\eqref{eq:denoise}, which iteratively predicts the sequence $(\hat{x}_{t_{n+m-1}}, \dots, \hat{x}_{t_{n}}|x_{t_{n+m}})$. This is practically found to improve generation quality. The student consistency function $f_\theta$ can be reparameterized from the neural network prediction, similar as in LCM~\cite{luo2023latent}:
\begin{align*}
    f_\theta(x_{t_n},t_n) &= c_{\text{skip}} x_{t_n} + c_{\text{out}}x_\theta(x_{t_n}, t_n)
\end{align*}

There are two different ways for student-teacher parameterization: \textbf{homogeneous} and \textbf{heterogeneous}. 

For homogeneous student-teacher parameterization, the networks of student and teacher both follow the same variable prediction, \emph{i.e.}, $v$-prediction in our setting, with a transformation:
\begin{align}
    x_\theta(x_t,t)&=\frac{x_t+\sqrt{1-\bar{\alpha}_t}v_\theta(x_t, t)}{\sqrt{\bar{\alpha}_t}+\sqrt{1-\bar{\alpha}_t}}
    \label{eq:x_v_trans}
\end{align}
which is proved in Appendix Sec.~\ref{sec:math}.
The student model $v_\theta$ will be initialized from teacher model $v_{\theta'}$ at the beginning of distillation. 

For heterogeneous student-teacher parameterization, the student network can directly predict $x_\theta$ without leveraging Eq.~\eqref{eq:x_v_trans}.
For the best usage of teacher model in student distillation, we adopt the homogeneous parameterization by default.

To enhance the distillation for conditional generation with conditional variable $c\in\mathcal{C}$ (\emph{e.g.}, text prompts), we applied the classifier-free guidance (CFG)~\cite{ho2022classifier} augmentation for the teacher denoising function, similar as VideoLCM~\cite{wang2023videolcm}, but for $v_\theta$-prediction in our case:
\begin{small}
\begin{align}
 v^w_\theta(x_{t_n}, t_n, c)=v_\theta(x_{t_n}, t_n, c) + w\big(v_\theta(x_{t_n}, t_n, c) - v_\theta(x_{t_n}, t_n, \varnothing)\big)
 \label{eq:cfg}
\end{align}
\end{small}
This is applied in replacement of $v_\theta$ in Eq.~\eqref{eq:x_v_trans} for conditional generation.

\subsection{Variational Score Distillation}
\label{sec:vsd}
Variational score distillation (VSD)~\cite{wang2024prolificdreamer} is proposed with the objective of distribution matching between the teacher and student models, by approximating the scores with properly trained diffusion models. Specifically, the loss of minimizing the Kullback-Leibler (KL) divergence between real (teacher) sample distribution $p_\text{real}$ and fake (student) sample distribution $p_\text{fake}$ has the form:
\begin{align}
    \mathcal{L}_\text{VSD}&\coloneqq D_\text{KL}(p_\text{fake}||p_\text{real})=\mathbb{E}_{x\sim p_\text{fake}}[\log \frac{p_\text{fake}(x)}{p_\text{real}(x)}]\\
    &=\mathbb{E}_{\varepsilon\sim \mathcal{N}(0, \mathbf{I}), x=G_\theta(\varepsilon)}[\log \frac{p_\text{fake}(x)}{p_\text{real}(x)}]
    \label{eq:vsd_loss}
\end{align}
and the derivative is,
\begin{small}
\begin{align}
    \nabla_\theta D_\text{KL}=\mathbb{E}_{\varepsilon\sim \mathcal{N}(0, \mathbf{I}), x=G_\theta(\varepsilon)}[-(s_\text{real}(x)-s_\text{fake}(x))\nabla_\theta G_\theta(\varepsilon)]
\end{align}
\end{small}
with score functions $s_\text{real}(x)=\nabla_x \log p_\text{real}(x)$ and $s_\text{fake}(x)=\nabla_x \log p_\text{fake}(x)$ for two distributions. $G_\theta(\cdot)$ is the generation process by the student network through iteratively denoising the noisy training samples. 

The scores are estimated with perturbed samples $x_t, t\in[0.02T, 0.98T]$~\cite{poole2022dreamfusion, yin2024improved} following: 
\begin{align}
   s(x_t, t)=-\frac{\epsilon_\theta(x_t, t)}{\sqrt{1-\bar{\alpha}_t}}=-\frac
{x_t-\sqrt{\bar{\alpha}_t}x_\theta(x_t,t)}{1-\bar{\alpha}_t}
\end{align}
with $x_\theta$ derived by Eq.~\eqref{eq:x_v_trans}.

The real score $s_\text{real}$ is estimated with the pretrained teacher model.
For accurately estimating the fake score $s_\text{fake}$, the fake diffusion model  $v_{\theta_\text{fake}}$ is initialized from the teacher and dynamically adapts according to the student sample distribution. For the score estimation purpose, the fake score model is updated with the diffusion loss $\mathcal{L}_\text{CV}(\theta_\text{fake})$ following Eq.~\eqref{eq:insta_loss}, on student generated samples.

Compared with distribution matching distillation~\cite{yin2024improved}, we abandon the adversarial loss in distillation since the GAN training can be unstable and the improvement can be marginal. We replace the adversarial training with the consistency distillation loss and reward model fine-tuning, which generates richer gradient signals.

\subsection{Latent Reward Fine-tuning}
\begin{figure}[htbp]
    \centering
\includegraphics[width=\columnwidth]{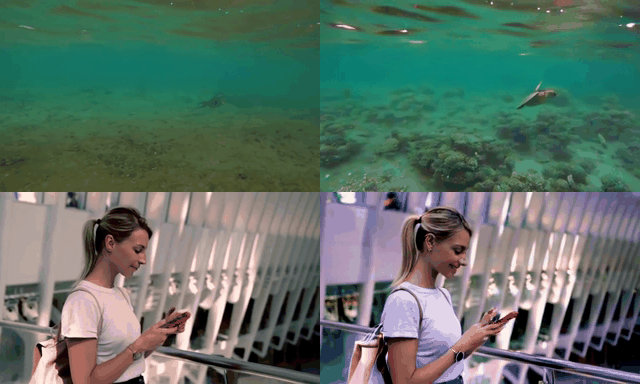}
    \caption{Visualization of samples in training dataset (left) and samples generated with reward tuning using HPSv2 reward (right).
    }
    \label{fig:compare_lrm_hps_x0}
\end{figure}

Reward fine-tuning is an effective approach to align the sample distribution with the specified preference metric in the post-training phase.
As shown in Fig.~\ref{fig:compare_lrm_hps_x0}, the samples generated after reward model tuning can have a substantial difference from the original training samples in dataset (left in the figure), for aspects of aesthetic quality, lighting condition, colors, etc.

\begin{figure}[b]
    \centering
\includegraphics[width=0.98\columnwidth]{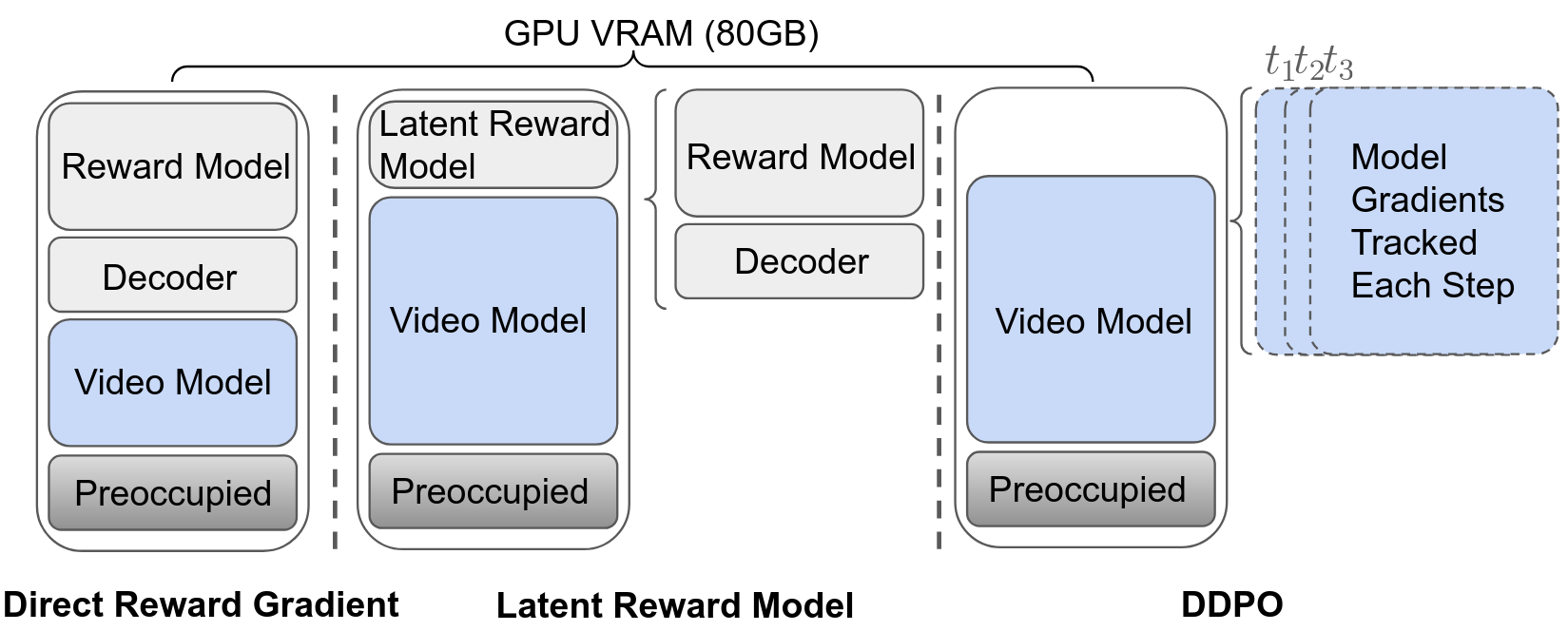}
    \caption{Comparison of different reward fine-tuning methods: 
    (1) Direct reward gradient methods are limited to small reward or video models or short input sequences, and they also require a differentiable reward model; (2) The latent reward model is compact and bypasses the decoder for gradient-based optimization, making it suitable when large reward models and decoders exceed available VRAM; (3) DDPO is similarly constrained by VRAM limits when handling large video models and tracking log-probabilities of samples over multiple steps.
    }
    \label{fig:compare_lrm}
\end{figure}

Previous reward-based optimization methods either (1) requires direct gradients from the pixel-space reward models~\cite{xu2024imagereward, clark2023directly}, or (2) relies on the log-probability estimation of the samples for multiple diffusion steps like DDPO~\cite{black2023training}, as compared in Fig.~\ref{fig:compare_lrm}. One major drawback of (1) is that it only works for differentiable reward function, while not feasible for non-differentiable ones like JPEG compressibility~\cite{black2023training}, etc. Apart from that, the reward models usually work for raw RGB pixel space, which requires the reward gradient to backpropagate through not only the large reward models, but also the decoder, as the practical framework usually follows LDM~\cite{rombach2022high} for latent space modeling. Method (2) is found to be less efficient in reward optimization due to lack of rich reward gradient information~\cite{clark2023directly}, and occupies more memory due to gradient estimation over multiple diffusion steps.

We propose to learn a dual latent reward model (LRM) for directly optimizing the diffusion model in the latent space, which supports any type of reward metrics as detailed in Appendix~\ref{app_sec:lrm_reward_types}. Here we take image rewards as an example. Consider a provided image reward model $\mathcal{R}:\mathcal{I}\rightarrow \mathbb{R}$ with RGB image $i\in\mathcal{I}$ as its input, we approximate the LRM $\mathcal{R}^l_\phi:\mathcal{X}\bigcup \mathcal{X}'\rightarrow \mathbb{R}$ with loss:
\begin{align}
    \mathcal{L}_\text{LRM}(\phi)=\mathbb{E}_{x\in{\mathcal{X}\bigcup \mathcal{X}'}}\big[\big(\mathcal{R}(\text{Dec}(x))- \mathcal{R}^l_\phi(x)\big)^2\big]
    \label{eq:lrm_loss}
\end{align}
where $\mathcal{X}'=\{G_\theta(\varepsilon)\}$ is the set of generated samples from the generator, and $\text{Dec}(\cdot)$ is the decoder. We use both real images and generated images to improve the robustness of learned LRM on generated samples. To alleviate the computational burden in training the LRM, we apply Eq.~\eqref{eq:x_v_trans} for single-step prediction of generated samples, rather than iteratively denoising along the entire trajectory. This approach significantly reduces memory usage by avoiding gradient backpropagation through the iterative sampling process. Although our distilled student models operate with a maximum of 4 sampling steps, memory usage can still be intensive if samples are generated with a full denoising process. Tab.~\ref{tab:model_comparison_memory} compares the parameter counts and memory costs for pixel-space reward models and LRMs on video samples with a batch size of 1. HPSv2 and PickScore are two pixel-space reward models used in our experiments (as described in Sec.~\ref{sec:reward_tune}).

\begin{table}[htbp]
\centering
\caption{Comparison of parameters and GPU memory (VRAM) costs and for pixel-space HPSv2, PickScore reward models and LRMs. }
\resizebox{\columnwidth}{!}{
\begin{tabular}{@{}l|c|c|c@{}}
\toprule
Model                     & \# Parameters   & Forward VRAM   & Backward VRAM   \\ \midrule
Image LRM       & 189,441            & 8.998 MB                 & 17.772 MB                     \\
Text-image LRM & 763,009            & 15.500 MB                & 26.277 MB                          \\
HPSv2/PickScore    & 632 million        & 5.926 GB                & $>$90 GB                             \\ \bottomrule
\end{tabular}
}
\label{tab:model_comparison_memory}
\end{table}

With the compact and differentiable LRM on the latent space, we apply direct reward gradient optimization to fine-tune the diffusion model:
\begin{align}
    \mathcal{L}_\text{FT}(\theta; \phi) = -\mathbb{E}_{\varepsilon\sim \mathcal{N}(0, \mathbf{I})}[\mathcal{R}^l_\phi(G_\theta(\varepsilon))]
    \label{eq:lrm_ft_loss}
\end{align}

In practice, we can either pre-train the LRM first and then fine-tune the diffusion model with a fixed LRM, or train the LRM and fine-tune the diffusion model iteratively. For simplicity, we adopt the second approach. If the original reward model is conditioned on additional text input, $\mathcal{R}(i, c)$, the LRM also operates conditionally as $\mathcal{R}^l(x, c), c \in \mathcal{C}$. \textit{The LRM method can accommodate any type of reward models, including image, video, text-image, and text-video rewards.} For image-only LRMs, we use a convolutional neural network, while for text-image LRMs, we apply a cross-attention module after the convolutional feature extractor to integrate text embeddings with image features. For video-based LRMs, the 2D convolution is replaced with a 3D convolutional neural network. Additional details are provided in Appendix~\ref{app_sec:lrm_reward_types}.

\subsection{Multi-Objective Distillation}
Distillation using VSD alone can result in severe mode collapse, while CD tends to produce lower-quality samples by averaging across sample distributions (Appendix Fig.~\ref{fig:infer_steps2}). By integrating consistency distillation, variational score distillation, and latent reward fine-tuning objectives, our method enables few-step generation of high-quality, diverse samples after distillation, optimized by the following loss function:
\begin{align}
    \mathcal{L}(\theta) = \mathcal{L}_\text{VSD}(\theta) + \beta_\text{CD} \mathcal{L}_\text{CD}(\theta) + \beta_\text{FT} \mathcal{L}_\text{FT}(\theta;\phi)
\end{align}
During distillation, the fake score network is updated with $\mathcal{L}_\text{CV}(\theta_\text{fake})$, and the LRM $\mathcal{R}^l_\phi$ is updated using $\mathcal{L}_\text{LRM}(\phi)$. 
The pseudo-code of our method is displayed as Alg.~\ref{alg:code}
\begin{algorithm}[H]
\caption{Training procedure of \texttt{DOLLAR}.}
\small
\begin{algorithmic}[1]
\State \textbf{Input:} Pretrained teacher model $v_{\theta'}$ by $\mathcal{L}_\text{CV}$ Eq.~\eqref{eq:insta_loss}, pretrained encoder and decoder, dataset $\mathcal{D}=\{(c, i)\}$
\State \textbf{Output:} Distilled student few-step generator $G_\theta$.
\State \texttt{//Initialize student and fake score model from teacher}
\State $\theta\leftarrow \theta'$, $\theta_\text{fake}\leftarrow \theta'$
\While{\text{train}}
\State Sample batch $(c,i)\sim\mathcal{D}$, encode $x\leftarrow \text{Encoder}(i)$
\State \texttt{//Update the generator with distillation}
\State $\hat{x}\leftarrow G_\theta(c, \varepsilon), \varepsilon\sim\mathcal{N}(0, \mathbf{I})$
\State Uniformly sample $t_n$, forward diffusion $x_{t_{n+m}}\leftarrow F(x, t_{n+m})$
\State $\mathcal{L}_\text{G}=\mathcal{L}_\text{VSD}(\theta;\theta', \theta_\text{fake}, \hat{x}, c)+\eta_1\mathcal{L}_\text{CD}(\theta;\theta', x_{t_{n+m}}, c)$ \texttt{//VSD by Eq.~\eqref{eq:vsd_loss}, CD by Eq.~\eqref{eq:cd_loss}}
\State $G_\theta\leftarrow \text{GradientDescent}(\theta, \mathcal{L}_\text{G})$
\State \texttt{//Update fake score model}
\State Uniformly sample $t$, forward diffusion $x_t\leftarrow F(\hat{x}, t)$
\State $\theta_\text{fake}\leftarrow \text{GradientDescent}(\theta_\text{fake}, \mathcal{L}_\text{CV}(x_t))$ \texttt{//Eq.~\eqref{eq:insta_loss}}
\State \texttt{//Train latent reward model}
\State Merge batch $\tilde{x}=x\bigcup \hat{x}$ 
\State $\mathcal{R}^l_\phi\leftarrow \text{GradientDescent}(\phi, \mathcal{L}_\text{LRM}(\phi; \tilde{x}, \mathcal{R}))$ \texttt{//Eq.~\eqref{eq:lrm_loss}}
\State \texttt{//Update the generator with latent reward fine-tuning}
\State $G_\theta\leftarrow \text{GradientDescent}(\theta, \mathcal{L}_\text{FT}(\theta; \hat{x}, \mathcal{R}^l))$ \texttt{//Eq.~\eqref{eq:lrm_ft_loss}}
\EndWhile
\end{algorithmic}
\label{alg:code}
\end{algorithm}

\section{Experiments}

\subsection{Implementation}
\label{sec:implement}
\paragraph{Student and teacher models.}
The video diffusion model in our experiments is based on the bidirectional diffusion transformer (DiT) architecture~\cite{peebles2023scalable}, similar as CogVideoX~\cite{yang2024cogvideox}. Although our methodology is architecture-agnostic and could be applied to models like U-Net~\cite{rombach2022high}, we select the transformer due to its scalability. The teacher and student T2V diffusion models are same as a modified variant of CogVideoX~\cite{yang2024cogvideox} and follow the LDM framework~\cite{rombach2022high, blattmann2023align}, utilizing DiT modeling in the latent space encoded with a pretrained 3D variational autoencoder (VAE)~\cite{kingma2013auto, opensora}. The 3D VAE encodes and decodes videos chunk-by-chunk to alleviate the computational burden, encoding chunks of 16 video frames into 5 latent embeddings. These embeddings are then patchified into sequences as inputs to the DiT. Leveraging the DiT architecture, the model can accommodate arbitrary video durations and resolutions; however, our experiments constrain the video generation setting to 128 frames at a resolution of $192 \times 320$.

\paragraph{Model training and inference.}
Following the setting of CogVideoX~\cite{yang2024cogvideox}, the student model is distilled with a mixture of internal image and video datasets with text captioning. All videos are resized and cropped with same resolution as $192 \times 320$ and the student distillation uses around 320K licensed single-shot videos. The teacher model employs standard DDPM settings with 1000 sampling steps: $t\in[1,\dots,1000]$. For inference, the teacher model utilizes DDIM sampling to generate high-quality samples in 50 steps, with $t_n\in[19, 39, \dots, 999]$. After distillation, the student model adopts a default 4-step sampling protocol, as in previous work~\cite{yin2024improved}, using timesteps $[249, 499, 749, 999]$. Additionally, we explore 1-step ($[999]$) and 2-step ($[499, 999]$) generation configurations for the student model in Sec.~\ref{sec:exp_ablation}. Consistency distillation (CD, discussed in Sec.~\ref{sec:cd}) follows a DDIM schedule with $N=50$ steps, as implemented in LCM~\cite{luo2023latent}. For teacher inference, we apply classifier-free guidance (CFG)~\cite{ho2022classifier} augmentation with a weight of $w=7.5$ in CD as specified in Eq.~\eqref{eq:cfg} and $w=3.5$ for the real score network in VSD, The fake score network and distilled student inference do not employ CFG. In the VSD loss, we adhere to the update ratio as 5 for the fake score update over generator update, as suggested in previous work~\cite{yin2024improved}, to ensure training stability. All experiments are conducted with a batch size of 1 per GPU due to the large model size and limited VRAM, utilizing 8 A100 GPUs in parallel for each run. All student models are distilled up to $4\times 10^4$ iterations using AdamW~\cite{loshchilov2017decoupled} optimizer and a learning rate of $2\times 10^{-5}$, with moderate model selection. Video samples are generated with 128 frames at a resolution of $192\times 320$. We set $\beta_\text{CD}=0.5$ and $\beta_\text{FT}=1.0$. 

To reduce VRAM occupancy on GPUs, we employ gradient checkpointing and fully sharded data parallel (FSDP)~\cite{zhao2023pytorch}, enabling sharding of model weights and gradients across GPUs in a data-parallel fashion. Additionally, we utilize mixed precision training with the Bfloat16 data type. For fine-tuning with LRMs, we apply gradient accumulation over 7 steps to stabilize training due to the small batch size (=1) used.

\paragraph{Reward Metrics.}
We utilize Human Preference Score v2 (HPSv2)~\cite{wu2023human} and PickScore~\cite{kirstain2023pick} as the text-image reward models. Both are fine-tuned CLIP-type models trained on extensive text-to-image datasets with human preferences. While our methods are compatible with directly optimizing the model using VBench reward metrics, we intentionally avoid doing so, as VBench scores serve as one of the final evaluation criteria. However, in Sec.~\ref{sec:exp_ablation}, we present ablation experiments where we optimize using VBench metrics, such as dynamics degree, and other image rewards, like JPEG compressibility. These experiments reveal that while reward scores can be significantly improved, it could result in overoptimization for specific metrics, leading to a degradation in overall generation quality. Consequently, we adopt the more general preference-based reward models, HPSv2 and PickScore, by default for reward fine-tuning. Nonetheless, our method remains compatible with other reward models.

\paragraph{Evaluation.}
To faithfully reflect model performance, we apply both automatic evaluation benchmark VBench~\cite{huang2024vbench} and human evaluation for our results. VBench assesses 16 dimensions encompassing both video visual quality and semantic alignment aspects for T2V models, with higher scores indicating better performance in each metric. Following the standard VBench evaluation protocol, we use a set of 946 long prompts, generating five videos per prompt. Final scores for each dimension are averaged across all generated videos for that metric.

Additionally, we examine the impact of prompt length, comparing the performance of long descriptive prompts with short prompts in VBench evaluation (details in Sec.~\ref{sec:exp_ablation}). To further assess text-video alignment capabilities, we sample the distilled student models with various styles and motions, with the results provided in Appendix Sec.~\ref{app_sec:style_motion}.

\begin{table*}[htbp]
\centering
\caption{Comparison of VBench scores for different models.}
\resizebox{0.95\textwidth}{!}{%
\begin{tabular}{@{}cccccp{1.6cm}|cp{1.5cm}p{1.5cm}@{}}
\toprule
Model & Pika & Gen-2 & Gen-3 & Kling & T2V-Turbo (VC2) & Teacher & \texttt{DOLLAR} (PickScore) & \texttt{DOLLAR} (HPSv2) \\ \midrule
Subject Consistency & \underline{96.76} & \underline{97.61} & 97.10 & \textbf{98.33} & 96.28 & 83.99 & 93.77 & 92.57 \\
Background Consistency & \textbf{98.95} & \underline{97.61} & 96.62 & \underline{97.60} & 97.02 & 93.78 & 96.80 & 96.14 \\
Temporal Flickering & \textbf{99.77} & \underline{99.56} & 98.61 & \underline{99.30} & 97.48 & 96.42 & 96.30 & 97.48 \\
Motion Smoothness & \underline{99.51} & \textbf{99.58} & 99.23 & \underline{99.40} & 97.34 & 98.09 & 97.76 & 98.59 \\
Dynamic Degree& 37.22 & 18.89 & 60.14 & 61.21 & 49.17 & \textbf{99.44} & \underline{75.83} & \underline{81.67} \\
Aesthetic Quality & 63.15 & \textbf{66.96} & \underline{63.34} & 46.94 & 63.04 & 61.21 & \underline{63.80} & 63.14 \\
Imaging Quality& 62.33 & \underline{67.42} & 66.82 & 65.62 & \textbf{72.49} & 63.87 & \underline{69.40} & 65.61 \\
Object Class& 87.45 & 90.92 & 87.81 & 87.24 & \textbf{93.96} & 85.79 & \underline{91.63} & \underline{93.84} \\
Multiple Objects & 46.69 & 55.47 & 53.64 & \underline{68.05} & 54.65 & 52.59 & \underline{69.71} & \textbf{72.21} \\
Human Action& 88.00 & 89.20 & 96.40 & 93.40 & 95.20 & \textbf{99.60} & \underline{99.00} & \underline{99.00} \\
Color & 85.31 & \underline{89.49} & 80.90 & \textbf{89.90} & \textbf{89.90} & 77.00 & 77.95 & 74.78 \\
Spatial Relationship& 65.65 & 66.91 & 65.09 & \textbf{73.03} & 38.67 & 51.40 & \underline{68.56} & \underline{68.35} \\
Scene & 44.80 & 48.91 & \underline{54.57} & 50.86 & \textbf{55.58} & 49.99 & \underline{55.06} & 52.72 \\
Temporal Style & 24.44 & 24.12 & 24.71 & 24.17 & \underline{25.51} & \textbf{26.45} & 24.64 & \underline{25.23} \\
Appearance Style & 21.89 & 24.31 & \textbf{24.86} & 19.62 & 24.42 & \underline{24.83} & \underline{24.45} & 23.50 \\
Overall Consistency & 25.47 & 26.17 & 26.69 & 26.42 & \textbf{28.16} & \underline{27.89} & \underline{26.93} & 26.85 \\
\midrule
Quality Score & 82.68 & 82.47 & \textbf{84.11} & 83.39 & 82.57 & 81.89 & \underline{83.49} & \underline{83.83} \\
Semantic Score & 71.26 & 73.03 & 75.17 & \underline{75.68} & 72.57 & 73.71 & \textbf{77.90} & \underline{77.51} \\
\textbf{Total Score} & 80.40 & 80.58 & \underline{82.32} & 81.85 & 81.01 & 80.25 & \underline{82.37} & \textbf{82.57} \\
\bottomrule
\end{tabular}%
}
\label{tab:compare_vbench}
\end{table*}

\subsection{Comparison with VBench Baselines}
\paragraph{Vbench Results.}
The VBench evaluation results are summarized in Tab.~\ref{tab:compare_vbench}, with the highest in bold and 2nd and 3rd underlined. Our distilled methods with VSD+CD+LRM achieve superior performance over the baselines including Pika~\cite{pikalabs2023}, Gen-2, Gen-3, Kling~\cite{kuaishou2024}, T2V-Turbo~\cite{li2024t2v}, and our teacher model. The highest semantic scores of our models indicate a significant improvement over baselines for text-video alignment. The quality score, which reflects the visual quality, is heavily affected by the frame consistency metrics like subject consistency, background consistency, temporal flickering and motion smoothness, which are usually high if there is a lack of motions in the videos. Our models have significantly higher dynamics degree for motions as shown in the table as well as visualization in Appendix.~\ref{app_sec:visual_methods}. The total score is a weighted sum of all metrics showing the general preference of the videos, and our method achieves 82.37 and 82.57 surpassing all models in the table, as well as outperforming the teacher model. The students achieve higher scores in 9-10 metrics (out of 16) than the teacher. It indicates that the performance of our method is not upper bounded by the teacher model, which is beyond the VSD loss for student and teacher distribution matching. The additional CD loss enforces the self-consistency of model prediction on noisy real images. It provides the source of signals to improve the student model over teacher model on quality and semantic performances, which are further boosted by LRM fine-tuning.

\paragraph{Human Evaluation.}
\begin{figure*}[htbp]
    \centering
\includegraphics[width=0.99\textwidth]{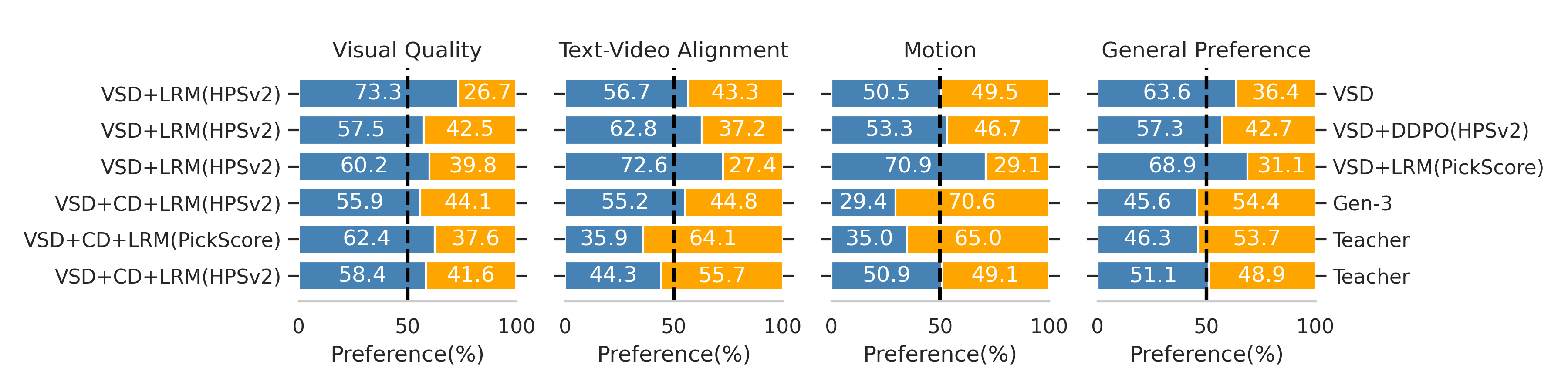}
    \caption{Human evaluation results over four independent metrics: visual quality, text-video alignment, motion and general preference.
    }
    \label{fig:human_eval}
\end{figure*}

We further conduct human evaluation to visually compare the generated videos for different models, over four independent metrics: visual quality, text-video alignment, motion, and general preference. The evaluation details are provided in Appendix Sec.~\ref{app_sec:human_eval}. From the evaluated results in Fig.~\ref{fig:human_eval}, our method with HPSv2 reward is preferred more than the DDPO method (by $57.3\%$) and teacher model (by $51.1\%$), and performs similarly with the Gen-3 model (by $45.6\%$) in  terms of general preference. The visual quality of our distilled students is significantly higher than both teacher (by $58.4\%$) and Gen-3 (by $55.9\%$). Moreover, we find that, PickScore increases visual quality, but likely leads to worse motion performance. HPSv2 for reward tuning not only increases the visual quality, but has better motion and text-video alignment.

\subsection{Comparison of Distillation Methods}
\label{sec:distill_compare}
Tab.~\ref{tab:diversity} shows the ablation of our distillation method by comparing it with VSD and VSD+CD. The breakdown results for each VBench metric refer to Appendix Sec.~\ref{app_sec:complete_vbench}. VSD has comparable performances with teacher, with additional CD loss it increases the sample diversity. Our VSD+CD+LRM method achieves high sample quality and diversity overall. We provide visualization of samples in Appendix Sec.~\ref{app_sec:visual}.

\paragraph{Diversity Measure.}
The diversity of model generation is not captured by the VBench. We conduct both qualitative and quantitative comparison of generation diversity for different distillation methods. We quantitatively measure the diversity of sampled videos with Vendi score~\cite{friedman2022vendi}, which is based on the similarity matrix for the sample set. The mean and standard deviations across prompted video samples are reported in Tab.~\ref{tab:diversity}. The videos are generated using VBench long prompts, with five videos produced for each prompt. To evaluate diversity, we randomly sample 500 prompts, resulting in a total of 2,500 videos. For assessing video sample diversity within a single prompt, we define the diversity metric as:
\begin{align}
   \text{Diversity}=\frac{1}{K}\sum_{k=1}^K \text{Vendi}\big([f^k_1, \dots, f^k_n]\big)
\end{align}
For images, the function $\text{Vendi}(\cdot)$ quantifies the diversity of a set of image features, which can be derived either from raw pixel vectors or embeddings obtained via the Torchvision Inception v3 model. For videos, we uniformly extract $K$ keyframes with an equal spacing of 20 frames between consecutive keyframes. We then calculate $\text{Vendi}\big([f^k_1, \dots, f^k_n]\big)$ for the 5 videos corresponding to the same frame index $k$. Finally, the diversity measure for a given prompt is obtained by averaging the Vendi values across all $K$ frames. The mean and standard deviations of this metric are computed and reported across all prompts to evaluate video diversity, as shown in Tab.~\ref{tab:diversity}.
Visual samples are provided in Appendix Sec.~\ref{app_sec:visual_diverse}.
We find that the Inception-based Vendi score aligns better with visual inspection than pixel-based alternatives. While VSD produces high-quality samples, it tends to lead to mode collapse. Our method addresses this diversity limitation by incorporating CD loss, and further enhances generation quality through LRM fine-tuning.

\begin{table}[htbp]
\centering
\caption{Comparison of teacher and students with different distillation methods, with 4-step sampling for student models.}
\resizebox{\columnwidth}{!}{
\begin{tabular}{p{2.7cm}|c|ccc@{}}
\toprule
Model & Teacher & \multicolumn{3}{|c}{Student} \\ \midrule
Method & DDIM 50 steps & VSD & VSD+CD & VSD+CD+LRM \\ \midrule
Quality Score & 81.89 & 80.95 & 82.16 & \textbf{83.83} \\ \hline
Semantic Score & 73.71 & 76.61 & 74.58 & \textbf{77.51} \\ \hline
Total Score & 80.25 & 80.08 & 80.65 & \textbf{82.57} \\ \hline
Vendi (Pixel)$\uparrow$ & $1.46\pm0.14$ & $1.49\pm0.14$  & $1.59\pm0.17$ &  $\mathbf{1.60\pm0.14}$ \\ \hline
Vendi (Inception)$\uparrow$ & $\mathbf{2.34\pm0.16}$ & $1.91\pm0.14$  & $2.14\pm0.15$ & $1.98\pm0.14$\\ 
\bottomrule
\end{tabular}
}
\label{tab:diversity}
\end{table}

\paragraph{Inference Time.}
Tab.~\ref{tab:inference_time_percentage} presents the per-sample inference time consumption for the teacher model using 50-step DDIM inference, and for student models with 1, 2, and 4 inference steps. Here, ``diffusion time" refers solely to the diffusion sampling in latent space, while ``inference time" encompasses the complete generation process for one video, including text encoding, diffusion sampling, and decoding of latent outputs.
The inference experiments are conducted on a single A100 80GB GPU, with mean and standard deviation calculated over 100 samples. With parallel sampling across multiple GPUs, the amortized time per sample can be further minimized. The reported values indicate the percentage of the teacher model's inference time, excluding amortization effects. Distilled student models significantly accelerate diffusion sampling compared to the teacher, achieving speedups from $\times 15.6$ (4 steps) to $\times 278.6$ (1 step). Absolute time costs are not reported, as they are influenced by hardware-specific factors and inference configurations such as batch size and the number of GPUs used. Instead, relative time consumption is emphasized as a more reliable metric for cross-configuration comparisons.

Notably, the relationship between diffusion sampling time and the number of sampling steps is not strictly linear. For example, the first diffusion sampling step accounts for only $0.33\%$ of the total inference time, making it approximately 6.2 times faster than subsequent steps. This discrepancy is likely due to the faster inference process for initial Gaussian noise inputs or the relatively low hardware cache occupation during early inference stages.

Furthermore, the difference between the total inference time and the diffusion sampling time includes additional costs for text preprocessing and encoding, as well as decoding from the latent space back to the original pixel space. These processes collectively account for approximately $7\%$ of the total inference time.

\begin{table}[htbp]
\centering
\caption{Time consumption of teacher and distilled student models (as percentage of teacher's total inference time) with different numbers of function evaluations (NFE)}
\resizebox{\columnwidth}{!}{
\begin{tabular}{@{}l|c|ccc@{}}
\toprule
Model & Teacher & \multicolumn{3}{c}{Student} \\ \midrule
Steps (NFE) & 50  & 4 & 2 & 1 \\ \hline
Diffusion Time (\%) & $91.94\pm0.32$ & $5.88\pm0.03$ & $2.16\pm0.01$ & $0.33\pm0.02$ \\ \hline
Inference Time (\%) & $100.00\pm0.66$ & $13.06\pm0.17$ & $9.30\pm0.11$ & $7.45\pm0.12$ \\
\bottomrule
\end{tabular}
}
\label{tab:inference_time_percentage}
\end{table}

\subsection{Reward Fine-Tuning}
\label{sec:reward_tune}
\begin{figure}[htbp]
    \centering
\includegraphics[width=\columnwidth]{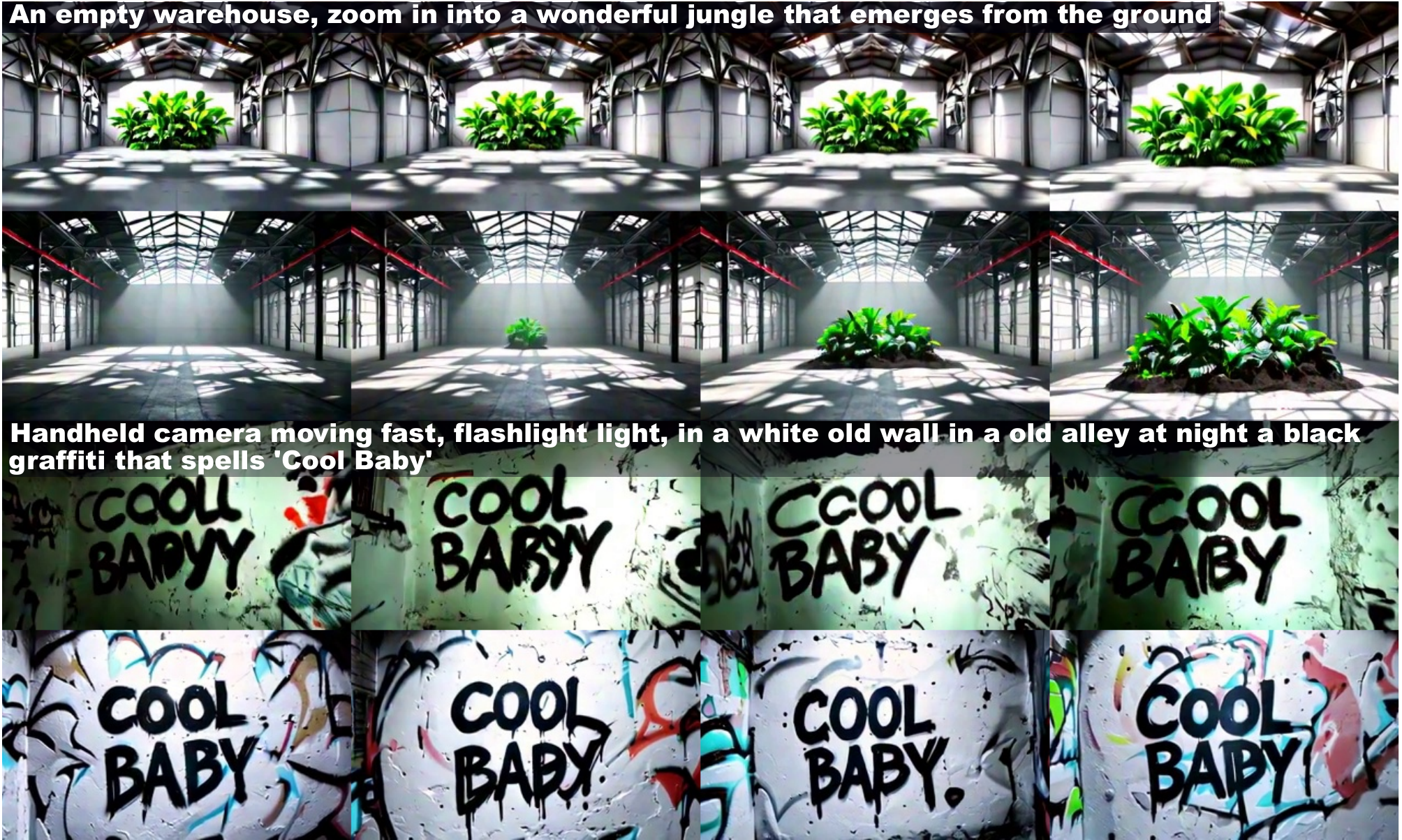}
    \caption{Compare the generated samples with (first line) and without (second line) reward fine-tuning for two samples. First sample: 4 frames are extracted from one sampled video per method along the time sequence. Second sample: one frame is extracted from one video, with 4 videos sampled by the same prompt.
    }
    \label{fig:compare_hps}
\end{figure}

\paragraph{Baselines.}
DDPO~\cite{black2023training} applies the REINFORCE algorithm to optimize the diffusion model by treating the diffusion process as a MDP. It requires to estimate the log-probabilities for the sample at all diffusion steps, which are then summed over and weighted by the final reward as the optimization objective. Considering memory constraints, our method is suited for few-step sampling models or configurations with gradient truncation along the diffusion trajectory. In our experiments, memory limitations prevent log-probability estimation over more than 2 steps. Therefore, we employ a truncation step of 2 for the student model (\emph{i.e.}, log-probability estimation at timesteps [249, 499]). This truncation approach has been validated in previous work~\cite{clark2023directly, ren2024diffusion}. We apply $\text{DDPO}_\text{SF}$ for online policy gradient. More details refer to Appendix Sec.~\ref{sec:ddpo}. Direct reward gradient methods like ReFL~\cite{xu2024imagereward} and DRaFT~\cite{clark2023directly} exceed single-GPU memory capacity in our case, thus are not included as baselines.

\begin{figure}[htbp]
    \centering
\includegraphics[width=0.99\columnwidth]{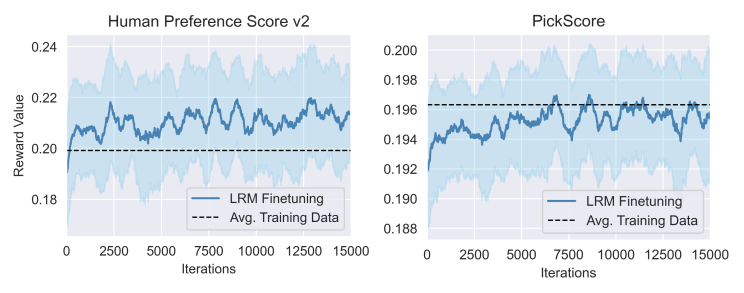}
    \caption{Latent reward model fine-tuning process under reward metrics HPSv2 and PickScore.
    }
    \label{fig:hpsv2}
\end{figure}

\paragraph{Method Comparison.} Tab.~\ref{tab:reward_method_compare} compares VBench scores and final reward values for LRM and DDPO. The last row ``Reward'' indicates the corresponding reward value after fine-tuning, for example, PickScore value is reported if the model is fine-tuned with PickScore reward model, and similar for HPSv2. The mean and standard deviation values are reported with 500 videos generated under VBench prompts. As visualized in Fig.~\ref{fig:compare_hps}, the reward fine-tuning helps to improve the text-image alignment for the first prompt by more explicitly exhibiting the ``emerging'' effect, and improves the accuracy of text display in frames for the second prompt. The lighting style is also improved through fine-tuning. Fig.~\ref{fig:hpsv2} displays the predicted reward values $\mathcal{R}^l_\phi(\hat{x}_0, c)$ with LRM for generated samples ($\hat{x}_0\sim\mathcal{X}'$, by Eq.~\eqref{eq:x_v_trans}) during the distillation process with VSD+LRM loss, for two reward metrics HPSv2 and PickScore, respectively. The horizontal dashed lines are the average reward values of the samples in training dataset. For HPSv2, the reward values of generated samples surpass the training data quickly with the LRM fine-tuning. For PickScore, the reward values of generated samples also gradually increase to be close to the training data.

Full results for VBench scores refer to Appendix Tab.~\ref{tab:vbench_lrm_ddpo}.
Human evaluation results of the LRM and DDPO refer to Appendix~\ref{app_sec:human_eval}.

\begin{table}[htbp]
\centering
\caption{Comparison of LRM with DDPO using VBench and training reward metrics. \texttt{DOLLAR} is our final method with VSD+CD+LRM.}
\resizebox{\columnwidth}{!}{
\begin{tabular}{@{}l|ccc|ccc@{}}
\toprule
Reward Model & \multicolumn{3}{c|}{PickScore} & \multicolumn{3}{c}{HPSv2} \\ \midrule
Method & VSD+DDPO & VSD+LRM & \texttt{DOLLAR} &  VSD+DDPO & VSD+LRM & \texttt{DOLLAR}   \\ \midrule
Quality & 82.99 & \textbf{84.01} & 83.49  & 82.97 & 83.53 & 83.83\\ \hline
Semantic & 77.26 & 72.51 & \textbf{77.90}  & 74.56 & 75.67 & 77.51\\ \hline
Total Score & 81.84  & 81.71 & 82.37  & 81.29 & 81.96 & \textbf{82.57}\\ \hline
Reward & 0.207 & 0.207 & \textbf{0.210 }& 0.271 & 0.276 & \textbf{0.277} \\
\bottomrule
\end{tabular}
}
\label{tab:reward_method_compare}
\end{table}

\subsection{Ablation Studies}
\label{sec:exp_ablation}
\paragraph{Distillation Timesteps.}
Our proposed method supports an arbitrary subset of timesteps for teacher sampling. By default, we use 4-step sampling for the student model to balance quality and efficiency, as discussed in Sec.~\ref{sec:implement}. Here, we investigate the impact of varying the number of sampling steps during distillation, specifically testing 1 step (timestep [999]), 2 steps (timesteps [499, 999]), and 4 steps (timesteps [249, 499, 749, 999]) with equal spacing. While our approach does not require equal spacing, this configuration is used for consistency in this experiment.
The evaluated VBench scores are reported in Tab.~\ref{tab:infer_steps}.
All three distilled student models with VSD loss demonstrate comparable or even superior performances relative to the teacher model with 50 inference steps. The slight differences can be attributed to checkpoint selection and evaluation variance. From visual inspection and human evaluation, we find that models with more inference steps tend to perform better, which may not be fully captured by the minor differences in VBench scores. Sample visualizations are provided in Appendix Sec.~\ref{app_sec:infer_steps}.
The breakdown results for each VBench metric refer to Appendix.

\begin{table}[htbp]
\centering
\caption{Comparison of the number of inference steps for distilled students with VSD using VBench (long prompt).}
\begin{tabular}{@{}l|cccc@{}}
\toprule
Model & Teacher & \multicolumn{3}{c}{Student (VSD)} \\ \midrule
Inference Steps & $50$ & $1$ & $2$ & $4$ \\ \midrule
Quality Score & 81.89 & 81.61  & 82.71 & 80.95 \\ \hline
Semantic Score & 73.71 & 76.66  & 73.86 & 76.61 \\ \hline
Total Score & 80.25  & 80.62  & 80.94 & 80.08 \\
\bottomrule
\end{tabular}
\label{tab:infer_steps}
\end{table}

\paragraph{Consistency Distillation Denoising Steps.}
In Sec.~\ref{sec:cd}, we introduced the consistency distillation method with a multi-step teacher denoising function: Denoise$^m(\cdot)$. We ablate the choice of $m$ in experiments and find that a larger value like $m=5$ improves distillation performance, as detailed in Tab.~\ref{tab:cd_steps}. The student models follow 4-step schedule, and CD loss is applied on a 50-step DDIM schedule with step size $20$ as previously discussed. Full results for VBench scores see Appendix Tab.~\ref{tab:vbench_vsd_cd}.

\begin{table}[htbp]
\centering
\caption{Effect of the number of teacher denoising steps in consistency distillation (CD), using VBench scores (with long prompt).}
\begin{tabular}{@{}l|ccc@{}}
\toprule
Model & Teacher & \multicolumn{2}{c}{Student (VSD+CD)} \\ \midrule
CD with Denoise$^m$ & - & $m=1$ & $m=5$ \\ \midrule
Quality Score & 81.89 & 80.75  & 82.16  \\ \hline
Semantic Score & 73.71 & 71.57  & 74.58 \\ \hline
Total Score & 80.25  & 78.92  & 80.65 \\
\bottomrule
\end{tabular}
\label{tab:cd_steps}
\end{table}

\paragraph{Homogeneous vs. Heterogeneous Parameterization.}
For the given teacher model $v_{\theta'}$ with $v$-prediction, we compare the student models with heterogeneous $x_\theta$ and homogeneous $v_\theta$ parameterization from the teacher, under the VSD+CD loss. The student model weights are initialized from the teacher model for both configurations. The evaluated VBench results are shown in Tab.~\ref{tab:parameterization}. The homogeneous parameterization leads to slightly better performance over the heterogeneous parameterization and even the teacher model.
Full results for VBench scores refer to Appendix Tab.~\ref{tab:vbench_vsd_cd}.

\begin{table}[htbp]
\centering
\caption{Comparison of different student-teacher 
parameterization for distillation with VSD+CD using VBench (long prompt).}
\resizebox{\columnwidth}{!}{
\begin{tabular}{@{}l|c|cc@{}}
\toprule
Model & Teacher & \multicolumn{2}{|c}{Student} \\ \midrule
Parameterization & $v_\theta$ & Heterogeneous ($x_\theta$) & Homogeneous ($v_\theta$) \\ \midrule
Quality Score & 81.89 & 81.65  & 82.16  \\ \hline
Semantic Score & 73.71 & 73.66  & 74.58 \\ \hline
Total Score & 80.25  & 80.05  & 80.65 \\
\bottomrule
\end{tabular}
}
\label{tab:parameterization}
\end{table}

\paragraph{Vbench Prompt Length.}
During our evaluation, we observed that the standard prompt suite in VBench includes very short prompts, such as ``a bus," which lack context or motion descriptions. This does not align well with the text-video data distribution used to train our model, where most images and videos are accompanied by richly detailed captions to enhance the model's semantic capabilities.
Our findings indicate that pretrained T2V models often exhibit a bias toward prompt length, performing better with longer, more descriptive prompts. To address this, VBench incorporates the prompt optimization technique introduced in CogVideoX~\cite{yang2024cogvideox}, which utilizes GPT-4o~\cite{achiam2023gpt} to extend the short prompts into more descriptive ``long prompts" while preserving their original meanings. We refer to these as ``long prompts", distinguishing them from the original ``short prompts".

The VBench score comparison for long prompts and short prompts are summarized in Tab.~\ref{tab:prompt_length}. The evaluation includes five models:
\begin{itemize}
    \item Teacher model 
    \item VSD model with 1-step inference (VSD1)
    \item VSD model with 4-step inference (VSD4)
    \item Model distilled with VSD and CD joint loss, using CD denoising step $m=1$ (VSD4+CD1)
    \item Model distilled with VSD and LRM joint loss, using PickScore as reward function (VSD4+LRM)
\end{itemize}
\begin{table}[htbp]
\centering
\caption{Effects on VBench scores by different prompt lengths: ``S'' for short prompt and ``L'' for long prompt.}
\resizebox{\columnwidth}{!}{
\begin{tabular}{@{}l|cccccccccc@{}}
\toprule
Model & \multicolumn{2}{c}{Teacher} & \multicolumn{2}{c}{VSD1} & \multicolumn{2}{c}{VSD4} & \multicolumn{2}{c}{VSD4+CD1} & \multicolumn{2}{c}{VSD4+LRM}\\ \midrule
Prompt & S & L & S & L & S & L & S & L & S & L \\ \midrule
Quality & 81.50 & 81.89  & 81.60 & 81.61 & 80.75 & 80.95 & 79.27 & 80.75 & 82.64 & 84.01\\ \hline
Semantic & 74.64 & 73.71  & 77.10 & 76.66 & 76.67 & 76.61 & 67.52 & 71.57 & 60.04 & 72.51\\ \hline
Total & 80.13  & 80.25$\uparrow$  & 80.70 & 80.62$\downarrow$ & 79.94 & 80.08$\uparrow$ & 76.92 & 78.92$\uparrow$ & 78.12 & 81.71$\uparrow$\\
\bottomrule
\end{tabular}
}
\label{tab:prompt_length}
\end{table}
Each pair of comparison is conducted using exactly the same model and evaluation protocol, differing only in prompt lengths. Most models achieve higher total scores when short prompts are replaced with long prompts, except for VSD1, which verifies our hypothesis on prompt length bias. According to this observation, we adopt the long prompt suite by default for VBench score evaluation. Full results of short-prompt VBench scores refer to Tab.~\ref{tab:vbench_short}. Sample visualization refers to Appendix Sec.~\ref{app_sec:visual_prompt_length}.

\paragraph{Reward Overoptimization}

\begin{figure}[htbp]
    \centering
\includegraphics[width=0.8\columnwidth]{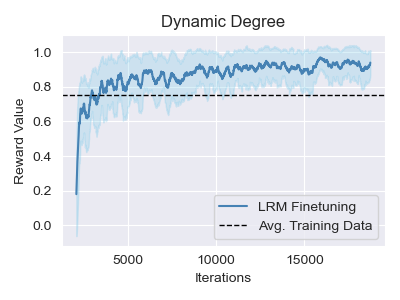}
    \caption{Latent reward model fine-tuning process for dynamic degree.}
    \label{fig:lrm_dd}
\end{figure}

We conduct additional experiments with latent reward fine-tuning on some VBench video-reward metrics, such as dynamic degree, and image-reward metrics, such as JPEG compressibility~\cite{black2023training}. Fig.~\ref{fig:lrm_dd} shows the progress of the latent reward model fine-tuning with the dynamic degree metric in VBench. As the dynamic degree score increases, the generated samples begin to exhibit a ``noise flow" effect that deteriorates the imaging quality. Despite this, the dynamic degree score can rise as high as $0.97$, compared to the average score of $0.75$ in the training data. These findings highlight the trade-off between optimizing for specific metrics and preserving overall visual quality. Appendix Sec.~\ref{app_sec:challenges} Fig.~\ref{fig:lrm_dd_vis} visualizes the training data and generated samples during reward fine-tuning.

As noted in \cite{black2023training}, reward-based optimization is prone to overoptimization, stemming from the divergence between the reward maximization objective and the distribution matching objective used during pre-training. In our video generation experiments, this issue is even more pronounced, with overoptimization sometimes occurring within just a few hundred iterations of fine-tuning. This rapid onset is likely exacerbated by the sample variance inherent in stochastic gradient descent when using a small batch size.

Simply reducing the learning rate or loss weight to mitigate overoptimization is not an ideal solution, as it significantly increases the training time and does not effectively address the core issue. This highlights the need for alternative strategies to balance reward maximization and distribution preservation during fine-tuning.

\section{Conclusion and Discussion}
\label{sec:discuss}
We propose the \texttt{DOLLAR} method for diffusion distillation, combining VSD, CD, and LRM objectives to dramatically accelerate teacher inference. With this approach, the distilled student models achieve significantly higher VBench scores than the teacher model, enhancing both visual quality and text alignment. Notably, this is accomplished within just 40,000 iterations using 8 GPUs, representing only a small fraction of the training dataset and computational resources required for teacher model pre-training. These results highlight the effectiveness of our method, particularly for reward fine-tuning in latent space.
Compared to large pixel-space reward models, LRM is compact and significantly reduces memory costs. Our experiments demonstrate that LRM approximates pixel-space reward models effectively on two representative metrics. 
Further investigation into other reward models remains a direction for future work. 
While the performance improvements are substantial, challenges persist in distillation, fine-tuning, and evaluation. See further discussion on these challenges in Appendix \ref{app_sec:challenges}.

{
    \small
    \bibliographystyle{ieeenat_fullname}
    \bibliography{main}
}
\clearpage
\setcounter{page}{1}
\maketitlesupplementary
\cftsetindents{section}{0em}{2em} 
\cftsetindents{subsection}{1.5em}{2.8em} 
\addtocontents{toc}{\protect\setcounter{tocdepth}{2}}
\renewcommand{\contentsname}{Table of Contents}  

\tableofcontents

\section{Derivations}
\label{sec:math}

\subsection{Proof of Eq.~\eqref{eq:denoise}}
We start from the forward diffusion process of DDPM~\cite{ho2020denoising}.
The distribution of one-step diffusion process $q(x_t|x_{t-1})=\mathcal{N}(x_t; \sqrt{\alpha_t}x_{t-1}, (1-\alpha_t)\mathbf{I})$ can be equivalently written as:
\begin{align}x_t=\sqrt{\alpha_t}x_{t-1}+\sqrt{1-\alpha_t}\varepsilon, \quad \varepsilon\sim\mathcal{N}(0, \mathbf{I})
    \label{eq:x_t}
\end{align}
with $t\in[T]$.

By chain rule, we have
\begin{align}x_t=\sqrt{\bar{\alpha}_t}x_0+\sqrt{1-\bar{\alpha}_t}\varepsilon
    \label{eq:xt_x0}
\end{align}
with $\bar{\alpha}_t=\Pi_{i=1}^t \alpha_i$. Equivalently, we have $x_t\sim q(x_t|x_0)=\mathcal{N}(x_t;\sqrt{\bar{\alpha}_t}x_0, (1-\bar{\alpha}_t)\mathbf{I})$.
This equation is also used to predict:
\begin{align}
\hat{x}_0=\frac{1}{\sqrt{\bar{\alpha}_t}}x_t-\frac{\sqrt{1-\bar{\alpha}_t}}{\sqrt{\bar{\alpha}_t}}\epsilon_\theta
\label{eq:tweedie}
\end{align}
which is called the Tweedie's formula. $\epsilon_\theta$ is the approximated prediction of $\varepsilon$ with a parameterized model by $\theta$.

Proof of the denoising function Eq.~\eqref{eq:denoise} in reverse diffusion process is as follows:
\begin{align*}
    &x_{t-1}=\sqrt{\bar{\alpha}_{t-1}}\hat{x}_0 + \sqrt{1-\bar{\alpha}_{t-1}-\sigma_t^2}\epsilon_\theta+\sigma_t \varepsilon \\
    &=\sqrt{\bar{\alpha}_{t-1}}\hat{x}_0 + \sqrt{1-\bar{\alpha}_{t-1}-\sigma_t^2}(\frac{1}{\sqrt{1-\bar{\alpha}_t}}x_t \\
    &- \frac{\sqrt{\bar{\alpha}_t}}{\sqrt{1-\bar{\alpha}_t}}\hat{x}_0)+\sigma_t \varepsilon\\
    &=(\sqrt{\bar{\alpha}_{t-1}}-\sqrt{1-\bar{\alpha}_{t-1}-\sigma_t^2}\frac{\sqrt{\bar{\alpha}_t}}{\sqrt{1-\bar{\alpha}_t}})\hat{x}_0 \\
    &+ \frac{\sqrt{1-\bar{\alpha}_{t-1}}}{\sqrt{1-\bar{\alpha}_t}}x_t+\sigma_t \varepsilon
\end{align*}
with the first equation follows the posterior sampling in DDIM paper~\cite{song2021denoising}. The second is to plug in the Tweedie's formula. We have the variance term $\sigma_t^2=\frac{(1-\alpha_t)(1-\bar{\alpha}_{t-1})}{1-\bar{\alpha}_t}$.

\subsection{Proof of Eq.~\eqref{eq:x_v_trans}}
Following the Instaflow objective as Eq.~\eqref{eq:insta_loss},
the network directly predicts $v_\theta$, to approximate the target velocity $\tilde{v}^y$ along the rectified flow (RF) trajectory, as the difference of the clean sample and Gaussian noise:
\begin{align}
    v_\theta\approx \tilde{v}^y = x_0 - \varepsilon
    \label{eq:v_predict}
\end{align}
Since the RF sample $y_t$ is a scaled version of diffusion sample $x_t$ as:
\begin{align}
y_t&=\frac{x_t}{\sqrt{\bar{\alpha}_t}+\sqrt{1-\bar{\alpha}_t}}=\gamma_t x_0 + (1-\gamma_t)\varepsilon, \label{eq:y_x}\\\gamma_t&=\frac{\sqrt{\bar{\alpha}_t}}{\sqrt{\bar{\alpha}_t}+\sqrt{1-\bar{\alpha}_t}},
\label{eq:rf_formula}
\end{align}
which satisfies $y_0=x_0$.

Given the velocity prediction $v_\theta$, we can derive the prediction of original sample $x_\theta$ as following, by replacing $x_0$ with prediction $x_\theta$ in Eq.~\eqref{eq:v_predict} and \eqref{eq:y_x}:
\begin{align}
    \gamma_t x_\theta &= y_t - (1-\gamma_t)(x_\theta - v_\theta^y)\\
    x_\theta &= y_t + (1-\gamma_t)v_\theta = y_t + \frac{\sqrt{1-\bar{\alpha}_t}}{\sqrt{\bar{\alpha}_t}+\sqrt{1-\bar{\alpha}_t}}v_\theta
\end{align}
which concludes the proof.

\section{Reward Model Fine-Tuning}
\label{app_sec:rm_finetune}


\subsection{Direct Reward Gradient}
In this section, we discuss in details why the direct reward gradient methods like ReFL~\cite{xu2024imagereward} and DRaFT~\cite{clark2023directly}, cannot fit into the memory efficiently.

Take the HPSv2~\cite{wu2023human} model as an example. It applies fine-tuned version of ViT-H/14 variant of CLIP model, which contains 32 image transformer layers and 24 text transformer layers, each with 16 heads. This constitutes a total of 633 million parameters. Even with FP16 data type, the model weights will occupy 1.25 GB memory. Even for a batch size of 1, the input video tensor of size $(128, 3, 192, 320)$ occupies about 6 GB memory for forward inference only. Backpropagation through the model will drastically increases the memory cost due to gradients storage. Moreover, the memory occupancy roughly scales linearly with the batch size, making it hard to scale up. PickScore~\cite{kirstain2023pick} with CLIP-H model has the similar memory cost in practice. Comparison of parameter numbers and memory costs for reward models and LRMs is shown in Tab.~\ref{tab:model_comparison_memory}. If we take sub-sampling in videos to extract frames for reward optimization, the backward memory (VRAM) cost for different number of frames $H$ is shown in Tab.~\ref{tab:model_comparison_memory_batch}. It indicates that even with frame sub-sampling, the memory cost can still be too large to afford in video model training.

\begin{table}[htbp]
\centering
\caption{Backward memory (VRAM) costs for HPSv2, PickScore reward models with different numbers ($H$) of image ($192\times320$) frames. }
\resizebox{\columnwidth}{!}{
\begin{tabular}{@{}l|cccc@{}}
\toprule
Model                     &  $H=12$   &  $H=24$   &  $H=64$ & $H=128$   \\ \midrule
HPSv2/PickScore    & 12.373 GB        & 20.577 GB  & 48.413 GB              & $>$90 GB                             \\ \bottomrule
\end{tabular}
}
\label{tab:model_comparison_memory_batch}
\end{table}

Given the diffusion modeling in latent space, direct reward gradient methods will also need to backpropagate the gradients from reward model through the large pretrained decoder, this further increases the burden on memory usage.

\subsection{Latent Reward Model For Different Reward Types}
\label{app_sec:lrm_reward_types}

\begin{table*}[htbp]
\centering
\caption{Summary of latent reward models for different pixel-space reward metrics.}
\resizebox{\textwidth}{!}{
\begin{tabular}{l|l|p{4cm}|p{5cm}}
\hline
\textbf{Reward Type} & \textbf{LRM Function} & \textbf{Architecture} & \textbf{Examples} \\ \hline
Image Reward & $\mathcal{R}^l_\phi(x):\mathcal{X} \rightarrow \mathbb{R}$ & 2D CNN backbone & LAION aesthetic~\cite{schuhmann2022laion}, JPEG compressibility~\cite{black2023training} \\ \hline
Text-Image Reward & $\mathcal{R}^l_\phi(x, c):\mathcal{X} \times \mathcal{C} \rightarrow \mathbb{R}$ & 2D CNN + text embedding, cross-attention & HPS~\cite{wu2023human_a, wu2023human}, ImageReward~\cite{xu2024imagereward}, PickScore~\cite{kirstain2023pick} \\ \hline
Video Reward & $\mathcal{R}^l_\phi(\boldsymbol{x}):\mathcal{X}^H \rightarrow \mathbb{R}$ & 2D CNN with average frame reward, or 3D CNN backbone & VBench quality scores~(subject consistency, motion smoothness, etc) \\ \hline
Text-Video Reward & $\mathcal{R}^l_\phi(\boldsymbol{x}, c):\mathcal{X}^H \times \mathcal{C} \rightarrow \mathbb{R}$ & 2D CNN with average frame reward, or 3D CNN backbone, + text embedding, cross-attention & ViCLIP~\cite{wang2023internvid}, VideoScore~\cite{he2024videoscore}, InternVideo2~\cite{wang2024internvideo2}, VBench semantic scores~(object class, human action, color, etc) \\ \hline
\end{tabular}
}
\label{tab:lrm_types}
\end{table*}

The proposed latent reward model method is compatible with any type of reward metrics as introduced previously, regardless of its differentiability and input formats. Here we consider several types of commonly used reward metrics: image reward, text-image reward, video reward and text-video reward. For each category, we provide examples and explain how LRM, with its diverse architectures, supports these metrics. A summary of this compatibility is provided in Tab.~\ref{tab:lrm_types}, with further details outlined below:

\begin{itemize}
    \item Image reward: $\mathcal{I}\rightarrow \mathbb{R}$.

    The LRM is $\mathcal{R}^l_\phi(x):\mathcal{X}\rightarrow \mathbb{R}, x=\text{Encode}(i),i\in\mathcal{I}$. It has the image backbone as a 2D convolutional neural network (CNN).

    Examples include LAION aesthetic quality~\cite{schuhmann2022laion}, JPEG compressibility~\cite{black2023training}.
    
    \item Text-image reward: $\mathcal{C}\times\mathcal{I}\rightarrow \mathbb{R}$. 

    The LRM is $\mathcal{R}^l_\phi(x, c):\mathcal{X}\times\mathcal{C}\rightarrow \mathbb{R}, x=\text{Encode}(i),i\in\mathcal{I}$. It has the image backbone as a 2D CNN and text embedding $e_c$ as inputs, with a cross-attention module for mixing image features $e_x$ and text features $e_c$: $\text{Softmax}(\mathbf{Q}(e_x) \cdot \mathbf{K}(e_c)^\top)\cdot \mathbf{V}(e_c)$.
    
    Examples include human preference score (HPS)~\cite{wu2023human_a, wu2023human}, ImageReward~\cite{xu2024imagereward}, PickScore~\cite{kirstain2023pick}.
    
    \item Video reward: $\mathcal{I}^H\rightarrow \mathbb{R}$ where $H$ is the number of frames in each video. 

    The LRM can be either (1). $\mathcal{R}^l_\phi(x):\mathcal{X}\rightarrow \mathbb{R}, x=\text{Encode}(i),i\in\mathcal{I}$ using a 2D CNN image backbone with average frame reward $\frac{1}{H}\sum_{k=1}^H\mathcal{R}^l_\phi(x_k)$ as video reward or (2). $\mathcal{R}^l_\phi(x_1, \dots, x_H):\mathcal{X}^H\rightarrow \mathbb{R}$ using a 3D CNN as video backbone.
    
    Examples include 7 quality scores in VBench (subject consistency, background consistency, motion smoothness, etc).
    
    \item Text-video reward: $\mathcal{C}\times\mathcal{I}^H\rightarrow \mathbb{R}$. 

    The LRM can be either (1). $\mathcal{R}^l_\phi(x, c):\mathcal{X}\times\mathcal{C}\rightarrow \mathbb{R}$ using a 2D CNN image backbone with average frame reward $\frac{1}{H}\sum_{k=1}^H\mathcal{R}^l_\phi(x_k, c)$ as video reward or (2). $\mathcal{R}^l_\phi(x_1, \dots, x_H, c):\mathcal{X}^H\times\mathcal{C}\rightarrow \mathbb{R}$ using a 3D CNN as video backbone, with additional text embedding $e_c$ as inputs, and cross-attention for mixing image features $e_x$ and text features $e_c$: $\text{Softmax}(\mathbf{Q}(e_x) \cdot \mathbf{K}(e_c)^\top)\cdot \mathbf{V}(e_c)$.

    Examples include  ViCLIP~\cite{wang2023internvid}, VideoScore~\cite{he2024videoscore}, InternVideo2~\cite{wang2024internvideo2} and 9 semantic score metrics in VBench (object class, human action, color, etc).
\end{itemize}

\paragraph{Architecture Details.}
The image only LRM $\mathcal{R}^l_\phi(x)$ has architecture detailed in Tab.~\ref{tab:conv_only_reward_model_architecture}.
The text-image LRM $\mathcal{R}^l_\phi(x,c)$ has architecture detailed in Tab.~\ref{tab:cross_attention_reward_model_architecture}. For video LRM and text-video LRM, we apply the same architectures with frame averaging in our experiments.

\paragraph{Discussions.}
The latent reward model can be utilized in two ways: it can either be pretrained or trained concurrently with the student model during fine-tuning, as demonstrated in our experiments. Furthermore, this approach can also be extended to fine-tune the teacher model. Alternatively, one could bypass the reward model in pixel space entirely and directly employ a latent reward model from the outset. However, we argue that such an approach is likely to be limited to specific fixed latent spaces and may lack generalizability across models. This is because pretrained encoder-decoder models can vary significantly and often do not share a unified latent space, particularly in existing image and video models.

\begin{table*}[htbp]
\centering
\caption{Architecture of the image latent reward model}
\resizebox{\textwidth}{!}{
\begin{tabular}{@{}lccccccc@{}}
\toprule
Layer                       & Input Shape           & Output Shape          & Kernel Size & Stride & Padding & Number of Parameters \\ \midrule
\textbf{Input}              & (batch, C, H, W)   &                       &            &        &         &                       \\
\textbf{Conv2d + GroupNorm + SiLU} & (batch, C, H, W) & (batch, 128, 6, 10)  & 4x4        & 4      & 1       & 24,704                \\
\textbf{Conv2d + GroupNorm + SiLU} & (batch, 128, 6, 10) & (batch, 128, 3, 5)   & 3x3        & 2      & 1       & 147,584               \\
\textbf{AdaptiveAvgPool2d}  & (batch, 128, 3, 5)    & (batch, 128, 1, 1)    & -          & -      & -       & 0                     \\
\textbf{Conv2d}             & (batch, 128, 1, 1)    & (batch, 128, 1, 1)    & 1x1        & 1      & 0       & 16,512                \\
\textbf{Flatten}            & (batch, 128, 1, 1)    & (batch, 128)          & -          & -      & -       & 0                     \\
\textbf{Linear}             & (batch, 128)          & (batch, 1)            & -          & -      & -       & 129                   \\ \midrule
\textbf{Total Parameters}   &                       &                       &            &        &         & 189,441               \\ \bottomrule
\end{tabular}
}
\label{tab:conv_only_reward_model_architecture}
\end{table*}

\begin{table*}[htbp]
\centering
\caption{Architecture of the text-image latent reward model}
\resizebox{\textwidth}{!}{
\begin{tabular}{@{}lccccccc@{}}
\toprule
Layer                                & Input Shape            & Output Shape           & Kernel Size / Projection & Stride & Padding & Number of Parameters \\ \midrule
\textbf{Input Image}                 & (batch, C, H, W)    &                        &                          &        &         &                       \\
\textbf{Conv2d + GroupNorm + SiLU}   & (batch, C, H, W)    & (batch, 128, 6, 10)    & 4x4                      & 4      & 1       & 24,704                \\
\textbf{Conv2d + GroupNorm + SiLU}   & (batch, 128, 6, 10)    & (batch, 128, 3, 5)     & 3x3                      & 2      & 1       & 147,584               \\
\textbf{AdaptiveAvgPool2d}           & (batch, 128, 3, 5)     & (batch, 128, 1, 1)     & -                        & -      & -       & 0                     \\
\textbf{Conv2d}                      & (batch, 128, 1, 1)     & (batch, 128, 1, 1)     & 1x1                      & 1      & 0       & 16,512                \\
\textbf{Flatten (Image Features)}    & (batch, 128, 1, 1)     & (batch, 128)           & -                        & -      & -       & 0                     \\ \midrule
\textbf{Input Text}                  & (batch, L, D)     &                        &                          &        &         &                       \\
\textbf{Text MLP}                    & (batch, L, D)     & (batch, 256, 128)      & -                        & -      & -       & 524,544               \\
\textbf{Average Pooling (Text Features)} & (batch, 256, 128) & (batch, 128)           & -                        & -      & -       & 0                     \\ \midrule
\textbf{Query Projection (Linear)}   & (batch, 128)           & (batch, 128)           & -                        & -      & -       & 16,512                \\
\textbf{Key Projection (Linear)}     & (batch, 128)           & (batch, 128)           & -                        & -      & -       & 16,512                \\
\textbf{Value Projection (Linear)}   & (batch, 128)           & (batch, 128)           & -                        & -      & -       & 16,512                \\
\textbf{Attention Mechanism (Softmax)} & (batch, 1, 1)       & (batch, 1, 1)          & -                        & -      & -       & 0                     \\
\textbf{Final Linear (Output Layer)} & (batch, 128)           & (batch, 1)             & -                        & -      & -       & 129                   \\ \midrule
\textbf{Total Parameters}            &                        &                        &                          &        &         & 763,009               \\ \bottomrule
\end{tabular}
}
\label{tab:cross_attention_reward_model_architecture}
\end{table*}

\subsection{Latent Reward Model Training}

\begin{figure}[htbp]
    \centering
\includegraphics[width=0.49\columnwidth]{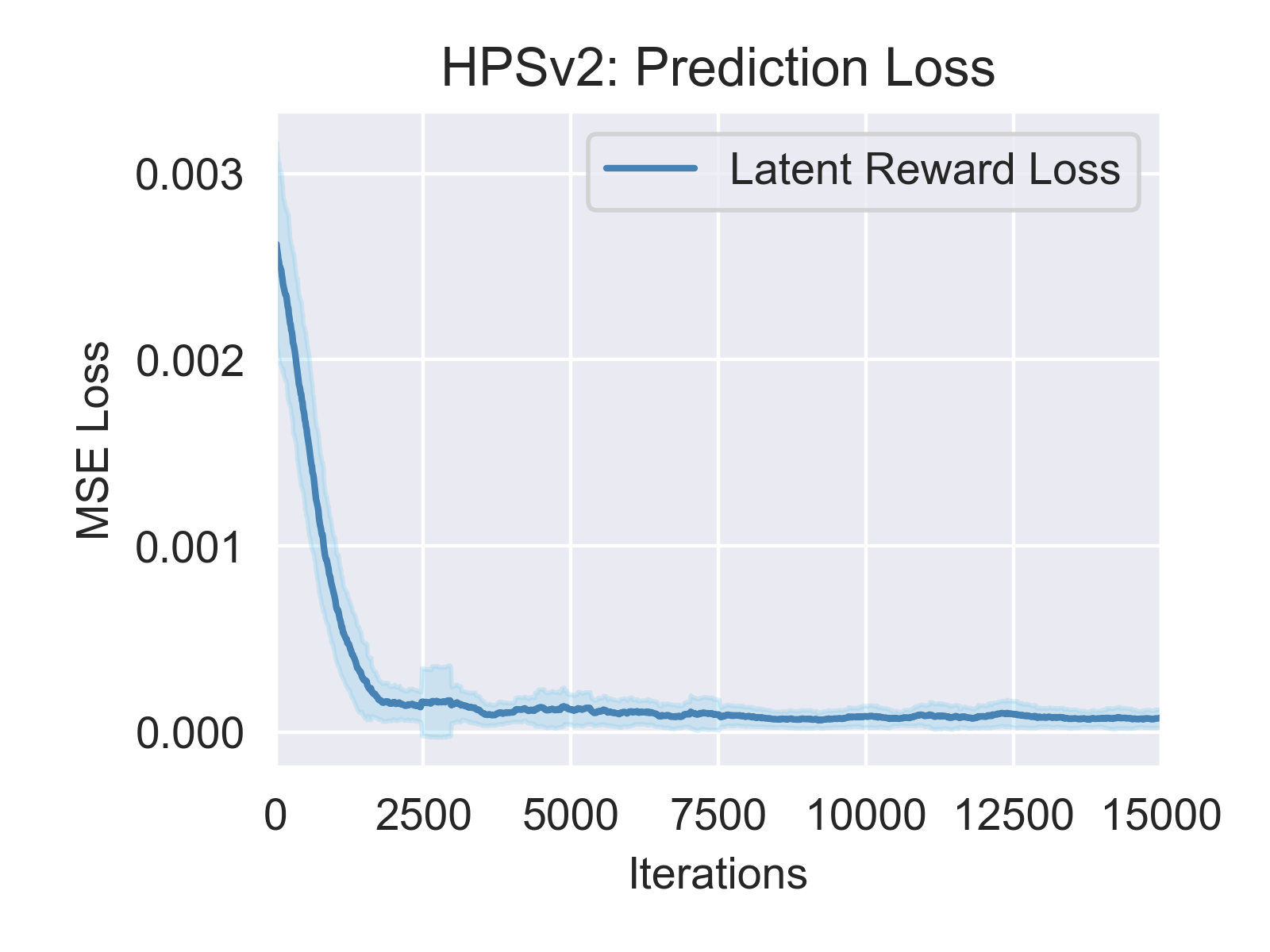}
\includegraphics[width=0.49\columnwidth]{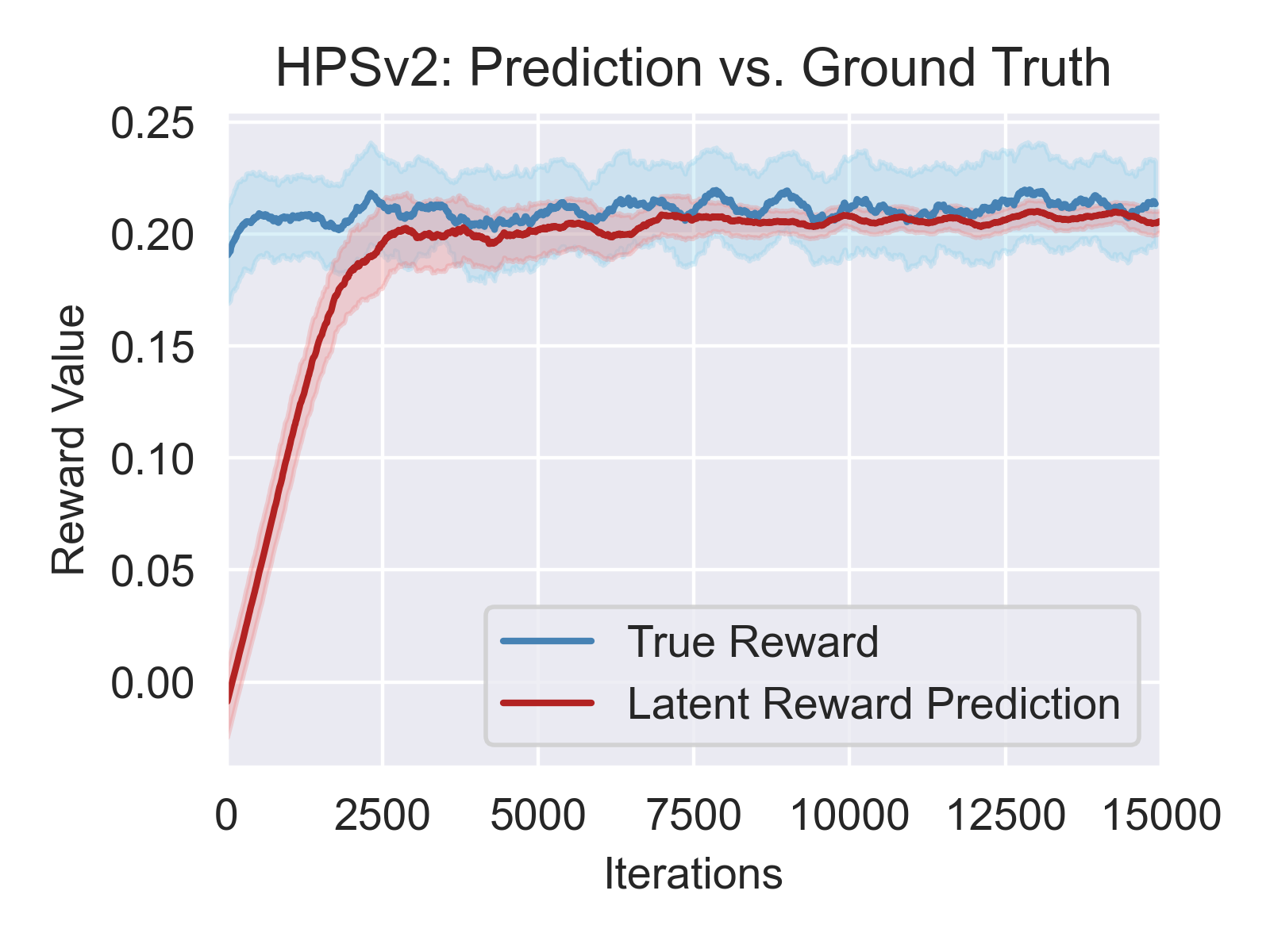}
    \caption{The learning process of LRM with HPSv2 reward.
    }
    \label{fig:lrm_hps_detail}
\end{figure}

\begin{figure}[htbp]
    \centering
\includegraphics[width=0.49\columnwidth]{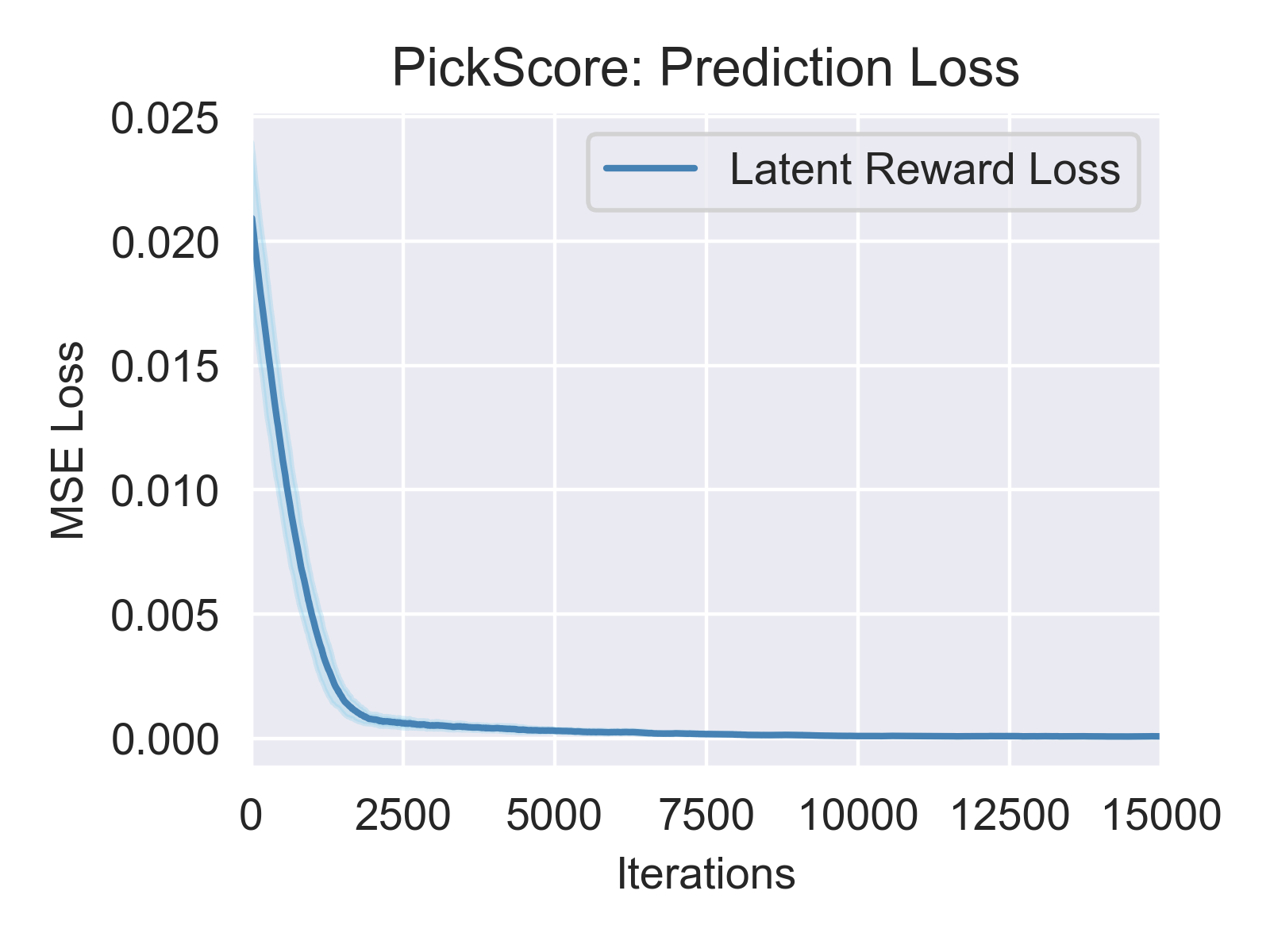}
\includegraphics[width=0.49\columnwidth]{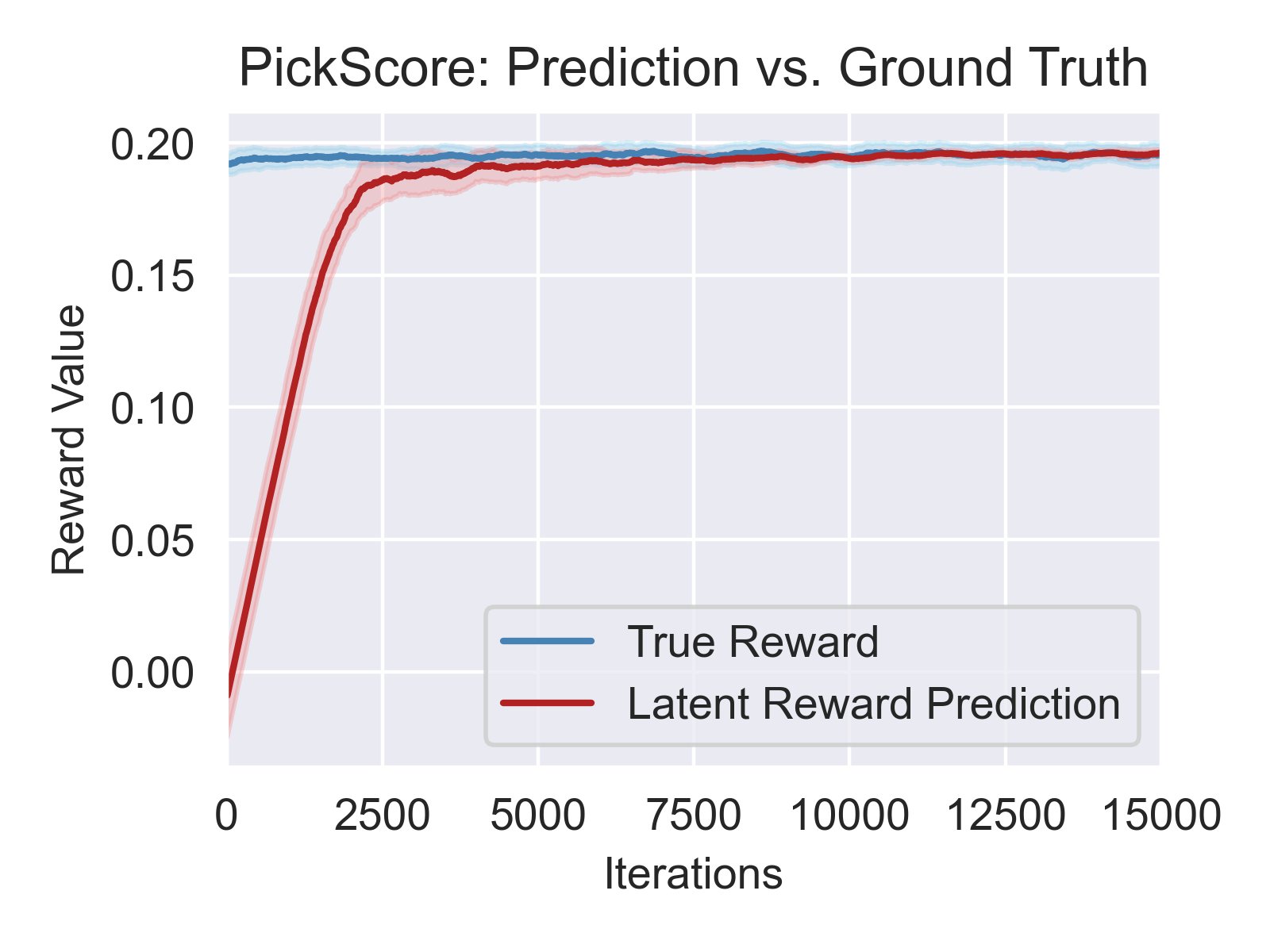}
    \caption{The learning process of LRM with PickScore reward.
    }
    \label{fig:lrm_pick_detail}
\end{figure}

Fig.~\ref{fig:lrm_hps_detail} and Fig.~\ref{fig:lrm_pick_detail} show the learning curves of latent reward models (LRMs) with two original pixel-space rewards HPSv2 and PickScore, respectively, during the distillation process. The loss for training is VSD+LRM. Left figure displays the MSE loss for LRM prediction against the ground-truth pixel-space reward value. Right figure displays the LRM predicted reward values $\mathcal{R}_\phi^l(x_0, c)$ and ground truth reward values $\mathcal{R}(x_0, c)$ on training samples from the dataset $x_0\sim\mathcal{X}$. This demonstrates that the LRM achieves rapid convergence within 2000–3000 training iterations, even when operating in a significantly lower-dimensional latent space.  The small approximation errors ensure the effectiveness of fine-tuning with learned LRM.

\subsection{Denoising Diffusion Policy Optimization}
\label{sec:ddpo}
Denoising Diffusion Policy Optimization (DDPO) serves as the baseline for comparison with our proposed LRM method. In this section, we delve into the implementation details of DDPO.
By applying the REINFORCE algorithm on denoising process of diffusion models, the DDPO$_\text{SF}$ algorithm follows the score function policy gradient:
\begin{align}
   \nabla_\theta \mathcal{J}=\mathbb{E}[\sum_{t=1}^T \nabla_\theta \log p_\theta (x_{t-1}|x_t, c)R(x_0, c)]
\end{align}
This is the online version for gradient estimation, which requires to sample $x_{t-1}$ as well as calculating the probabilities $p_\theta (x_{t-1}|x_t, c)$ along the sampling process at the same time, such that the model parameters $\theta$ remain the same for sampling and probability estimation. The update will only take one step to preserve the online estimation property. Original paper~\cite{black2023training} also proposes another version for offline policy gradient estimation with importance sampling to allow multi-step updates. As log-probability $\log p_\theta (x_{t-1}|x_t, c)$ needs to be estimated during the sampling process, we cannot take sampling process as Eq.~\eqref{eq:denoise}, but estimating the posterior mean $\mu_\theta$ and standard deviation $\sigma$ instead:
\begin{align}
    \mu_\theta(x_{t-1};x_t)&= \frac{(1 - \alpha_t)\sqrt{\bar{\alpha}_t}}{1-\bar{\alpha}_t}x_\theta  + \frac{\sqrt{\alpha}_t(1-\bar{\alpha}_{t-1})}{1-\bar{\alpha}_t} x_t\nonumber\\
    \sigma_t &= \sqrt{(1 - \alpha_t)\frac{1-\bar{\alpha}_{t-1}}{1-\bar{\alpha}_t}}
    \label{eq:mean_std}
\end{align}
with $x_\theta$ following Eq.~\eqref{eq:x_v_trans}. $x_{t-1}$ will be sampled from $\mathcal{N}(\mu_\theta(x_{t-1};x_t), \sigma_t)$, with log-probability of the sample as:
\begin{align}
    \log p_\theta(x_{t-1}|\mu_\theta, \sigma, c)=-\frac{1}{2} \left( \frac{(x_{t-1} - \mu_\theta)^2}{\sigma^2} + \log(2\pi \sigma^2)\right)
    \label{eq:log_prob}
\end{align}
The practical procedure of DDPO$_\text{SF}$ is outlined in Alg.~\ref{alg:ddpo}. Due to VRAM memory constraints, we employ the REINFORCE policy gradient with truncation, allowing gradient tracking for a maximum of $N=2$ steps during training.
Specifically, for a student model with a sampling time sequence $[T, \dots, t_\text{min}]=[999, 749, 499, 249 ]$, the gradient update steps will only take the last two steps $t_n\in\{499, 249\}$, rather than all timesteps. This truncation is used to estimate the log-probabilities of samples at $t_{n-1}$. Here, $\text{Dec}(\cdot)$ represents the pretrained video decoder, while the reward model $R$ operates in the original pixel space. We use $.\text{detach}()$ to indicate a stop-gradient function.

\begin{algorithm}[H]
\caption{DDPO practical procedure.}
\small
\begin{algorithmic}[1]
\State \textbf{Input:} Distilled student model $G_{\theta}$, dataset $\mathcal{D}=\{(c, i)\}$
\State \textbf{Output:} Fine-tuned student few-step generator $G_\theta$.
\While{\text{train}}
\State \texttt{//Sample from random noise along entire diffusion trajectory}
\State $x_T\leftarrow \varepsilon\sim\mathcal{N}(0, \mathbf{I})$
\For{$t_n\in [T, \dots, t_\text{min}]$}
\State Get posterior Gaussian $(\mu_\theta, \sigma)$ with $v_\theta(x_{t_n}\text{.detach()},t_n)$ \texttt{//Eq.~\eqref{eq:mean_std}}
\State Sample $x_{t_{n-1}}\sim\mathcal{N}(\mu_\theta, \sigma \mathbf{I})$
\State Estimate $\log p_\theta(x_{t_{n-1}}|x_{t_n}, c)$  \texttt{//Eq.~\eqref{eq:log_prob}}
\EndFor
\State Get reward $R=R(\text{Dec}(\hat{x}_0), c)\text{.detach()}$
\State \texttt{//REINFORCE policy gradient with truncation}
\State $\mathcal{L}_{\text{DDPO}_\text{SF}}=-\sum_n^N \log p_\theta(x_{t_{n-1}}\text{.detach()}|x_{t_n}, c) \cdot R$
\State $G_\theta\leftarrow \text{GradientDescent}(\theta, \mathcal{L}_{\text{DDPO}_\text{SF}})$
\EndWhile
\end{algorithmic}
\label{alg:ddpo}
\end{algorithm}

\begin{figure}[htbp]
    \centering
\includegraphics[width=0.48\columnwidth]{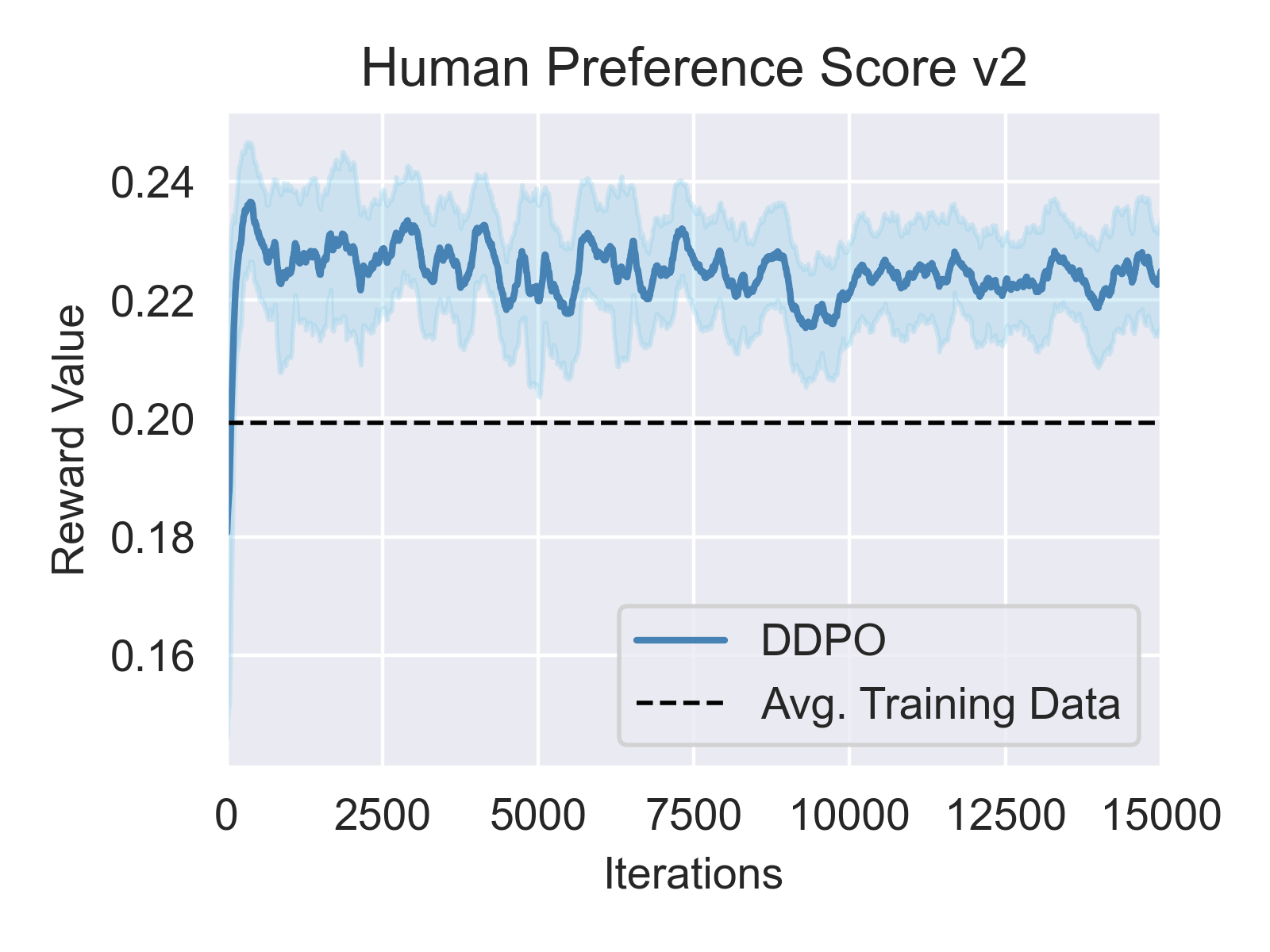}
\includegraphics[width=0.48\columnwidth]{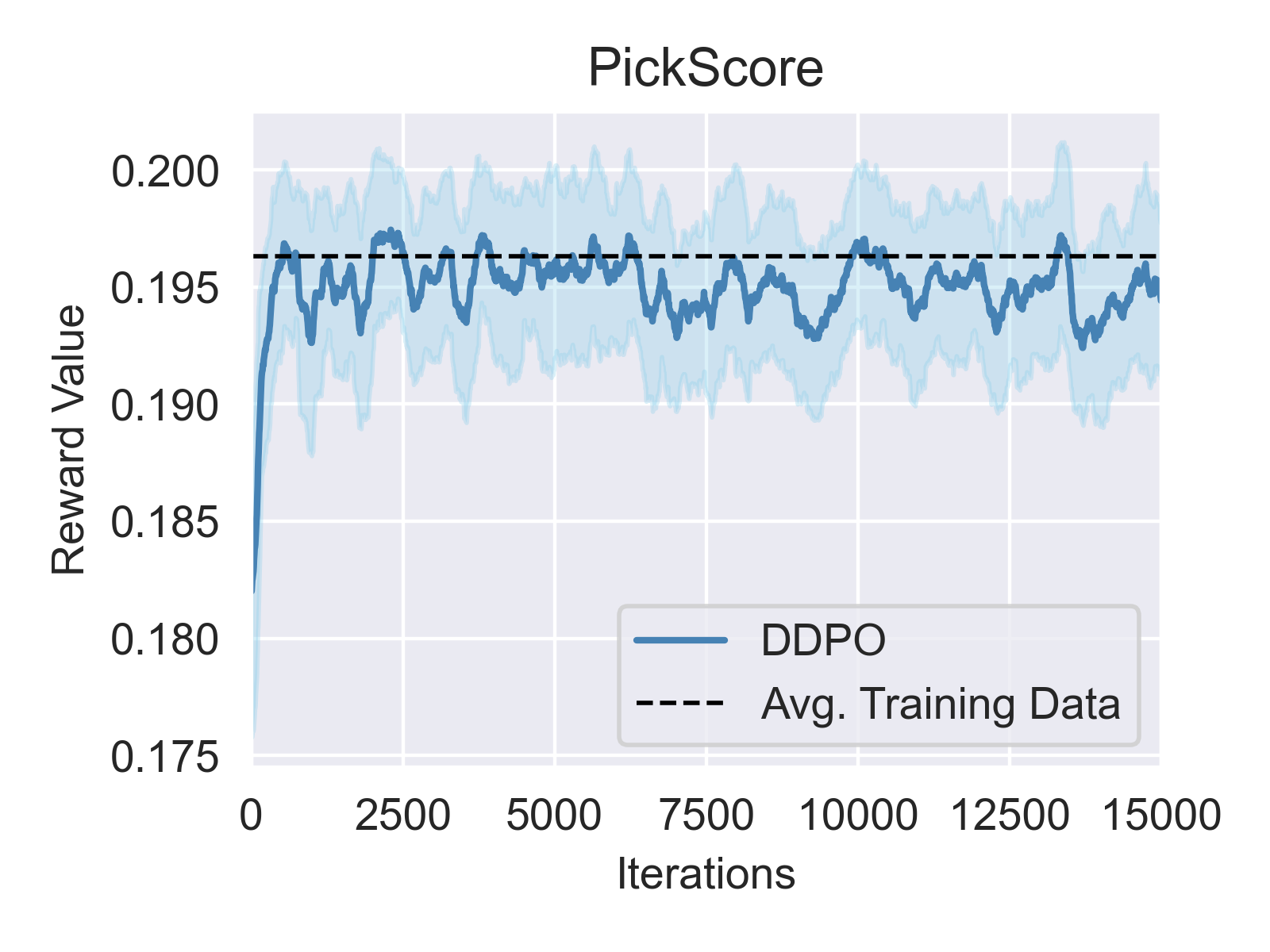}
    \caption{Reward model fine-tuning process with VSD+DDPO under reward models HPSv2 and PickScore.
    }
    \label{fig:ddpo_curves}
\end{figure}

\paragraph{Learning Curves.}
The training process of VSD+DDPO for two reward metrics are shown in Fig.~\ref{fig:ddpo_curves}. The learning curve shows the reward values $\mathcal{R}(x_0,c)$ for generated samples $\hat{x}_0$ through iterative denoising along the full diffusion trajectories, during the fine-tuning process.

The learning curves of DDPO are not directly comparable to those of the LRM methods shown in Fig.~\ref{fig:hpsv2}. This difference arises because DDPO samples across the entire diffusion trajectory to obtain the predicted $\hat{x}_0$ for reward evaluation, whereas LRM performs one-step prediction using $x_\theta=\frac{x_t+\sqrt{1-\bar{\alpha}_t}v^w_\theta(x_t, t)}{\sqrt{\bar{\alpha}_t}+\sqrt{1-\bar{\alpha}_t}}$, as defined in Eq.~\eqref{eq:x_v_trans}. Consequently, the LRM samples tend to be noisier and yield lower rewards during fine-tuning. A fair comparison involves evaluating the rewards of the final generated samples after the model fine-tuning, as presented in Tab.~\ref{tab:reward_method_compare} of main paper.

\section{Additional Experimental Results}
\begin{figure*}[htbp]
    \centering
\includegraphics[width=0.99\textwidth]{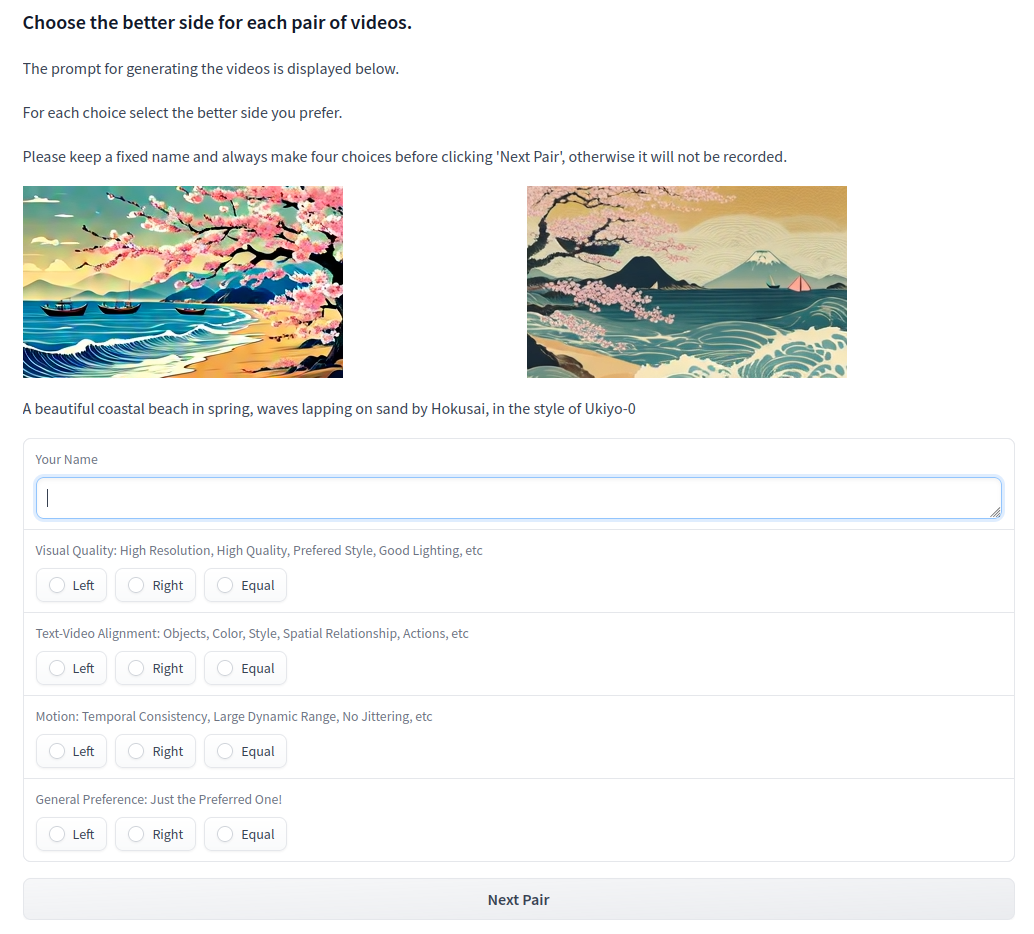}
    \caption{The user interface for human evaluation experiments.
    }
    \label{fig:human_eval_web}
\end{figure*}
\subsection{Human Evaluation}
\label{app_sec:human_eval}
\paragraph{Human Evaluation Details.}
Fig.~\ref{fig:human_eval_web} displays the user interface for human evaluation experiments. The four choices include visual quality, text-video alignment, motion and general preference, which correspond to the four reported metrics in Fig.~\ref{fig:human_eval}.
For the pairwise comparison of methods, the videos are randomly sampled from 4730 videos with 946 VBench long prompts, with 5 videos generated for each prompt under different random seeds. The videos are all displayed at a resolution of $192\times 320$ with 128 frames for our methods. For a fair comparison, videos for the baseline method Gen-3 ($768\times1280$) are resized to $192\times 320$.
Each pair of videos requires approximately 20–30 seconds for evaluation. To prevent positional bias, the left and right placement of the videos is randomly shuffled for each evaluation session.

\paragraph{Human Evaluation Results.}
We conduct 6 rounds of human evaluation on sampled videos with different methods, comparing models under the following settings:
\begin{itemize}
    \item VSD+LRM with HPSv2 as reward model versus VSD method, to verify the effectiveness of LRM for fine-tuning.
    \item VSD+LRM versus VSD+DDPO, both with HPSv2 as the reward model, to compare the LRM and DDPO methods for reward fine-tuning. 
    \item VSD+LRM with HPSv2 reward versus PickScore reward, to testify the effectiveness of two reward models.
    \item VSD+CD+LRM with HPSv2 reward versus Gen-3 model results, to compare our distilled models with one of the best present models in Tab.~\ref{tab:compare_vbench} according to VBench.
    \item VSD+CD+LRM with PickScore as reward model versus the teacher model.
    \item VSD+CD+LRM with HPSv2 as reward model versus the teacher model.
\end{itemize}

The results for above 6 experiments are summarized in Fig.~\ref{fig:human_eval}. Each value indicates the winning rate, with the equal performance option excluded.

\paragraph{Discussions.}
In the comparison of VSD+CD+LRM with PickScore versus the teacher model, human evaluation results indicate that the student underperforms the teacher in text-video alignment, motion and general preference, although it has a much higher score in VBench evaluation (82.37 vs. 80.25) as Tab.~\ref{tab:compare_vbench}. Specifically, the semantic score in VBench is 77.90 for the student and 73.71 for the teacher, while human evaluation arrives at the opposite conclusion. This discrepancy highlights a mismatch between VBench and human evaluation metrics, posing a challenge in accurately assessing video generation quality.
Our empirical findings suggest that humans tend to reject videos exhibiting subtle flaws such as shape distortions, unnatural motions, or other elements that appear less natural or physically realistic. Humans are highly sensitive to these imperfections, which influence their preference. By contrast, VBench metrics, primarily based on pretrained image understanding models, are more influenced by factors such as coloring, lighting, aesthetics, and imaging quality, while being less sensitive to the naturalness and physical realism of videos. Measuring physical realism directly from pixels remains a challenge in general. We hypothesize that this difference contributes to the observed divergence between VBench scores and human preferences in our experiments.

\subsection{Complete VBench Scores}
\label{app_sec:complete_vbench}
The breakdown VBench scores and reward scores for Tab.~\ref{tab:reward_method_compare} of main paper are shown in Tab.~\ref{tab:vbench_lrm_ddpo} and Tab.~\ref{tab:reward_method_compare_complete}. The breakdown VBench scores for Tab.~\ref{tab:infer_steps} of main paper are shown in Tab.~\ref{tab:vbench_infer_steps}.  The breakdown VBench scores for main paper Tab.~\ref{tab:cd_steps} and Tab.~\ref{tab:parameterization} are shown in Tab.~\ref{tab:vbench_vsd_cd}. VSD4 indicates the VSD loss for 4-step inference of the student, as our default setting. CD1 and CD5 indicate the CD loss with denoising steps $m=1$ and $m=5$, respectively.

\begin{table*}[htbp]
\centering
\caption{Comparison of LRM with DDPO using VBench (long prompt) and fine-tuning reward metrics HPSv2 and PickScore.}
\begin{tabular}{@{}l|ccc|ccc@{}}
\toprule
Reward Model & \multicolumn{3}{c|}{PickScore} & \multicolumn{3}{c}{HPSv2} \\ \midrule
Method & VSD+DDPO & VSD+LRM & VSD+CD+LRM &  VSD+DDPO & VSD+LRM & VSD+CD+LRM   \\ \midrule
Quality Score & 82.99 & \textbf{84.01} & 83.49  & 82.97 & 83.53 & \textbf{83.83}\\ \hline
Semantic Score & 77.26 & 72.51 & \textbf{77.90} & 74.56 & 75.67 & \textbf{77.51}\\ \hline
Total Score & 81.84  & 81.71 & \textbf{82.37} & 81.29 & 81.96 & \textbf{82.57}\\ \hline
Reward & $0.207\pm0.011$ & $0.207\pm0.011$ & $\mathbf{0.210\pm0.011}$ &  $0.271\pm0.027$  & $0.276\pm0.028$ & $\mathbf{0.277\pm0.029}$   \\
\bottomrule
\end{tabular}
\label{tab:reward_method_compare_complete}
\end{table*}

\begin{table*}[htbp]
\centering
\caption{Comparison of VBench scores for DDPO and LRM methods (values in percentage).}
\begin{tabular}{@{}p{3.5cm}p{1.5cm}p{1.5cm}p{2.cm}p{2.cm}p{2.cm}p{2.cm}@{}}
\toprule
Model & Teacher & VSD4 & VSD4+DDPO (PickScore) & VSD4+LRM (PickScore) & VSD4+DDPO (HPSv2) & VSD4+LRM (HPSv2) \\ \midrule
Subject Consistency & 83.99 & 93.26 & \textbf{95.26} & 94.34 & 93.27 & 91.99 \\
Background Consistency & 93.78 & 95.82 & 96.21 & 96.08 & 96.22 & \textbf{96.93} \\
Temporal Flickering & 96.42 & 95.79 & 96.56 & 95.85 & 96.64 & \textbf{96.80} \\
Motion Smoothness & \textbf{98.09} & 97.48 & 96.45 & 97.30 & 97.56 & 97.39 \\
Dynamic Degree & \textbf{99.44} & 58.61 & 85.83 & 94.44 & 81.67 & 85.56 \\
Aesthetic Quality & 61.21 & 61.34 & 61.85 & 61.84 & 61.66 & \textbf{63.14} \\
Imaging Quality & 63.87 & 68.21 & 65.98 & \textbf{68.49} & 66.39 & 67.35 \\
Object Class & 85.79 & \textbf{94.72} & 94.29 & 87.28 & 91.91 & 90.65 \\
Multiple Objects & 52.59 & 69.24 & \textbf{72.33} & 55.11 & 65.32 & 60.34 \\
Human Action & 99.60 & \textbf{99.80} & 98.00 & 98.80 & 98.20 & 99.40 \\
Color & \textbf{77.00} & 71.81 & 76.28 & 76.12 & 70.00 & 73.69 \\
Spatial Relationship & 51.40 & 64.80 & \textbf{65.19} & 54.25 & 61.75 & 63.81 \\
Scene & 49.99 & 51.89 & 52.60 & 49.17 & 49.65 & \textbf{53.43} \\
Temporal Style & \textbf{26.45} & 24.93 & 25.00 & 24.39 & 24.53 & 25.02 \\
Appearance Style & \textbf{24.83} & 24.31 & 23.99 & 23.68 & 23.66 & 24.53 \\
Overall Consistency & \textbf{27.89} & 26.38 & 26.43 & 25.97 & 26.67 & 26.76 \\
\midrule
Quality Score & 81.89 & 80.95 & 82.99 & 84.01 & 82.97 & \textbf{83.53} \\
Semantic Score & 73.71 & 76.61 & \textbf{77.26} & 72.51 & 74.56 & 75.67 \\
Total Score & 80.25 & 80.08 & 81.84 & 81.71 & 81.29 & \textbf{81.96} \\
\bottomrule
\end{tabular}
\label{tab:vbench_lrm_ddpo}
\end{table*}

\begin{table*}[htbp]
\centering
\caption{VBench scores with short prompts (values in percentage) for some models.}
\begin{tabular}{@{}p{3.5cm}p{1.5cm}p{1.5cm}p{2.cm}p{2.cm}p{2.cm}p{2.cm}@{}}
\toprule
Model & Teacher & VSD1 & VSD4 & VSD4+LRM (PickScore) & VSD4+LRM (HPSv2) & VSD4+CD1 \\ \midrule
Subject Consistency & 84.80 & 89.39 & 92.98 & 94.13 & 91.72 & 84.83 \\
Background Consistency & 94.10 & 94.91 & 96.12 & 95.14 & 96.34 & 93.87 \\
Temporal Flickering & 96.12 & 96.96 & 96.55 & 95.29 & 96.50 & 94.84 \\
Motion Smoothness & 97.99 & 97.57 & 97.12 & 96.77 & 96.65 & 97.08 \\
Dynamic Degree & 97.78 & 91.94 & 61.39 & 93.06 & 94.17 & 93.61 \\
Aesthetic Quality & 57.74 & 57.33 & 58.24 & 58.08 & 60.20 & 55.32 \\
Imaging Quality & 65.41 & 62.10 & 67.79 & 68.97 & 67.12 & 62.28 \\
Object Class & 88.45 & 89.89 & 93.12 & 57.34 & 92.67 & 80.54 \\
Multiple Objects & 56.54 & 73.86 & 72.29 & 38.43 & 66.45 & 47.90 \\
Human Action & 99.60 & 98.00 & 98.20 & 92.00 & 96.60 & 96.60 \\
Color & 77.75 & 86.19 & 79.55 & 78.36 & 82.90 & 67.55 \\
Spatial Relationship & 51.21 & 70.13 & 70.09 & 44.49 & 63.32 & 55.34 \\
Scene & 50.89 & 35.32 & 42.95 & 13.31 & 36.90 & 29.53 \\
Temporal Style & 26.52 & 26.21 & 24.91 & 23.03 & 25.11 & 25.02 \\
Appearance Style & 24.76 & 23.93 & 23.87 & 23.47 & 24.45 & 23.03 \\
Overall Consistency & 27.96 & 28.06 & 26.42 & 24.81 & 27.05 & 27.13 \\
\midrule
Quality Score & 81.50 & 81.60 & 80.75 & 82.64 & 83.03 & 79.27 \\
Semantic Score & 74.64 & 77.10 & 76.67 & 60.04 & 75.08 & 67.52 \\
Total Score & 80.13 & 80.70 & 79.94 & 78.12 & 81.44 & 76.92 \\
\bottomrule
\end{tabular}
\label{tab:vbench_short}
\end{table*}

\begin{table}[htbp]
\centering
\caption{Comparison of VBench scores across models with different inference steps (values in percentage).}
\resizebox{\columnwidth}{!}{%
\begin{tabular}{@{}l|c|ccc@{}}
\toprule
Model & Teacher & \multicolumn{3}{|c}{Student (VSD)} \\ \midrule
Inference Steps & 50 & 1 & 2 & 4 \\ \midrule
Subject Consistency & 83.99 & 90.09 & 92.27 & \textbf{93.26} \\
Background Consistency & 93.78 & 94.39 & 95.17 & \textbf{95.82} \\
Temporal Flickering & 96.42 & \textbf{96.79} & 95.73 & 95.79 \\
Motion Smoothness & \textbf{98.09} & 97.72 & 96.96 & 97.48 \\
Dynamic Degree & \textbf{99.44} & 86.39 & 93.33 & 58.61 \\
Aesthetic Quality & 61.21 & 60.26 & \textbf{61.55} & 61.34 \\
Imaging Quality & 63.87 & 61.82 & 66.09 & \textbf{68.21} \\
Object Class & 85.79 & 90.03 & 87.86 & \textbf{94.72} \\
Multiple Objects & 52.59 & 67.71 & 58.06 & \textbf{69.24} \\
Human Action & 99.60 & 98.40 & 99.60 & \textbf{99.80} \\
Color & \textbf{77.00} & 74.43 & 65.44 & 71.81 \\
Spatial Relationship & 51.40 & \textbf{69.17} & 63.96 & 64.80 \\
Scene & 49.99 & 49.74 & \textbf{52.21} & 51.89 \\
Temporal Style & \textbf{26.45} & 26.03 & 25.19 & 24.93 \\
Appearance Style & \textbf{24.83} & 23.90 & 23.77 & 24.31 \\
Overall Consistency & \textbf{27.89} & 27.14 & 26.91 & 26.38 \\
\midrule
Quality Score & 81.89 & 81.61 & \textbf{82.71} & 80.95 \\
Semantic Score & 73.71 & \textbf{76.66} & 73.86 & 76.61 \\
Total Score & 80.25 & 80.62 & \textbf{80.94} & 80.08 \\
\bottomrule
\end{tabular}
}
\label{tab:vbench_infer_steps}
\end{table}

\begin{table}[htbp]
\centering
\caption{Comparison of VBench scores for VSD+CD methods (values in percentage).}
\resizebox{\columnwidth}{!}{%
\begin{tabular}{@{}lcccc@{}}
\toprule
Metric & Teacher & VSD4+CD1 & VSD4+CD5 & VSD4+CD5\\ \midrule
Parameterization & $v_\theta$ & $v_\theta$ & $v_\theta$ & $x_\theta$ \\ \midrule
Subject Consistency & 83.99 & 86.36 & \textbf{86.37} & 85.47 \\
Background Consistency & 93.78 & \textbf{94.84} & 94.70 & 93.37 \\
Temporal Flickering & 96.42 & 95.60 & 96.48 & \textbf{96.51} \\
Motion Smoothness & \textbf{98.09} & 97.70 & 98.04 & 98.05 \\
Dynamic Degree & \textbf{99.44} & 87.50 & 90.28 & 95.28 \\
Aesthetic Quality & 61.21 & 60.16 & \textbf{62.16} & 60.95 \\
Imaging Quality & 63.87 & 62.85 & \textbf{65.24} & 63.39 \\
Object Class & 85.79 & 85.84 & \textbf{89.79} & 87.07 \\
Multiple Objects & 52.59 & 52.53 & \textbf{63.86} & 54.51 \\
Human Action & \textbf{99.60} & 99.40 & 99.20 & \textbf{99.60} \\
Color & \textbf{77.00} & 64.02 & 71.38 & 69.35 \\
Spatial Relationship & 51.40 & 55.34 & \textbf{59.50} & 54.89 \\
Scene & 49.99 & 49.29 & 49.49 & \textbf{53.85} \\
Temporal Style & \textbf{26.45} & 25.82 & 25.30 & 26.04 \\
Appearance Style & \textbf{24.83} & 23.22 & 23.81 & 24.09 \\
Overall Consistency & \textbf{27.89} & 27.24 & 27.08 & 27.73 \\
\midrule
Quality Score & \textbf{81.89} & 80.75 & 82.16 & 81.65 \\
Semantic Score & 73.71 & 71.57 & \textbf{74.58} & 73.66 \\
Total Score & 80.25 & 78.92 & \textbf{80.65} & 80.05 \\
\bottomrule
\end{tabular}
}
\label{tab:vbench_vsd_cd}
\end{table}

\clearpage

\section{Challenges and Discussions}
\label{app_sec:challenges}
\paragraph{Long Prompt Bias.}
The experiments in Sec.~\ref{sec:exp_ablation} show that,
current models perform better for long and more descriptive prompt, which is inherited from the teacher model. The reason is hypothesized to be the well-captioned text-to-video training dataset, which emphasize detailed descriptions. With longer prompts, the text-video alignment, understanding of object relationships, and depiction of motion are generally more robust and accurate. To address this issue, the performance gap with short prompts could be reduced by incorporating more short-prompt datasets during training or fine-tuning.
Additional results illustrating this phenomenon are provided in  Fig.~\ref{fig:prompt_length_comparison}, with corresponding video samples available on the website.

\begin{figure}[htbp]
\begin{center}
\includegraphics[width=0.95\columnwidth]{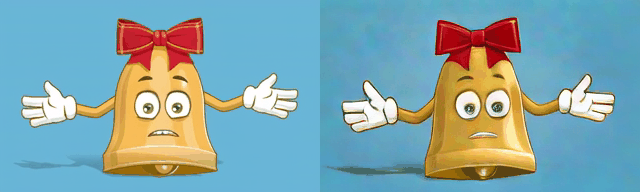}
\includegraphics[width=0.95\columnwidth]{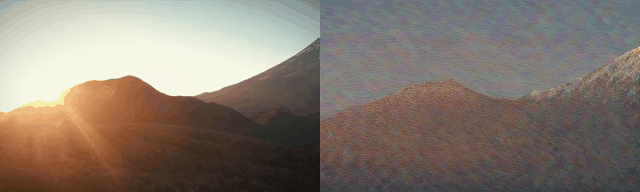}
\includegraphics[width=0.95\columnwidth]{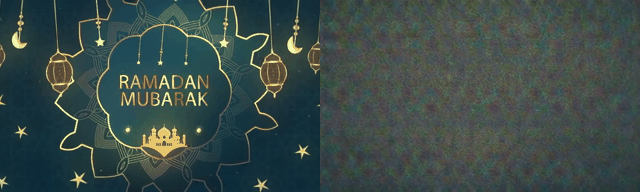}
\end{center} 
\caption{Reward model fine-tuning with dynamic degree: (left) ground-truth training samples, (right) generated samples. The noise level increases as training goes longer (from top to bottom).}
\label{fig:lrm_dd_vis}
\end{figure}

\paragraph{Reward Overoptimization.} As demontrated by experiments in Sec.~\ref{sec:exp_ablation}, the reward overoptimization issue sometimes happens for some certain reward metrics for both LRM and DDPO methods, with examples visualized in Fig.~\ref{fig:lrm_dd_vis}. To address this issue, early stopping or checkpoint selection can be one rescue. Another approach involves incorporating additional explicit regularization to constrain the student model with the teacher model during the reward fine-tuning process, and implicit regularization like diffusion loss or VSD loss may not be sufficient for this purpose. Beyond early stopping and careful tuning of the loss coefficients between data modeling and reward tuning, adopting the memoryless noise schedule~\cite{domingo2024adjoint} shows promise in steering the model toward correctly converging to tilted distributions. Further investigation into these strategies and their effectiveness in resolving overoptimization remains an important direction for future work.

\paragraph{Diversity.} From the Vendi score diversity measure of generated samples in main paper, we verifies the effectiveness of incorporating additional CD loss for improving the sample diversity. However, both qualitative comparisons and visual inspections reveal that a diversity gap remains between the distilled student models and the teacher model. 

While prior research predominantly emphasizes sample quality, the diversity of T2V models is crucial for practical applications, where diverse outputs are often necessary. This aspect of diversity remains underrepresented even in the comprehensive VBench evaluation, highlighting an area that warrants further attention and improvement.

\paragraph{Misalignment in Evaluation.}
As discussed in Sec.~\ref{app_sec:human_eval}, our experiments reveal a misalignment between VBench scores and human evaluations for videos generated using the same set of prompts. Humans may be more sensitive to unnatural flaws in videos, which can influence their preferences differently from the automatic evaluation metrics used in VBench. This discrepancy highlights the difficulty of aligning weighted score metrics with human preferences. As a result, models that achieve higher VBench scores may not necessarily be preferred by humans, and vice versa. Given the inherent complexity of video content, relying on a single or limited set of metrics may fail to fully capture video quality. This presents a challenge for the research community to develop more comprehensive evaluation protocols that are better aligned with human preferences.


\section{Visualization}
\label{app_sec:visual}
\subsection{More Qualitative Results}
\label{app_sec:visual_methods}
\begin{figure*}[htbp]
    \centering
\includegraphics[width=0.95\textwidth]{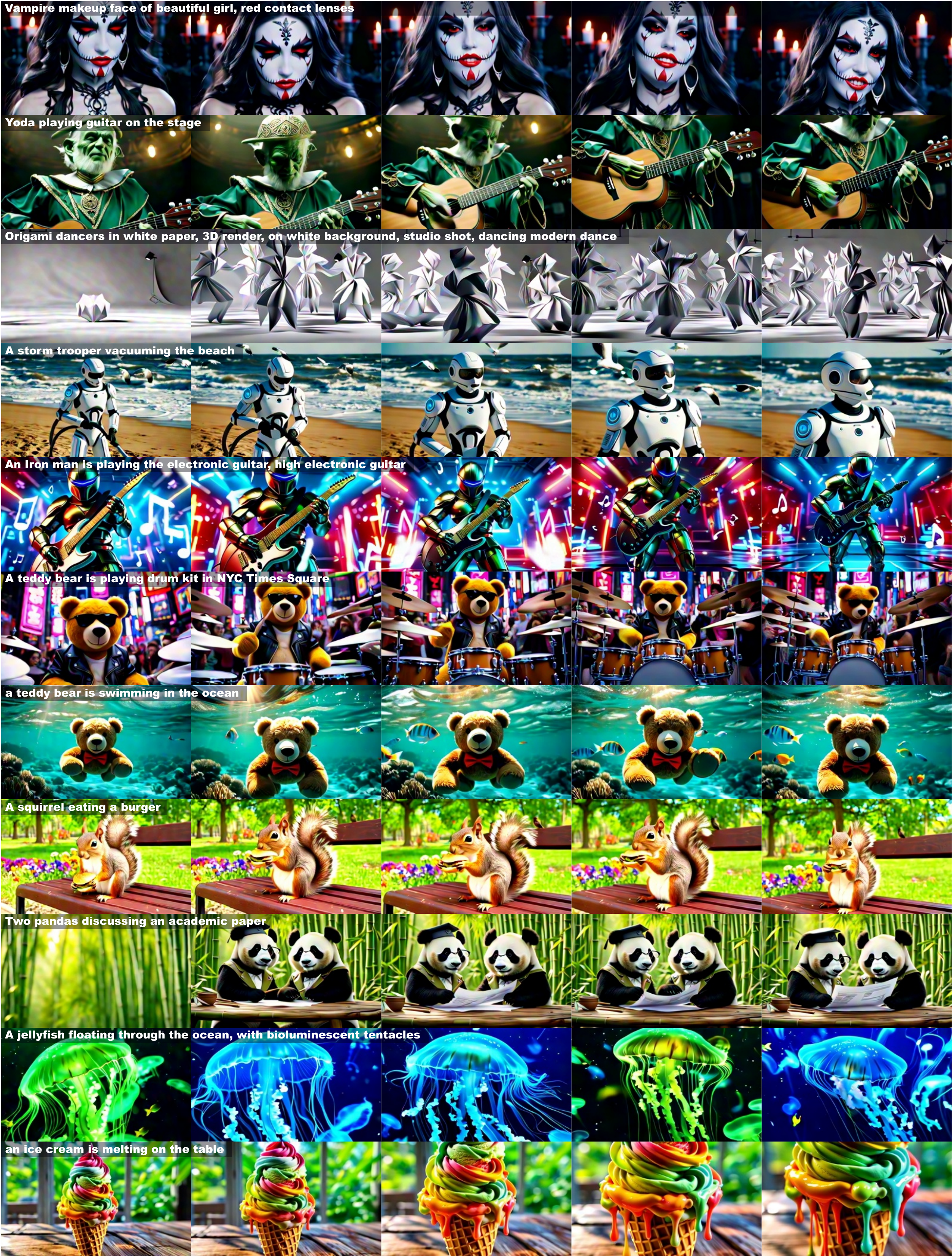}
    \caption{More qualitative results of our method (VSD+CD+LRM). Five frames are displayed for each video (frame index: 0, 30, 60, 90, 120).
    }
    \label{fig:visual_more1}
\end{figure*}

\begin{figure*}[htbp]
    \centering
\includegraphics[width=0.95\textwidth]{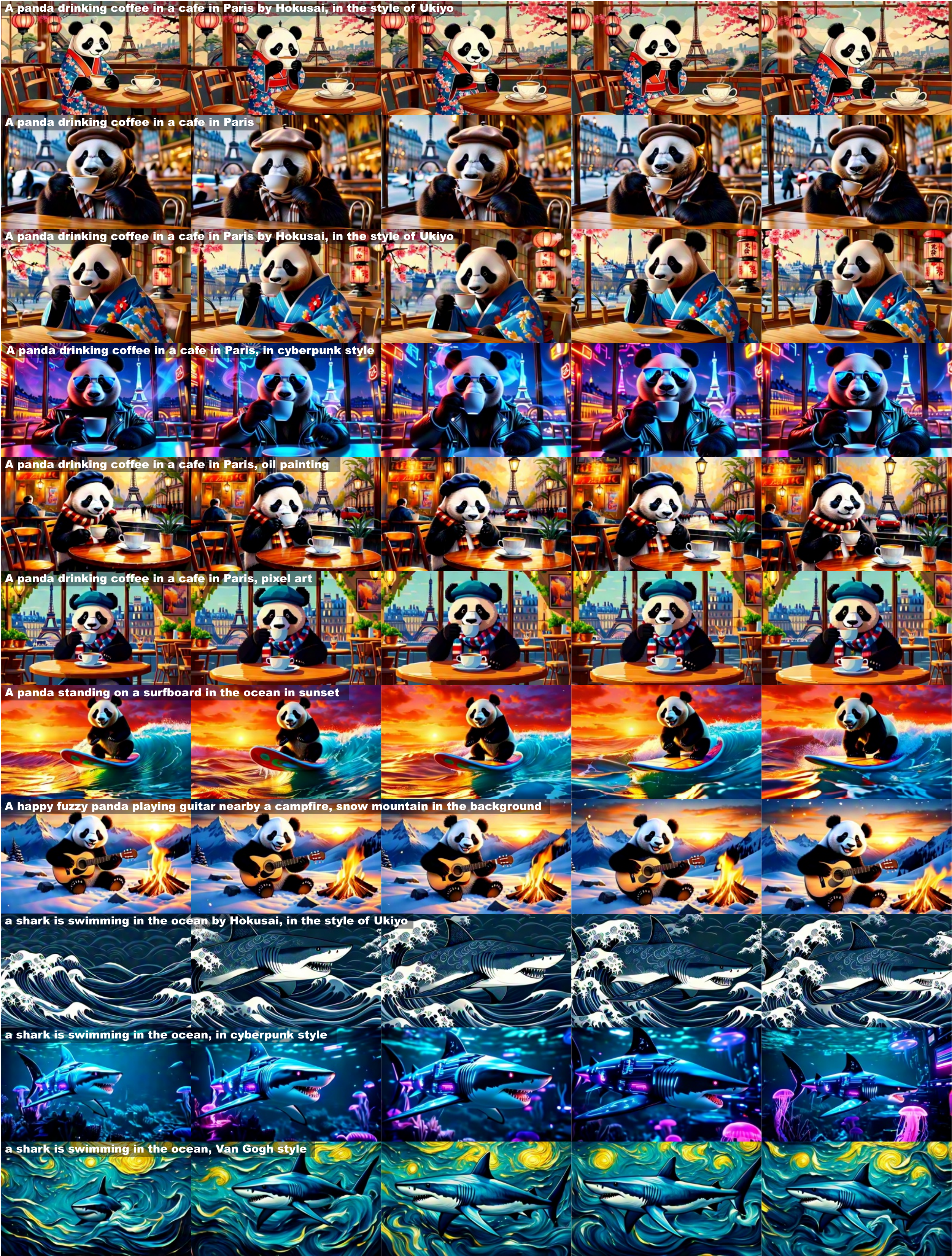}
    \caption{More qualitative results of our method (VSD+CD+LRM). Five frames are displayed for each video (frame index: 0, 30, 60, 90, 120).
    }
    \label{fig:visual_more2}
\end{figure*}

\begin{figure*}[htbp]
    \centering
\includegraphics[width=0.95\textwidth]{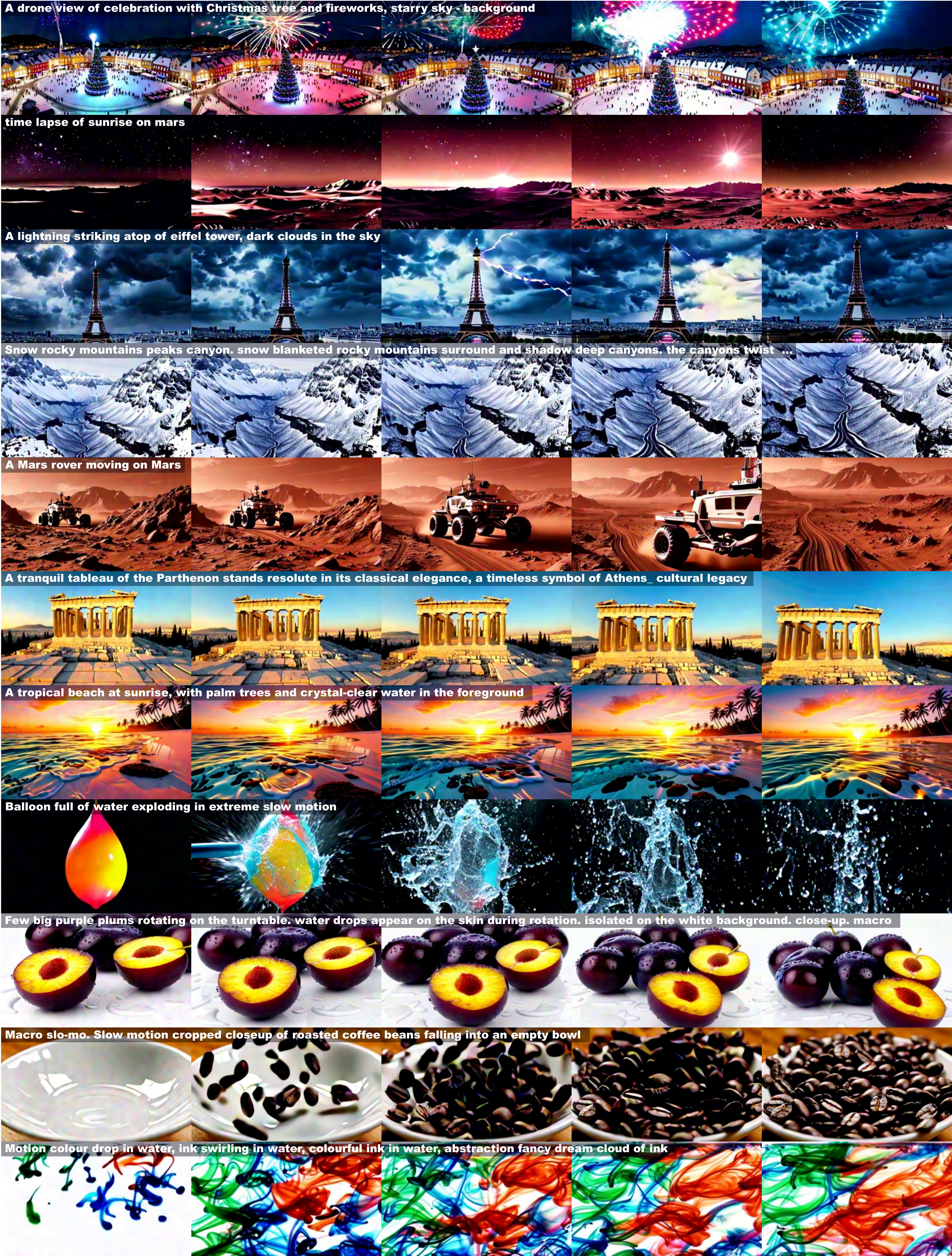}
    \caption{More qualitative results of our method (VSD+CD+LRM). Five frames are displayed for each video (frame index: 0, 30, 60, 90, 120).
    }
    \label{fig:visual_more3}
\end{figure*}

More qualitative results of our methods (VSD+CD+LRM) are displayed in Fig.~\ref{fig:visual_more1}, \ref{fig:visual_more2} and \ref{fig:visual_more3}.

\begin{figure*}[htbp]
    \centering
\includegraphics[width=0.95\textwidth]{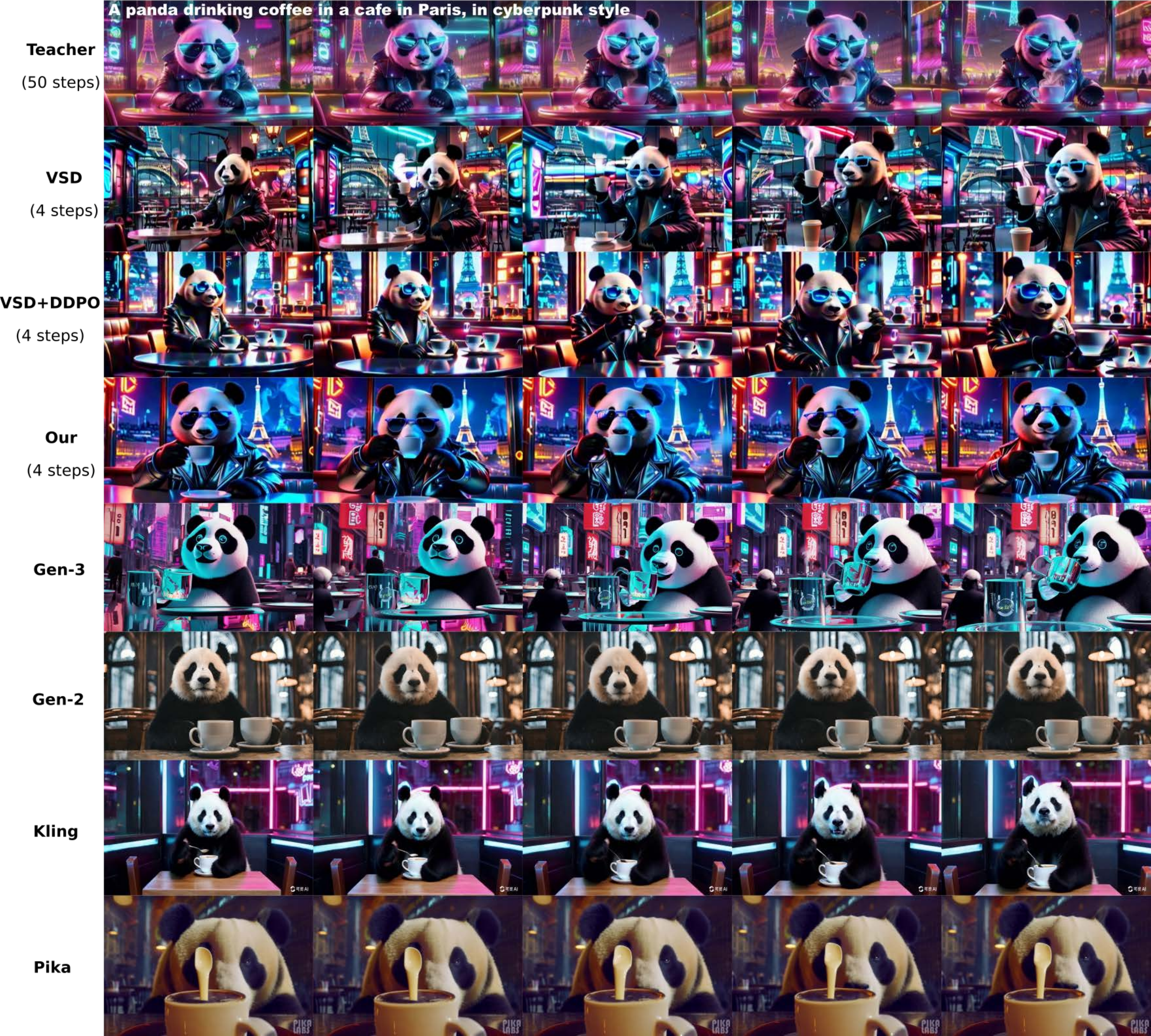}
    \caption{Comparison of our method (VSD+CD+LRM) against several baselines. Five frames are displayed for each video (frame index: 0, 30, 60, 90, 120). Videos from all baseline methods are transformed into $192\times320$ resolution for fair comparison, including Gen-2, Gen3, Kling, Pika. Our model shows superior performances in text-video alignment, motions, visual quality and fidelity.
    }
    \label{fig:compare_baseline1}
\end{figure*}

\begin{figure*}[htbp]
    \centering
\includegraphics[width=0.95\textwidth]{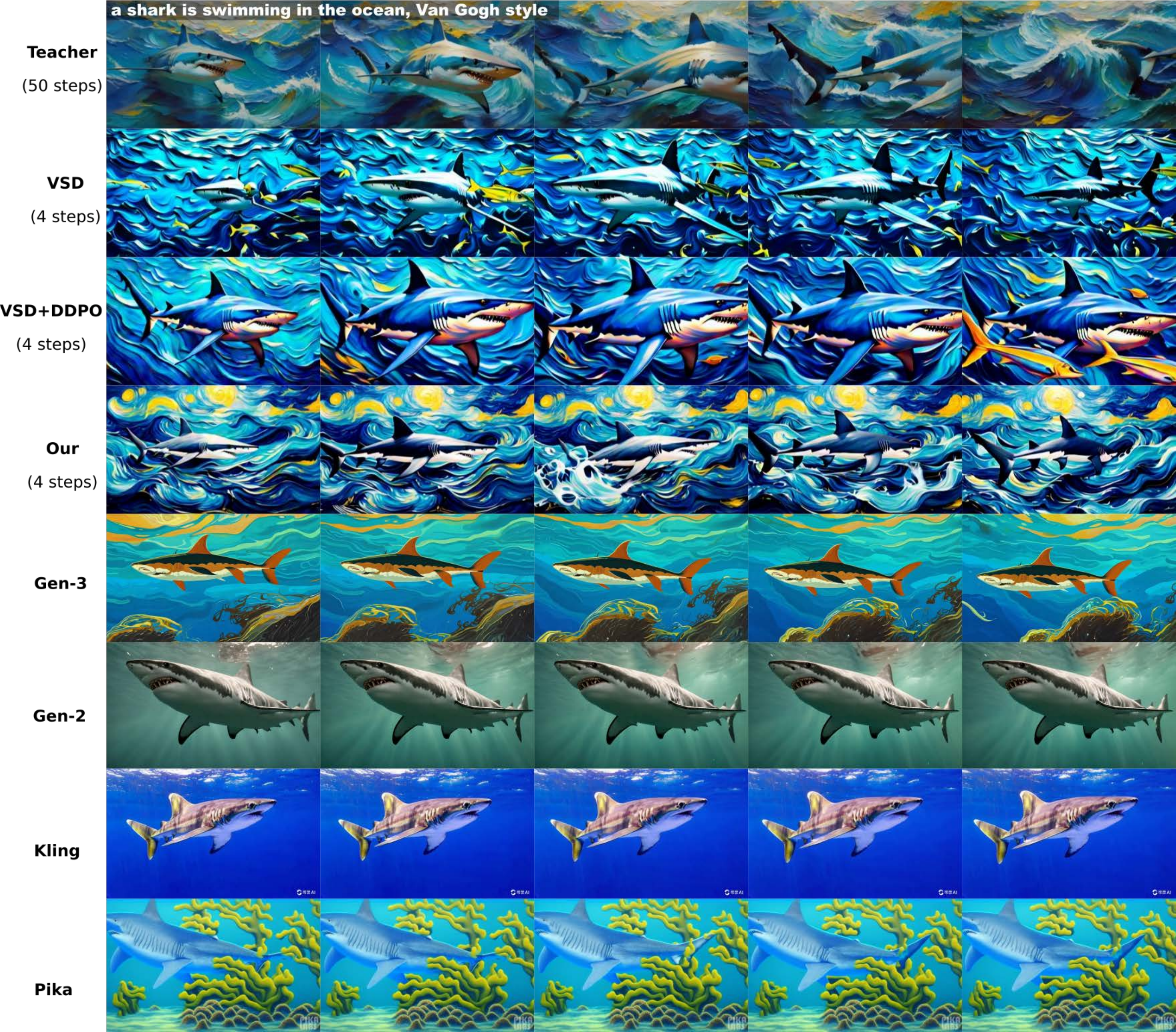}
    \caption{Comparison of our method (VSD+CD+LRM) against several baselines. Five frames are displayed for each video (frame index: 0, 30, 60, 90, 120). Videos from all baseline methods are transformed into $192\times320$ resolution for fair comparison, including Gen-2, Gen3, Kling, Pika. Our model shows superior performances in text-video alignment, motions, visual quality and fidelity.
    }
    \label{fig:compare_baseline2}
\end{figure*}

Visual comparison of our methods with baselines in Tab.~\ref{tab:compare_vbench} for generated samples with the same prompt is shown in Fig.~\ref{fig:compare_baseline1} and \ref{fig:compare_baseline2}. For fair of comparison, we visualize all sampled frames with resolution $192\times 320$ as the typical sample size of our models.

\subsection{Comparison of Reward Model Fine-tuning}
\label{app_sec:visual_reward_tune}
\begin{figure*}[htbp]
    \centering
\includegraphics[width=\textwidth]{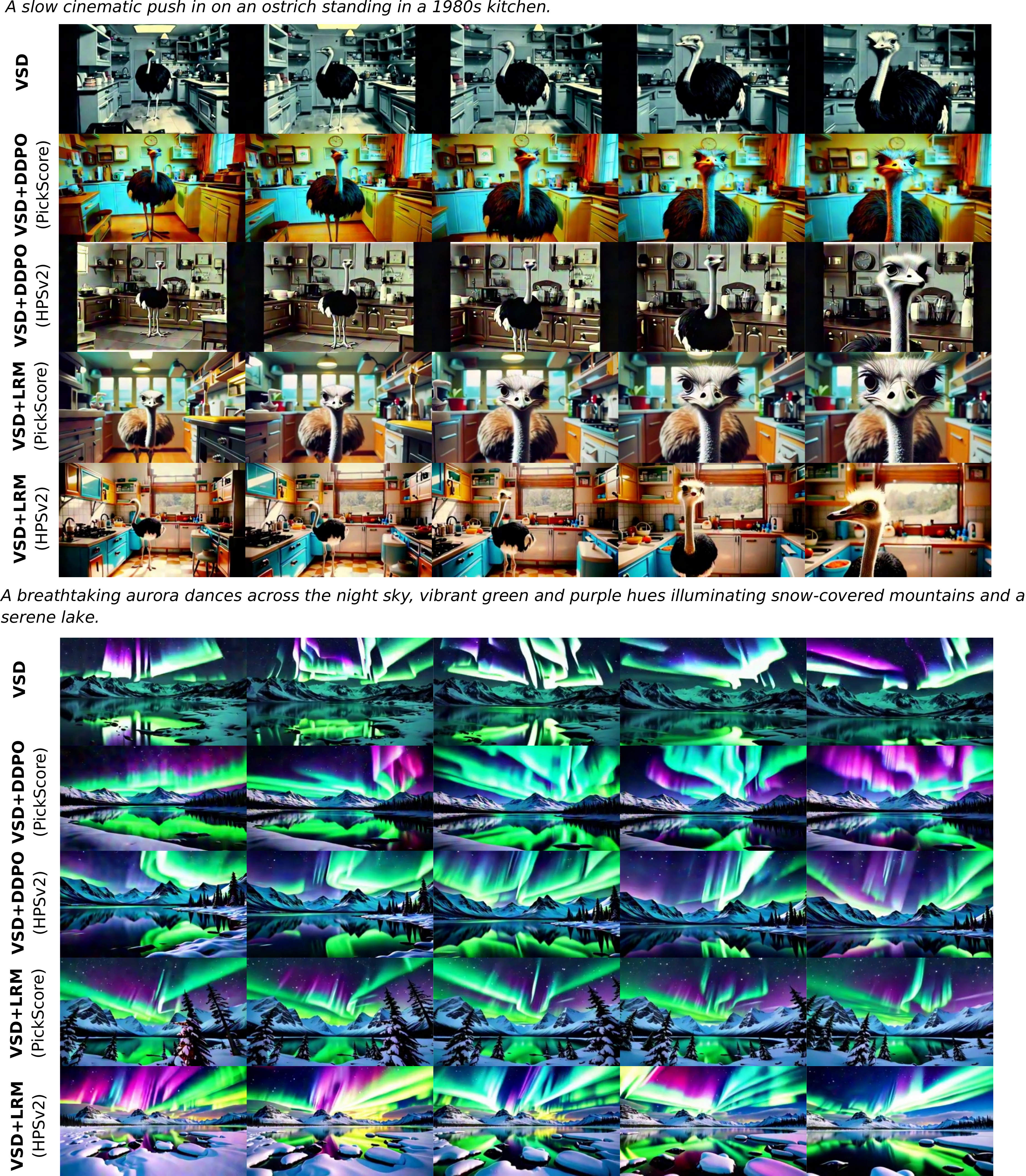}
    \caption{Visualization of video samples using different methods: VSD, VSD+DDPO(PickScore), VSD+DDPO(HPSv2), VSD+LRM(PickScore), VSD+LRM(HPSv2). Five frames are displayed for each video (frame index: 0, 30, 60, 90, 120).
    }
    \label{fig:method_compare}
\end{figure*}

\begin{figure*}[htbp]
    \centering
\includegraphics[width=\textwidth]{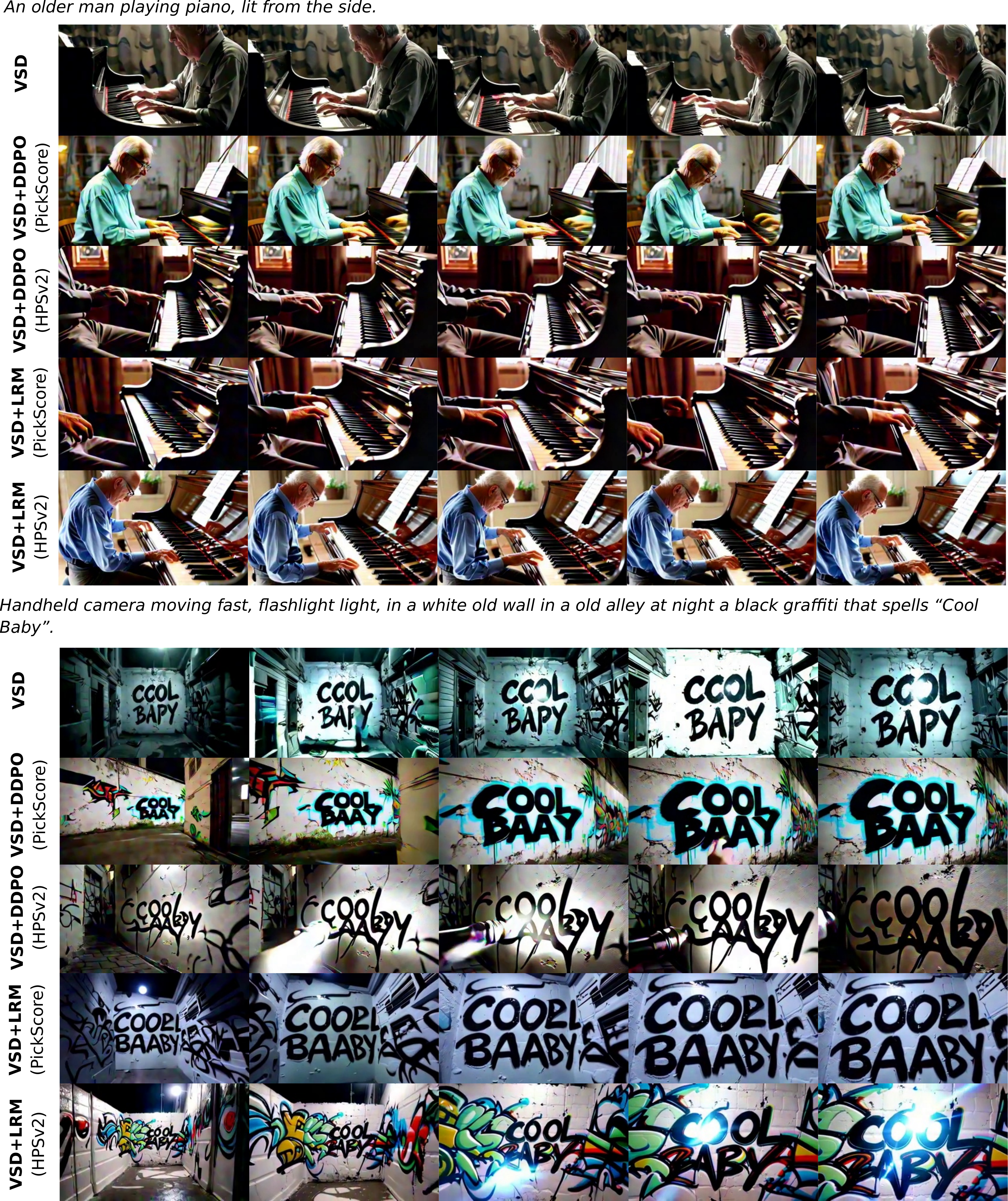}
    \caption{Visualization of video samples using different methods: VSD, VSD+DDPO(PickScore), VSD+DDPO(HPSv2), VSD+LRM(PickScore), VSD+LRM(HPSv2). Five frames are displayed for each video (frame index: 0, 30, 60, 90, 120).
    }
    \label{fig:method_compare2}
\end{figure*}

As additional results for Sec.~\ref{sec:distill_compare},
we provide visualization of samples with different reward model fine-tuning methods in Fig.~\ref{fig:method_compare} and \ref{fig:method_compare2}. It compares:
\begin{itemize}
    \item VSD;
    \item VSD with DDPO fine-tuning, using reward PickScore;
    \item VSD with DDPO fine-tuning, using reward HPSv2;
    \item VSD with LRM fine-tuning, using reward PickScore;
    \item VSD with LRM fine-tuning, using reward HPSv2.
\end{itemize}
All results are for 4-steps sampling after the distillation process.

\subsection{Inference Steps}
\label{app_sec:infer_steps}
\begin{figure*}[htbp]
    \centering
\includegraphics[width=\textwidth]{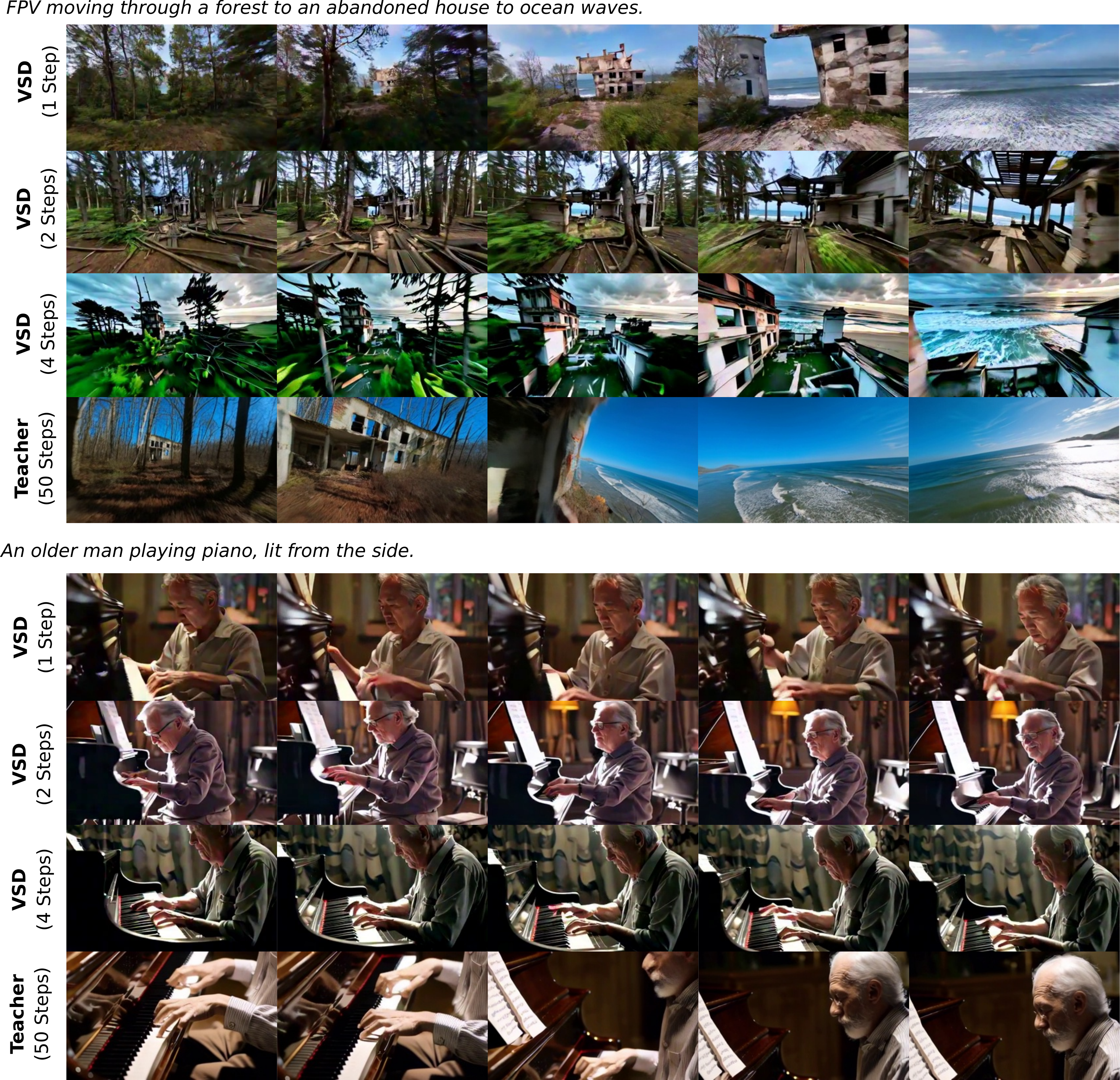}
    \caption{VSD 1, 2, 4 steps, teacher with 50 steps. Five frames are displayed for each video (frame index: 0, 30, 60, 90, 120).
    }
    \label{fig:infer_steps1}
\end{figure*}

\begin{figure*}[htbp]
    \centering
\includegraphics[width=\textwidth]{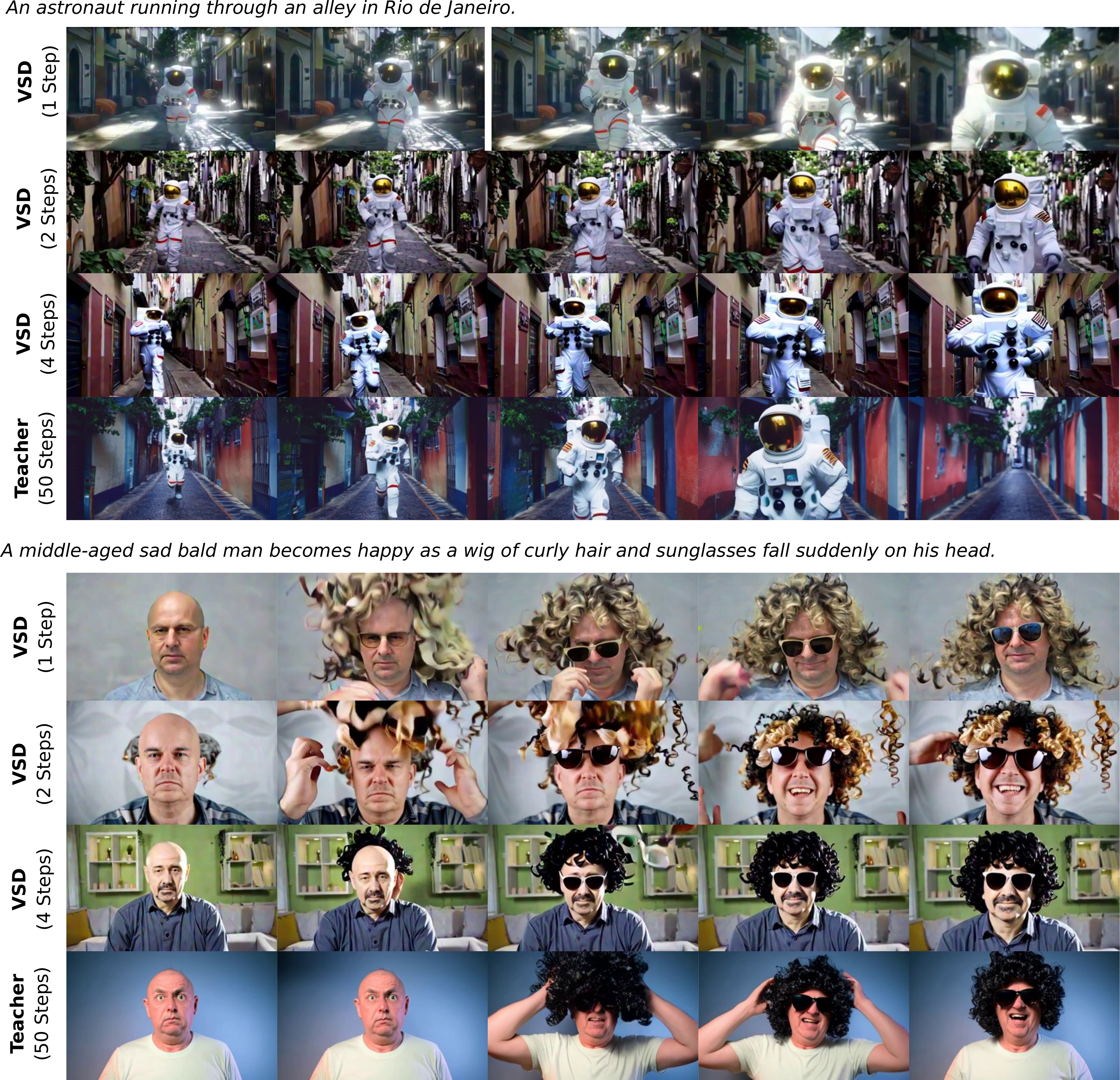}
    \caption{VSD 1, 2, 4 steps, teacher with 50 steps. Five frames are displayed for each video (frame index: 0, 30, 60, 90, 120).
    }
    \label{fig:infer_steps2}
\end{figure*}

We provide visualization of samples with different sampling steps for the VSD method, as shown in Fig.~\ref{fig:infer_steps1} and \ref{fig:infer_steps2}. During the distillation process, the sampling steps is set to be 1, 2, 4, and at inference time it follows the same step number as in distillation. From visual inspection, it is clear to show that a larger number of sampling steps usually leads to better performances, which may not be well captured by the slight difference of VBench scores.

\begin{figure*}[htbp]
    \centering
\includegraphics[width=\textwidth]{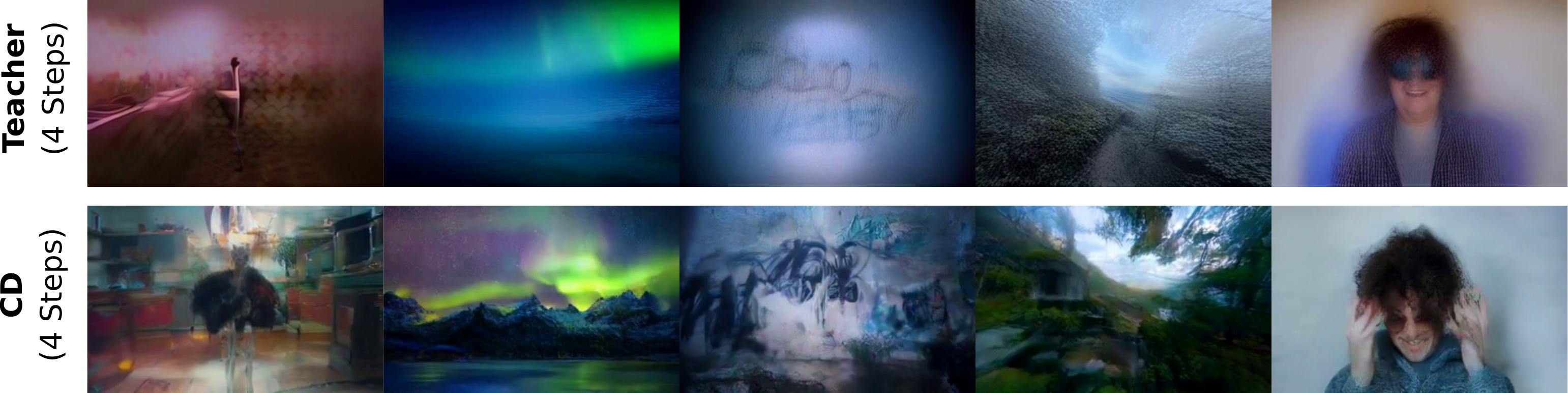}
    \caption{Sample results of 4 steps DDIM by the teacher model and 4 steps student model with CD loss. One frame for each video. Same five prompts as previous results: (from left to right) (1). \textit{A slow cinematic push in on an ostrich standing in a 1980s kitchen.} (2). \textit{A breathtaking aurora dances across the night sky, vibrant green and purple hues illuminating snow-covered mountains and a serene lake.} (3). \textit{Handheld camera moving fast, flashlight light, in a white old wall in a old alley at night a black graffiti that spells ``Cool Baby”.} (4). \textit{FPV moving through a forest to an abandoned house to ocean waves.} (5). \textit{A middle-aged sad bald man becomes happy as a wig of curly hair and sunglasses fall suddenly on his head.}
    }
    \label{fig:teacher_cd}
\end{figure*}

Fig.~\ref{fig:teacher_cd} visualizes the samples with 4-step teacher DDIM sampling, and with only CD loss for student distillation. Few-step teacher sampling without any distillation cannot generate high-quality samples. CD loss only tends to generate overly smoothed samples.

\subsection{Diversity}
\label{app_sec:visual_diverse}
\begin{figure*}[htbp]
    \centering
\includegraphics[width=\textwidth]{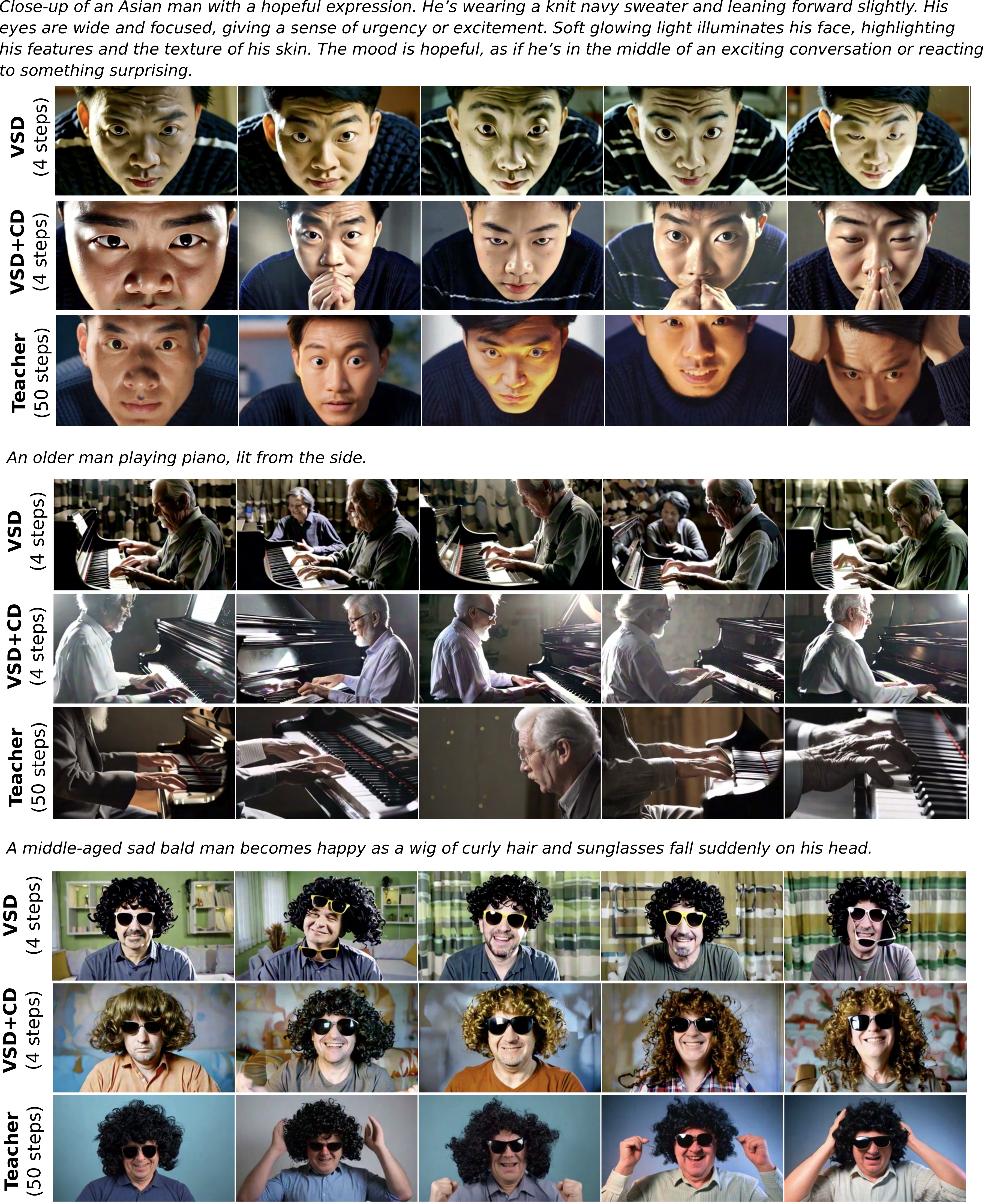}
    \caption{Diversity: VSD (top line), VSD+CD (middle  line), teacher (bottom line), one frame from each video (5 videos).
    }
    \label{fig:diverse_lrm}
\end{figure*}

\begin{figure*}[htbp]
    \centering
\includegraphics[width=\textwidth]{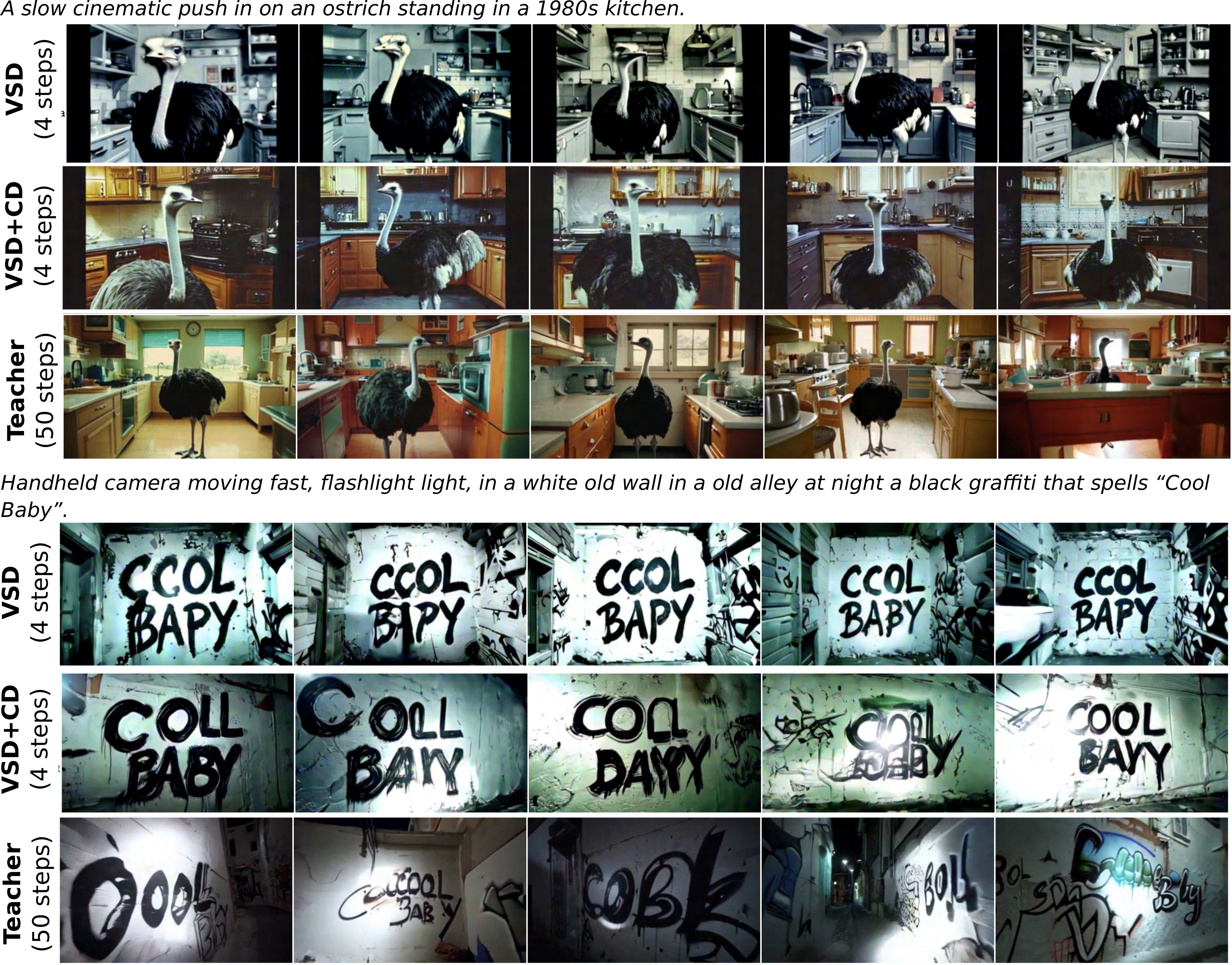}
    \caption{Diversity: VSD (top line), VSD+CD (middle  line), teacher (bottom line), one frame from each video (5 videos).
    }
    \label{fig:diverse_lrm2}
\end{figure*}

For visualizing the difference of sample diversity across different methods, we provide sample visualization as in Fig.~\ref{fig:diverse_lrm} and Fig.~\ref{fig:diverse_lrm2} for several models after training:
\begin{itemize}
    \item VSD for 4-step sampling;
    \item VSD with CD for 4-step sampling and $m=5$ for CD;
    \item Teacher model with 50 steps DDIM sampling.
\end{itemize}
The CD improves the sample diversity from both visual inspection and the quantitative measurement with Vendi score as in Tab.~\ref{tab:diversity} of the main paper.

\subsection{Prompt Length}
\label{app_sec:visual_prompt_length}
\begin{figure*}[htbp]
    \centering
    \parbox{0.98\textwidth}{\texttt{Long prompt}: \textit{A sleek, modern laptop, its screen displaying a vibrant, paused scene, sits on a minimalist wooden desk. The room is bathed in soft, natural light filtering through sheer curtains, casting gentle shadows. The laptop's keyboard is mid-illumination, with a faint glow emanating from the keys, suggesting a moment frozen in time. Dust particles are suspended in the air, caught in the light, adding to the stillness. A steaming cup of coffee beside the laptop remains untouched, with wisps of steam frozen in mid-air. The scene captures a serene, almost magical pause in an otherwise bustling workspace.}
}
\\
\includegraphics[width=0.19\textwidth]{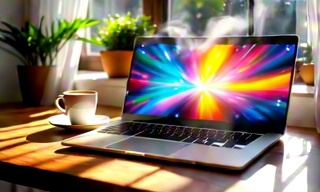}
\includegraphics[width=0.19\textwidth]{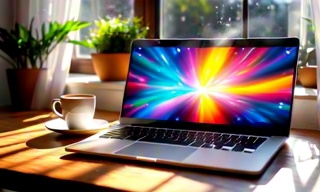}
\includegraphics[width=0.19\textwidth]{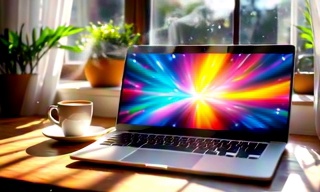}
\includegraphics[width=0.19\textwidth]{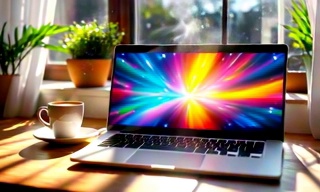}
\includegraphics[width=0.19\textwidth]{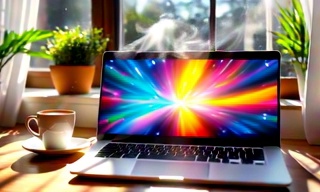}
\parbox{0.98\textwidth}{\texttt{Short prompt}: \textit{a laptop, frozen in time.}
}
\\
\includegraphics[width=0.19\textwidth]{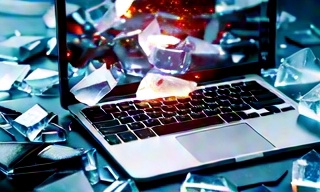}
\includegraphics[width=0.19\textwidth]{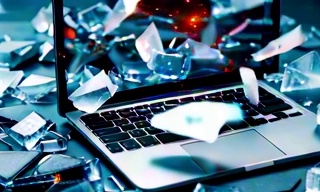}
\includegraphics[width=0.19\textwidth]{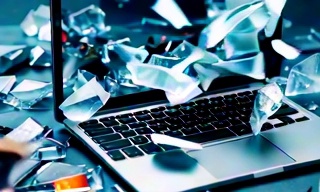}
\includegraphics[width=0.19\textwidth]{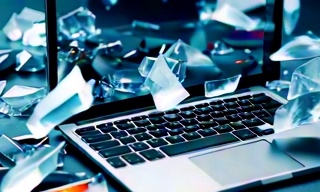}
\includegraphics[width=0.19\textwidth]{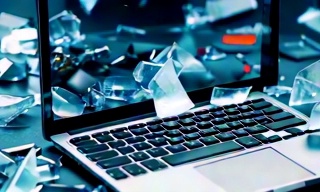}
\parbox{0.98\textwidth}{\texttt{Long prompt}: \textit{A serene nursery bathed in soft morning light reveals a cozy crib with pastel-colored bedding. A baby, dressed in a cute onesie adorned with tiny stars, stirs gently. The camera captures the baby's delicate eyelashes fluttering open, revealing curious, sleepy eyes. The baby stretches tiny arms and legs, yawning adorably. A mobile with soft, plush animals gently spins above, casting playful shadows. The room is filled with the soft hum of a lullaby, creating a peaceful atmosphere as the baby slowly awakens, ready to greet the new day with innocent wonder.}
}
\\
\includegraphics[width=0.19\textwidth]{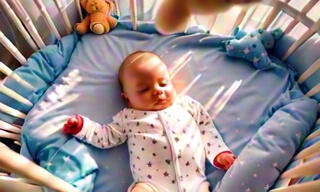}
\includegraphics[width=0.19\textwidth]{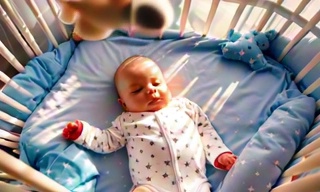}
\includegraphics[width=0.19\textwidth]{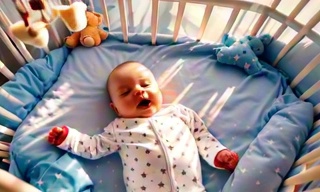}
\includegraphics[width=0.19\textwidth]{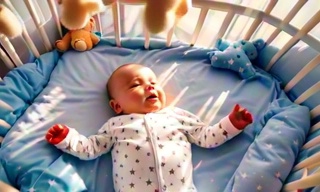}
\includegraphics[width=0.19\textwidth]{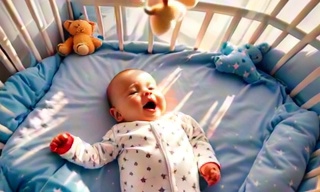}
\parbox{0.98\textwidth}{\texttt{Short prompt}: \textit{A person is baby waking up.}
}
\\
\includegraphics[width=0.19\textwidth]{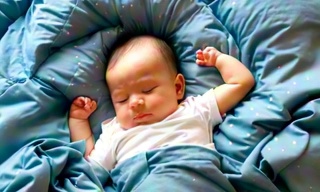}
\includegraphics[width=0.19\textwidth]{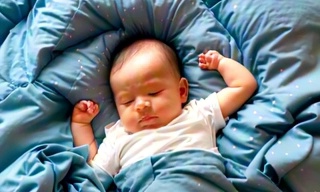}
\includegraphics[width=0.19\textwidth]{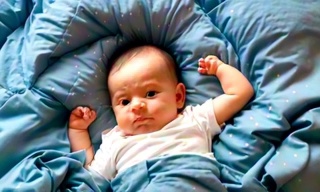}
\includegraphics[width=0.19\textwidth]{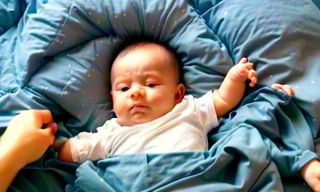}
\includegraphics[width=0.19\textwidth]{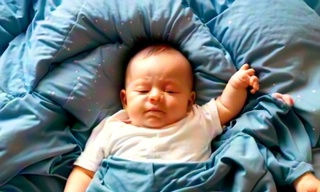}
\parbox{0.98\textwidth}{\texttt{Long prompt}: \textit{A single, perfectly ripe pear rests on a rustic wooden table, its golden-green skin glistening under soft, natural light. The pear's surface is dotted with tiny, delicate freckles, and its curved stem casts a gentle shadow. The background is a blurred, warm-toned kitchen scene, with hints of vintage decor and a window letting in a soft, diffused glow. The stillness of the frame captures the pear's natural beauty and simplicity, evoking a sense of calm and timelessness.}
}
\\
\includegraphics[width=0.19\textwidth]{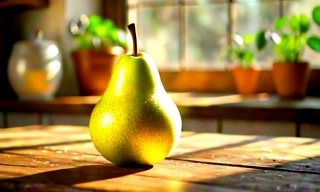}
\includegraphics[width=0.19\textwidth]{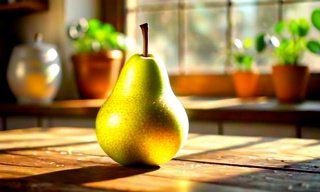}
\includegraphics[width=0.19\textwidth]{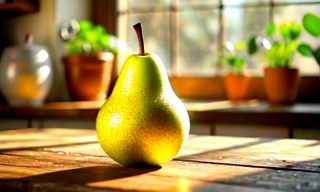}
\includegraphics[width=0.19\textwidth]{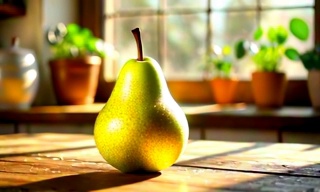}
\includegraphics[width=0.19\textwidth]{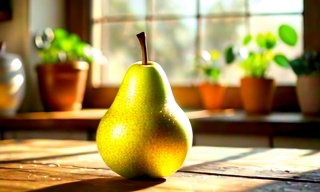}
\parbox{0.98\textwidth}{\texttt{Short prompt}: \textit{In a still frame, a pear.}
}
\\
\includegraphics[width=0.19\textwidth]{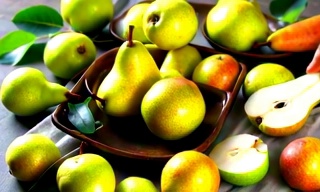}
\includegraphics[width=0.19\textwidth]{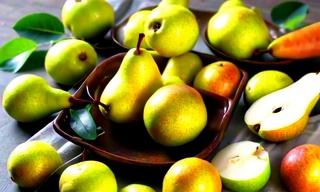}
\includegraphics[width=0.19\textwidth]{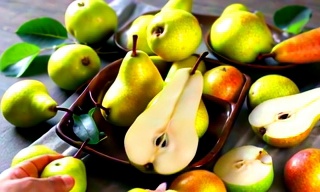}
\includegraphics[width=0.19\textwidth]{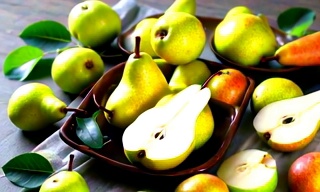}
\includegraphics[width=0.19\textwidth]{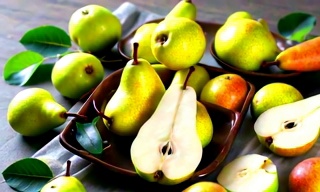}
    \caption{Comparison of sampled videos for VBench long and short prompts. Five frames are displayed for each video (frame index: 0, 30, 60, 90, 120).
    }
    \label{fig:prompt_length_comparison}
\end{figure*}

As additional results to Sec.~\ref{sec:exp_ablation}, we visualize samples with long descriptive prompts and corresponding short prompts in Fig.~\ref{fig:prompt_length_comparison}. It further verifies the hypothesis that the trained models tend to align the videos better with longer and more descriptive prompts. According to this results, the VBench evaluation in our experiments takes the long prompts for video generation by default.

\subsection{Sampling with Various Styles and Motions}
\label{app_sec:style_motion}
\begin{figure*}[htbp]
    \centering
    \parbox{0.98\textwidth}{A beautiful coastal beach in spring, waves lapping on sand by Hokusai, in the \textbf{style of Ukiyo}
}
\\
\includegraphics[width=0.19\textwidth]{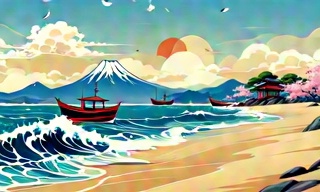}
\includegraphics[width=0.19\textwidth]{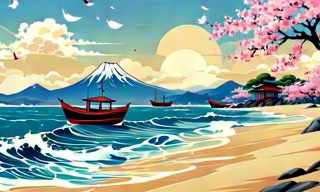}
\includegraphics[width=0.19\textwidth]{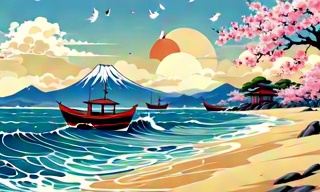}
\includegraphics[width=0.19\textwidth]{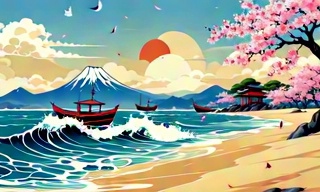}
\includegraphics[width=0.19\textwidth]{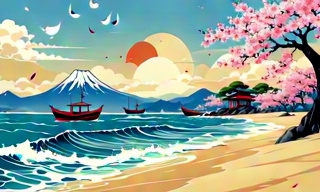}
\parbox{0.98\textwidth}{A beautiful coastal beach in spring, waves lapping on sand, in \textbf{cyberpunk style}
}
\\
\includegraphics[width=0.19\textwidth]{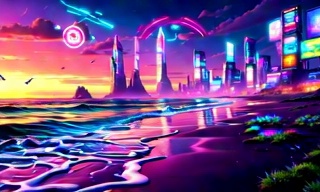}
\includegraphics[width=0.19\textwidth]{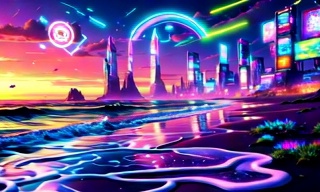}
\includegraphics[width=0.19\textwidth]{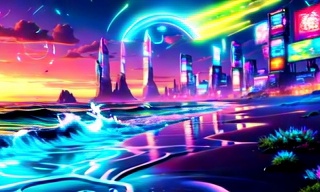}
\includegraphics[width=0.19\textwidth]{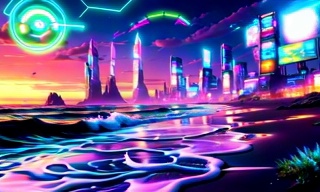}
\includegraphics[width=0.19\textwidth]{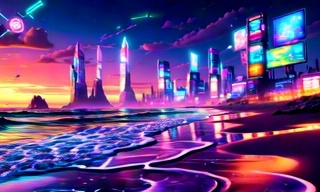}
\parbox{0.98\textwidth}{A beautiful coastal beach in spring, waves lapping on sand, \textbf{oil painting}
}
\\
\includegraphics[width=0.19\textwidth]{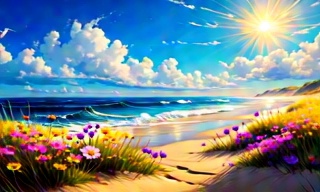}
\includegraphics[width=0.19\textwidth]{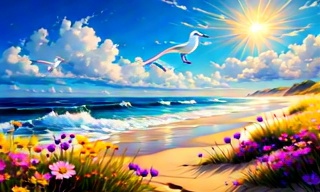}
\includegraphics[width=0.19\textwidth]{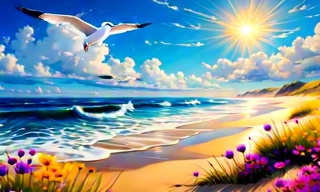}
\includegraphics[width=0.19\textwidth]{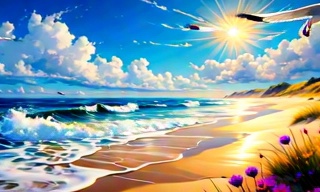}
\includegraphics[width=0.19\textwidth]{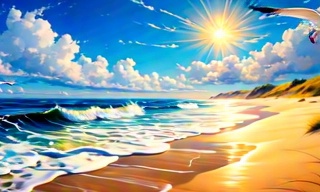}
\parbox{0.98\textwidth}{A beautiful coastal beach in spring, waves lapping on sand, \textbf{pixel art}
}
\\
\includegraphics[width=0.19\textwidth]{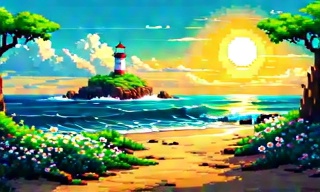}
\includegraphics[width=0.19\textwidth]{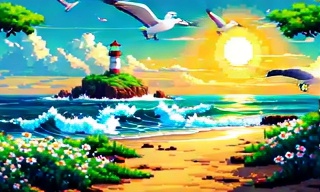}
\includegraphics[width=0.19\textwidth]{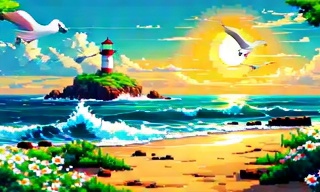}
\includegraphics[width=0.19\textwidth]{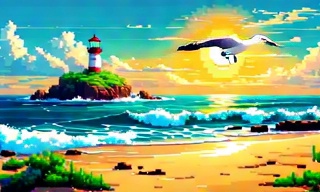}
\includegraphics[width=0.19\textwidth]{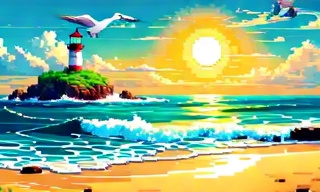}
\parbox{0.98\textwidth}{A beautiful coastal beach in spring, waves lapping on sand, \textbf{surrealism style}
}
\\
\includegraphics[width=0.19\textwidth]{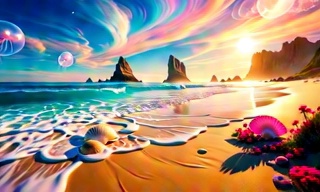}
\includegraphics[width=0.19\textwidth]{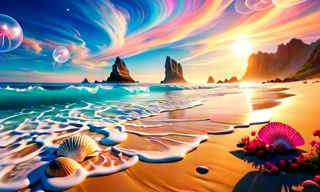}
\includegraphics[width=0.19\textwidth]{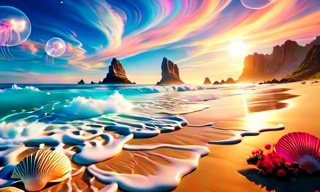}
\includegraphics[width=0.19\textwidth]{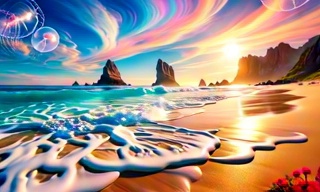}
\includegraphics[width=0.19\textwidth]{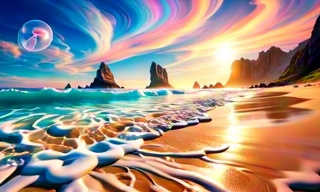}
\parbox{0.98\textwidth}{A beautiful coastal beach in spring, waves lapping on sand, \textbf{black and white}
}
\\
\includegraphics[width=0.19\textwidth]{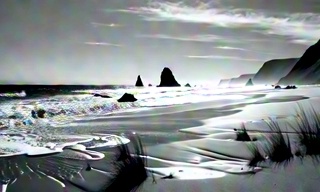}
\includegraphics[width=0.19\textwidth]{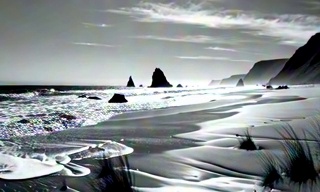}
\includegraphics[width=0.19\textwidth]{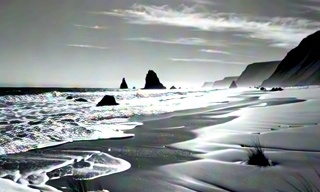}
\includegraphics[width=0.19\textwidth]{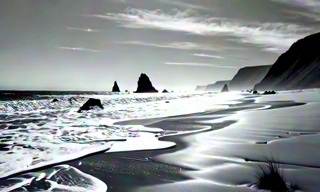}
\includegraphics[width=0.19\textwidth]{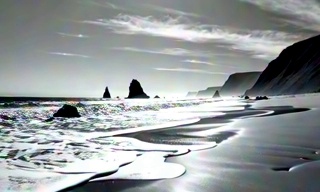}
\parbox{0.98\textwidth}{A beautiful coastal beach in spring, waves lapping on sand, \textbf{watercolor painting}
}
\\
\includegraphics[width=0.19\textwidth]{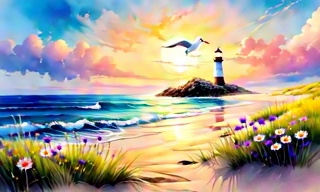}
\includegraphics[width=0.19\textwidth]{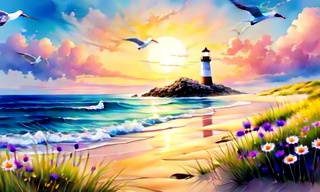}
\includegraphics[width=0.19\textwidth]{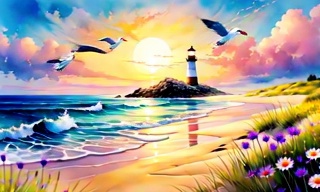}
\includegraphics[width=0.19\textwidth]{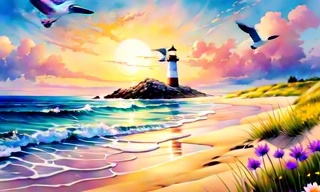}
\includegraphics[width=0.19\textwidth]{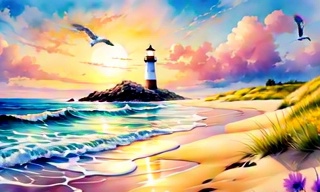}
    \caption{Video generation with diverse styles, using prompts from VBench. Five frames are extracted uniformly from one video for each prompt (with frame index: 0, 30, 60, 90, 120).
    }
    \label{fig:samples_styles}
\end{figure*}

\begin{figure*}[htbp]
    \centering
    \parbox{0.98\textwidth}{A beautiful coastal beach in spring, waves lapping on sand, \textbf{pan left}
}
\\
\includegraphics[width=0.19\textwidth]{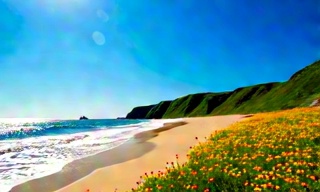}
\includegraphics[width=0.19\textwidth]{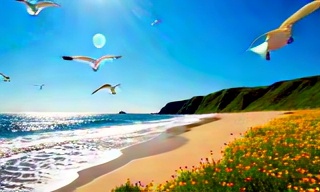}
\includegraphics[width=0.19\textwidth]{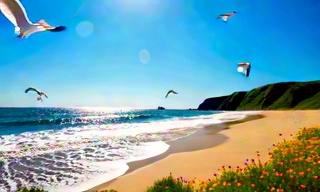}
\includegraphics[width=0.19\textwidth]{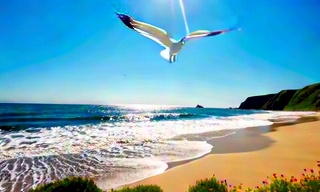}
\includegraphics[width=0.19\textwidth]{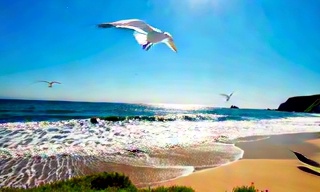}
    \parbox{0.98\textwidth}{A beautiful coastal beach in spring, waves lapping on sand, \textbf{tilt down}
}
\\
\includegraphics[width=0.19\textwidth]{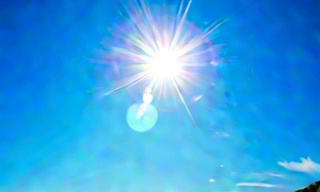}
\includegraphics[width=0.19\textwidth]{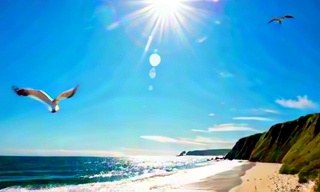}
\includegraphics[width=0.19\textwidth]{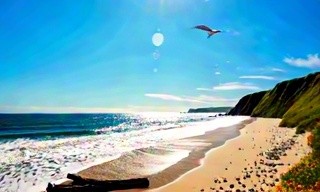}
\includegraphics[width=0.19\textwidth]{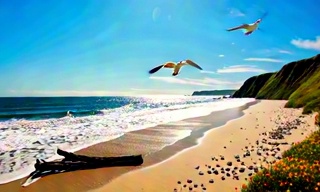}
\includegraphics[width=0.19\textwidth]{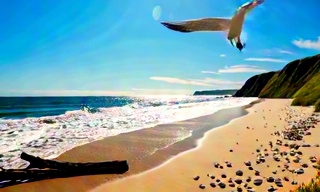}
    \parbox{0.98\textwidth}{A beautiful coastal beach in spring, waves lapping on sand, \textbf{tilt up}
}
\\
\includegraphics[width=0.19\textwidth]{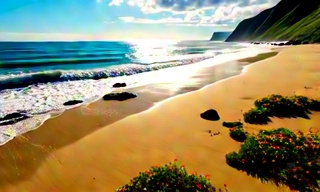}
\includegraphics[width=0.19\textwidth]{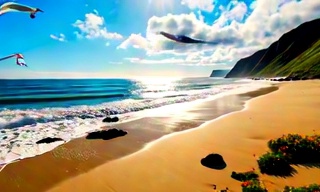}
\includegraphics[width=0.19\textwidth]{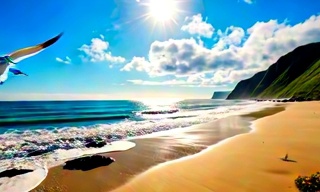}
\includegraphics[width=0.19\textwidth]{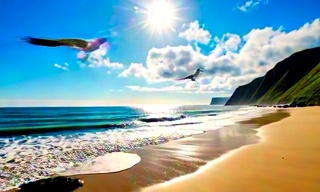}
\includegraphics[width=0.19\textwidth]{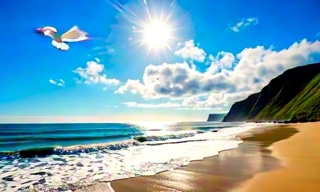}
    \parbox{0.98\textwidth}{A beautiful coastal beach in spring, waves lapping on sand, \textbf{zoom in}
}
\\
\includegraphics[width=0.19\textwidth]{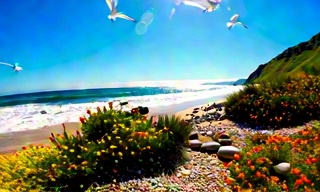}
\includegraphics[width=0.19\textwidth]{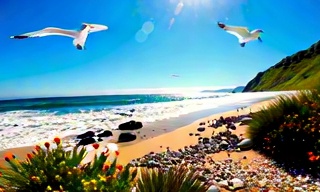}
\includegraphics[width=0.19\textwidth]{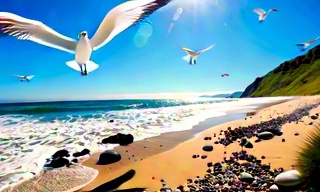}
\includegraphics[width=0.19\textwidth]{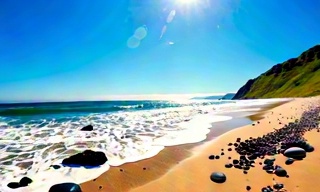}
\includegraphics[width=0.19\textwidth]{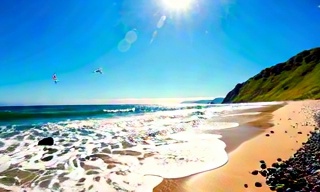}
    \parbox{0.98\textwidth}{A beautiful coastal beach in spring, waves lapping on sand, \textbf{racking focus}
}
\\
\includegraphics[width=0.19\textwidth]{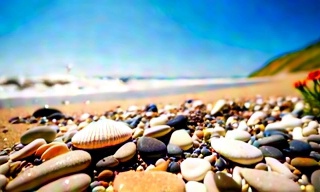}
\includegraphics[width=0.19\textwidth]{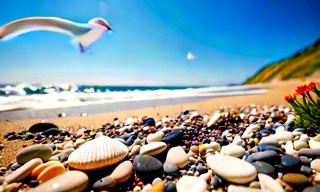}
\includegraphics[width=0.19\textwidth]{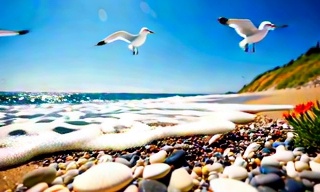}
\includegraphics[width=0.19\textwidth]{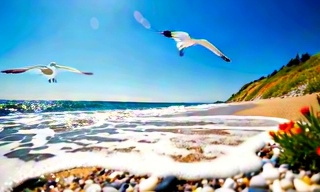}
\includegraphics[width=0.19\textwidth]{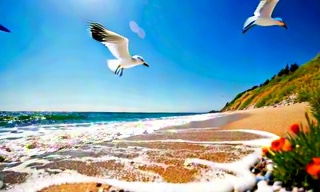}
    \parbox{0.98\textwidth}{A beautiful coastal beach in spring, waves lapping on sand, \textbf{in super slow motion}
}
\\
\includegraphics[width=0.19\textwidth]{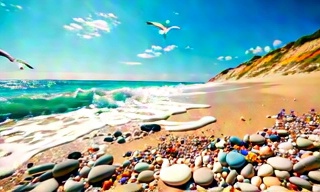}
\includegraphics[width=0.19\textwidth]{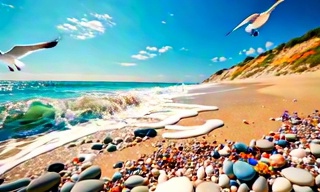}
\includegraphics[width=0.19\textwidth]{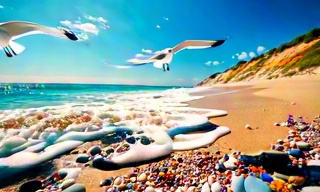}
\includegraphics[width=0.19\textwidth]{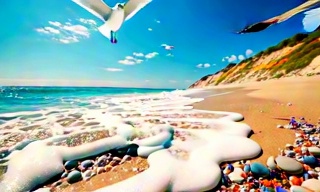}
\includegraphics[width=0.19\textwidth]{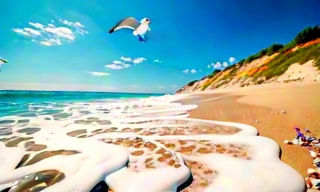}
    \caption{Video generation with diverse camera motions, using prompts from VBench. Five frames are extracted uniformly from one video for each prompt (with frame index: 0, 30, 60, 90, 120).
    }
    \label{fig:samples_motions}
\end{figure*}

The distilled student models with the proposed methods demonstrate great performances over various styles in prompts, including different artistic styles like \textit{Ukiyo style, cuberpunk, surrealism, pixel art, oil painting, watercolor painting, black and white}, etc. It also supports different camera motions in the video, like \textit{pan left, pan right, tilt down, tilt up, zoom in, racking focus}, etc. The visualization for generated samples with various styles and camera motions is shown in Fig.~\ref{fig:samples_styles} and Fig.~\ref{fig:samples_motions}.

\newpage

\end{document}